\documentclass[10pt,twocolumn,letterpaper]{article}

\usepackage[pagenumbers]{iccv} % To force page numbers, e.g. for an arXiv version

\usepackage{comment}
\usepackage{wrapfig}
\usepackage{multirow}
\usepackage{amsmath}
\usepackage{dsfont}
\usepackage{amssymb}
\usepackage{listings}
\usepackage{placeins}

\usepackage{soul}
\newcommand{\customformat}[3]{
    \noindent\textbf{\textcolor{violet}{#1}} % Bold text and color
    {\footnotesize\textcolor{darkgray}{#2}} % Normal text and color
    \par
    #3 % Normal text
    \\ \\ 
}

\usepackage[accsupp]{axessibility}  % Improves PDF readability for those with disabilities.

\definecolor{iccvblue}{rgb}{0.21,0.49,0.74}
\usepackage[pagebackref,breaklinks,colorlinks,allcolors=iccvblue]{hyperref}

\makeatletter
\def\thanks#1{\protected@xdef\@thanks{\@thanks
        \protect\footnotetext{#1}}}
\makeatother
\newcommand*\samethanks[1][\value{footnote}]{\footnotemark[#1]}
\def\paperID{4181} 
\def\confName{ICCV}
\def\confYear{2025}

\title{CNS-Bench: Benchmarking Image Classifier Robustness \\Under Continuous Nuisance Shifts}

\author{%
Olaf Dünkel$^{1}$\thanks{$^\dag$\ Corresponding author: \texttt{oduenkel@mpi-inf.mpg.de}.}\samethanks[2] \quad Artur Jesslen$^{2}$\thanks{*\ Equal contribution.}\samethanks[1] \quad Jiahao Xie$^{1}$\samethanks[1]\\ Christian Theobalt$^{1,3}$ \quad Christian Rupprecht$^{4}$ \quad Adam Kortylewski$^{1,2}$\\
{\normalsize$^{1}$Max Planck Institute for Informatics, Saarland Informatics Campus \quad $^{2}$University of Freiburg}\\ {\normalsize $^{3}$Saarbrücken Research Center for Visual Computing, Interaction and AI \quad $^{4}$University of Oxford}
}

\begin{document}
\maketitle

\begin{abstract}

An important challenge when using computer vision models in the real world is to evaluate their performance in potential out-of-distribution (OOD) scenarios. 
While simple synthetic corruptions are commonly applied to test OOD robustness, they often fail to capture nuisance shifts that occur in the real world. 
Recently, diffusion models have been applied to generate realistic images for benchmarking, but they are restricted to binary nuisance shifts.
In this work, we introduce \textbf{CNS-Bench}, a \textbf{C}ontinuous \textbf{N}uisance \textbf{S}hift \textbf{Bench}mark to quantify OOD robustness of image classifiers for continuous and realistic generative nuisance shifts.
CNS-Bench allows generating a wide range of individual nuisance shifts in continuous severities by applying LoRA adapters to diffusion models. To address failure cases, we propose a filtering mechanism that outperforms previous methods, thereby enabling reliable benchmarking with generative models.
With the proposed benchmark, we perform a large-scale study to evaluate the robustness of more than 40 classifiers under various nuisance shifts.
Through carefully designed comparisons and analyses, we find that model rankings can change for varying shifts and shift scales, which cannot be captured when applying common binary shifts.
Additionally, we show that evaluating the model performance on a continuous scale allows the identification of model failure points, providing a more nuanced understanding of model robustness. 
Project page including code and data: {\small\url{https://genintel.github.io/CNS}}.

\end{abstract}

\begin{figure}[t]
  \centering
  \includegraphics[width=1.\linewidth]{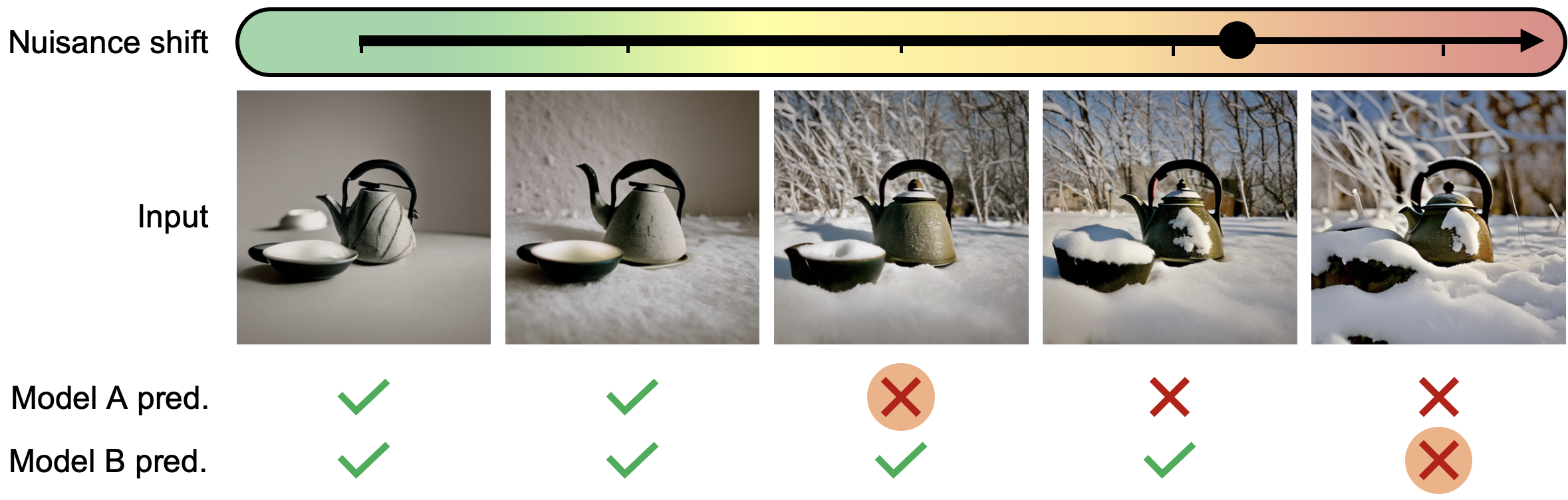}
  \caption{\textbf{Benchmarking under continuous nuisance shifts.} We evaluate the robustness of different models under gradually increasing nuisance shifts. This enables failure point identification (highlighted in red).}
  \label{fig:teaser}
  \vspace{-5pt}
\end{figure}

\section{Introduction}
Machine learning models are typically validated and tested on fixed datasets under the assumption of independent and identically distributed samples. 
However, this does not fully cover the true capabilities and potential vulnerabilities of models when deployed in dynamic real-world environments.
The robustness in out-of-distribution (OOD) scenarios is important, and decision-makers might need to know how models perform under various distribution shifts and levels severity.
Therefore, it is crucial to continue building richer and more systematic benchmarks.

Strategies for collecting out-of-distribution (OOD) images for such benchmarks involve manual data collection, perturbations with synthetic corruptions \citep{zhao_ood-cv_2022,hendrycks_many_2021,hendrycks2021nae}, or rendering from synthetic objects \cite{bordes2024pug,shu_identifying_2019}.
Recently, text-to-image (T2I) diffusion models have been introduced as promising tools for benchmarking images in a scalable manner \citep{mofayezi_benchmarking_2023, metzen_identification_2023, vendrowDatasetInterfacesDiagnosing2023, zhang_imagenet-d_2024}.

However, all previous approaches define \emph{categorical} or \emph{binary} nuisance shifts by considering the existence or absence of a shift, which contradicts their continuous realization in real-world scenarios.
For example, as shown in \cref{fig:teaser}, the snow level in an environment can range from light snowfall to objects fully covered with snow. 
While one model might fail at all snow levels, some models may only fail under heavy occlusion.
In a real-world application, an autonomous driving company might want to know how the system's performance deteriorates for stronger distribution shifts.
The seminal work ImageNet-C~\cite{hendrycks_benchmarking_2019} has illustrated through simple corruptions that classifier A can have a lower overall performance than classifier B, even though classifier A degrades more gracefully in case of corruptions and hence might be preferable over classifiers that degrade suddenly.
However, this is not yet possible for continuous real-world nuisance shifts.

To overcome this shortcoming in current benchmarks, we establish a \textbf{C}ontinuous \textbf{N}uisance \textbf{S}hift \textbf{Bench}mark for image classifier robustness, dubbed as \textbf{CNS-Bench}.
Building on top of T2I diffusion models (\eg, Stable Diffusion~\cite{rombach2022high}), we enable realistic and continuous nuisance shifts.
Specifically, we leverage LoRA~\citep{hu_lora_2021} adapters to learn ImageNet~\cite{deng2009imagenet} class-specific shift sliders \cite{vendrowDatasetInterfacesDiagnosing2023}. 
In contrast to previous works conducting analysis on \textit{binary} shifts, our study motivates the consideration of multiple shift scales.
This led to the observation that model rankings can change when considering different shift severities.
Generally, measuring robustness as a spectrum instead of aggregating it into a single average metric allows a more comprehensive understanding of OOD robustness~\citep{drenkow_systematic_2022,hendrycks_many_2021}.
As a necessity for scaling up the robustness analysis, we propose a filtering mechanism that automatically removes generated samples from the benchmarking dataset that do not represent the considered class.

With the benchmark, we evaluate more than 40 classifiers and study their robustness along the following axes: 
(\romannumeral1) architecture, (\romannumeral2) number of parameters, and (\romannumeral3) pre-training paradigm and data.
Through rigorous comparisons, we reveal multiple findings:
1) Model performance drops differently across different shifts and magnitudes. 2) Visual state-space models are more robust than other architectures like vision Transformers and CNNs. 3) Self-supervised pre-training leads to stronger robustness to the presented shifts than supervised pre-training on a larger dataset.
This demonstrates that generative benchmarks open a new path for systematically studying the robustness of vision models in a controlled and scalable manner.

In summary, our work makes the following contributions:
\textbf{1)} We propose CNS-Bench to benchmark ImageNet classifiers under continuous nuisance shifts.
We publish a dataset with 14 diverse and realistic nuisance shifts representing various style and weather variations at five severity levels.
In addition, we also provide trained LoRA sliders for all shifts that can be used to compute shift levels in a fully continuous manner.
\textbf{2)} We collect an annotated dataset to benchmark OOC filtering strategies and propose a novel filtering mechanism that achieves higher filter accuracies than previously applied text-alignment-based strategies.
\textbf{3)} We evaluate the robustness of more than 40 classifiers along different axes and reveal multiple valuable findings, underlining the importance of considering continuous shift severities of real-world nuisance shifts.

\begin{table}
\caption{\textbf{Image sources for benchmarking robustness to nuisance shifts.} Existing benchmarks for evaluating classifier robustness include images collected by humans, corrupted by synthetic perturbations, generated by rendering pipelines, and generated by a text-to-image (T2I) diffusion model. Our benchmark is the first that enables benchmarking \wrt realistic and continuous nuisance shifts, scalable with respect to the number of classes and shifts.}\label{tab:related_work}
\centering
\begin{small}
  \centering
  \begin{tabular}{lccc}
    \toprule
    Image source& Real. & Scalable&Continuous\\
    \midrule
    Human \cite{zhao_ood-cv_2022,zhao_ood-cv-v2_2023,hendrycks_many_2021}& \checkmark& &\\ 
    Synthetic \cite{hendrycks_benchmarking_2019,kar_3d_2022}& & \checkmark&\checkmark\\
    Rendered \cite{bordes2024pug,shu_identifying_2019,kar_3d_2022,li_imagenet-e_2023}& \checkmark& & \checkmark\\ %\cite{bordes2024pug}
    Gen. T2I \cite{vendrowDatasetInterfacesDiagnosing2023,zhang_imagenet-d_2024,mofayezi_benchmarking_2023, metzen_identification_2023}& \checkmark& \checkmark&\\
    Ours& \checkmark& \checkmark&\checkmark\\
    \bottomrule
  \end{tabular}
  \end{small}
  \vspace{-5pt}
\end{table}

\section{Related Work}

\noindent\textbf{Robustness.} 
When referring to robustness, we consider the relative accuracy drop of a classifier \wrt interventions that alter images from a base distribution, building upon the formalism introduced by~\citet{drenkow_systematic_2022}.
While the averaged accuracy drops provide an aggregated measure of the robustness, we consider the robustness \wrt specific nuisance shifts that can be modeled as causal interventions on the environment, the appearance, the object, or the renderer.

\noindent\textbf{Benchmarking robustness.}
Early approaches for benchmarking the performance and generalizability of models use fixed datasets, assuming independent and identically distributed samples~\citep{deng2012mnist,deng2009imagenet,cocodataset}.
However, this does not capture the performance in real-world applications where out-of-distribution (OOD) scenarios that deviate from the training distribution might occur~\cite{sun2020test,schneider2020improving,gandelsman2022test,xie2025test}. 
To tackle this challenge, various datasets have been presented that involve the manual collection of data with nuisance shifts~\citep{zhao_ood-cv_2022,hendrycks_many_2021,wang_learning_2019,geirhos2018imagenet,barbu_objectnet_2019,idrissi_imagenet-x_2022,hendrycks2021nae,recht2019imagenet,taori_measuring_2020}. 
However, these methods are often time-consuming and labor-intensive since they require data crawling and human annotations.
Moreover, they usually capture only a subset of nuisance shifts that models may encounter in the real world, and it is challenging to ensure the disentanglement of these annotated nuisances. 

On the other hand, synthetic datasets offer opportunities to evaluate deep neural networks since various instances of an object class with specified context and nuisance shifts can be generated.
One line of work applies simple synthetic corruptions to evaluate the robustness of classifiers~\cite{hendrycks_benchmarking_2019,press2023rdumb}, lacking real-world distribution shifts.

Furthermore, rendering pipelines allow the precise control of several variables and are applied for benchmarking~\citep{bordes2024pug,shu_identifying_2019,kar_3d_2022,li_imagenet-e_2023,sun_shift_2022}. However, some nuisance shifts, such as weather variations (\eg, snow), are very hard to model using traditional pipelines.
Additionally, scaling to a variety of classes is challenging since 3D assets need to be available for all considered classes.

Recent developments in diffusion models have enabled the creation of realistic and diverse synthetic benchmark datasets~\cite{mofayezi_benchmarking_2023, metzen_identification_2023, vendrowDatasetInterfacesDiagnosing2023, zhang_imagenet-d_2024}, offering greater control over nuisances (\eg, text-guided corruptions, counterfactuals).
However, unlike synthetic corruptions or rendering pipelines, these works do not support continuous modeling of distribution shifts, even though such shifts typically occur gradually in the real world and have varying effects on model performance.
While a previous study~\cite{hendrycks_benchmarking_2019} has examined simple synthetic corruptions, no prior work has addressed the need to handle continuous and realistic distribution shifts.
To bridge this gap, in this work, we propose a framework to benchmark vision models \wrt realistic nuisance shifts across continuous severity levels.

\noindent\textbf{Filtering out-of-class cases of generative models.}
When using synthetic images for benchmarking, one essential requirement is to ensure that the generated images represent the class of interest, \ie, no out-of-class (OOC) \cite{metzen_identification_2023} samples are contained.
Manually checking the quality of images to find those not aligned with the desired condition is still a common practice~\citep{zhang_imagenet-d_2024}.
However, it has difficulty scaling up the analysis~\citep{angelopoulos2023prediction}.
Removing failure cases from a set of generated images is still an open research question, which receives surprisingly low attention in the field of generative benchmarking.
With this in mind, we collect a dataset of manually annotated OOC-generated images and propose an improved filtering mechanism that outperforms a strategy relying solely on CLIP text alignment to automatically remove OOC samples \cite{vendrowDatasetInterfacesDiagnosing2023,prabhu2023lance}.

\begin{figure}
  \centering
    \includegraphics[width=1.\linewidth]{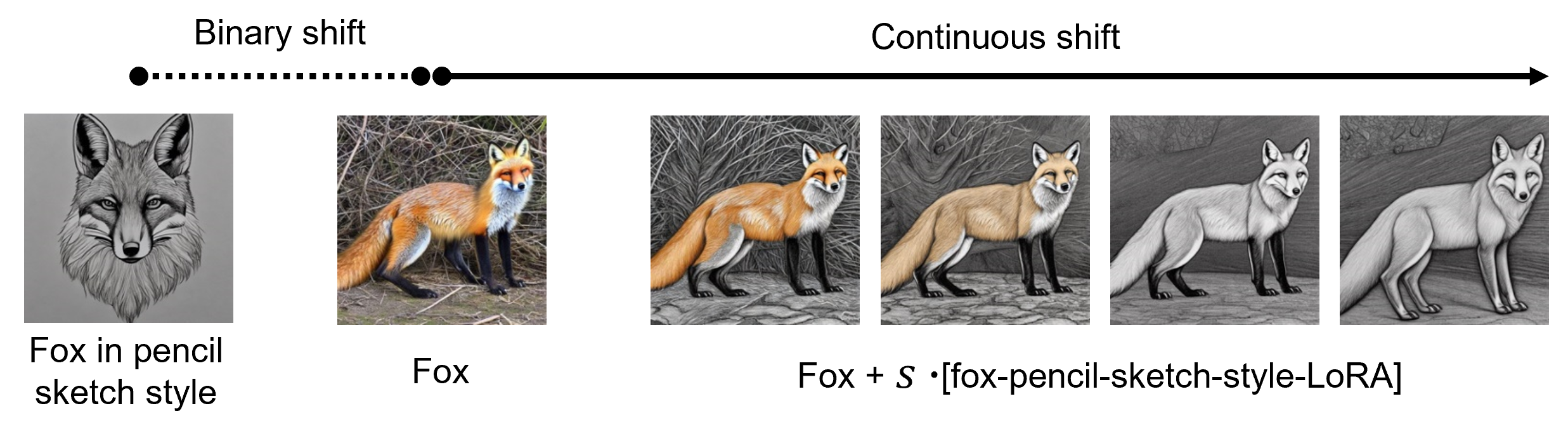}
  \caption{\textbf{Illustration of binary and continuous nuisance shift.} Existing methods using text-to-image diffusion models only enable binary distribution shifts. In contrast, our approach considers gradual and continuous nuisance shifts via weighting of class-specific LoRA sliders. All images are generated using the same diffusion model and seed.
  }
  \label{fig:qualitative_binary_vs_lora_sliding}
  \vspace{-15pt}
\end{figure}

\section{Continuous Nuisance Shift Benchmark}
In this section, we present how our CNS-Bench is created.
First, we discuss how to close the distribution gap between diffusion-generated images and ImageNet in \cref{sec:in_star}.
Then, we introduce how to enable continuous shifts to evaluate the model's sensitivity to various nuisance factors in \cref{sec:cns_bench} and further define the concept of failure points in \cref{sec:failure_point}.
Finally, we detail our filtering dataset and the proposed filtering strategy in \cref{sec:bench_data}.

\subsection{Replicating the ImageNet Distribution} \label{sec:in_star}
We aim to evaluate a model's robustness to specific nuisance shifts that alter the base ImageNet \citep{deng2009imagenet} distribution $p(X_\text{IN}|c)$, conditioned on an ImageNet class $c$.
However, as pointed out by \citet{vendrowDatasetInterfacesDiagnosing2023,kim2024datadream}, the distribution of Stable Diffusion (SD) \citep{rombach2022high} generated images $p(X_\text{SD}|c)$ differs from the ImageNet distribution, significantly lowering classification accuracies. % when evaluating on the diffusion-generated images.
To generate images that are more similar to the ImageNet images, we apply textual inversion~\cite{galimage_2023} to learn new ``words'' in the embedding space of a text encoder that capture the ImageNet-specific class concepts. 
Specifically, these text embeddings are optimized for all ImageNet images by minimizing the noise prediction error of diffusion models  $||\epsilon-\epsilon_{\psi}(\cdot,f_\psi(c)||^2$ with the text encoder $f_\psi(\cdot)$ and parameters $\psi$ for all diffusion time steps.
Following \citep{vendrowDatasetInterfacesDiagnosing2023}, we call this distribution IN*: $p(X|c)=p(X_\text{IN*}|c)$. 

\subsection{Continuous Nuisance Shifts for Benchmarking}
\label{sec:cns_bench}
To evaluate the robustness of image classifiers \wrt continuous nuisance shifts, the following characteristics are desirable: (\romannumeral1) the shift severity should be controllable, (\romannumeral2) the nuisance shift application should not alter the class-specific properties of an object, and (\romannumeral3) the variations should not drastically change the object shape.

A natural way to perform synthetic nuisance shifts is to use methods based on text prompts~\cite{metzen_identification_2023,liu_instruct2attack_2023,vendrowDatasetInterfacesDiagnosing2023}.
They follow the two prompt (2P) templates: ``\verb|A picture of a <class>|'' and ``\verb|A picture of a <class> in <shift>|''.
However, this approach does not allow for the gradual increase of a nuisance for a given image.
Additionally, the semantic structure of the generated image can be significantly changed, as shown in \cref{fig:qualitative_binary_vs_lora_sliding}.

To perform continuous shifts, we leverage LoRA \citep{hu_lora_2021} adapters that represent low-rank matrices added to the original weight matrices.
Such adapters are trained to capture the effect of a considered nuisance shift. 
\citet{gandikota_concept_2023} propose a strategy to learn such concept sliders using LoRA adapters that allow a continuous modulation of the considered concept, which is achieved by learning low-rank matrices that increase the expression of a specific attribute when applied to a class concept $c$. 
The low-rank parameters $\theta_\text{LoRA}$ modify the original model parameters $\theta$ to $\theta^*=\theta+s\cdot \theta_\text{LoRA}$ with scale $s$ and are trained to capture a concept $c_+$: 
\begin{equation}
    p_{\theta^*}(X|c)\leftarrow p_{\theta}(X|c) \cdot p_{\theta}(X|c_+)^{\eta},
\end{equation}
where $\eta$ refers to a weighting factor that is fixed during training.
Following \cite{gandikota_concept_2023}, we optimize with the MSE objective \citep{sohl_2015_thermo} using the Tweedie's formula \citep{efron2011tweedie} and the reparametrization trick \citep{ho_denoising_diffusion} by formulating the scores as a denoising prediction $\epsilon(X,c,t)$ with diffusion timestep $t$: 
\begin{equation}
\text{MSE}(\epsilon_{\theta^*}(X,c,t); \epsilon_{\theta}(X,c,t)+\epsilon_{\theta}(X,c_+,t)).
\end{equation}
We model the class concept $c$ and the nuisance concept $c_+$ by two text embeddings ``\verb|<class>|'' and ``\verb|<class> in <shift>|''.
Different from \citep{gandikota_concept_2023}, we specifically perform distribution shifts for ImageNet classes captured by the IN* distribution.
For this purpose, we introduce ImageNet class-conditional concept sliders $p(X|c,s)$ that allow capturing the class-specific characteristics and confounders of the considered shifts that occur in the real world.
Hence, we train separate LoRA adapters for each ImageNet class and shift.

After training, the learned LoRA adapters capture the direction between the two language concepts, \ie, characterizing attributes of the concept of interest $c_+$. 
The effect of the applied shift is modulated by changing the scale $s$. 
As shown in \cref{fig:qualitative_binary_vs_lora_sliding}, applying these learned directions enables gradual nuisance shifts. 
More examples are provided in \cref{fig:shifts_1} and \cref{fig:shifts_2} in the supplementary.

Activating the LoRA adapter at different timesteps throughout the diffusion process will modulate the effect of the adapter on the generation process \citep{meng2021sdedit}.
If the LoRA adapter is active for all noise steps, it will significantly influence the semantic structure and appearance of the generated image. Conversely, deactivating the adapter for earlier time steps will preserve the semantic structure.
Since we aim to perform edits that do not heavily change the semantic structure, we deactivate the LoRA adapter for early steps. 
This allows applying edits for which the semantic structure remains similar but the appearance changes (\eg,  \cref{fig:qualitative_binary_vs_lora_sliding}).

\subsection{Failure Point Concept}\label{sec:failure_point}
Applying continuous nuisance shifts also enables the computation of failure points, \ie, the nuisance shift scale at which a model fails for a given clean image for the first time, which adds an additional dimension to evaluate model robustness.
We define a failure point 
\begin{equation}
    s=\min\{S\in\mathbb{R}|f(X(S))\neq c\}
\end{equation}
as the smallest shift scale where a classifier $f(X(s))$ fails to correctly classify an image $X(s)$ with a class $c$ and a scale $s$ of a considered shift. 
See \cref{fig:teaser} for an illustration.
Since we are not only interested in the failure for a single image, we define the  failure point distribution that captures the ratio of failed samples in a dataset for all considered scales.
We compute this distribution via a histogram, where the number of elements in one bin corresponds to the number of wrongly classified images at the corresponding scale.

\subsection{Filtering Dataset and Strategy}
\label{sec:bench_data}

\textbf{Filtering of OOC samples.}
The proposed generation strategy enables the generation of diverse and realistic images $x\sim p(X|\textbf{z})$ that are conditioned on $\textbf{z}$, which contains the considered ImageNet class, the considered nuisance shift, and the desired shift scale. % %$c \in \mathbb{N} \mid 1 \leq c \leq 1000$ and %$s_i\in\mathbb{R}$
However, the generated sample might deviate from the condition $\mathbf{z}$ if the influence of the weighted LoRA adapter is too large, distorting the original class condition.
%While low-likely samples are in general not necessarily desired, long-tail samples also occur in the real world.
For benchmarking applications, we are particularly concerned about generated samples deviating from the original class $c$, \ie, the considered class cannot be characterized anymore, and we call such samples out-of-class (OOC) samples \citep{metzen_identification_2023}. 

To evaluate the sliding process and to benchmark OOC filtering mechanisms, we collect a dataset of generated images of various shift scales.
Details on the labeling strategy and the dataset statistics are provided in \cref{sec:labeling_details}.

\noindent\textbf{OOC filtering strategy.} 
An OOC filter serves its purpose if it removes all OOC samples, \ie, a high true positive rate (TPR), while retaining in-class samples, \ie, a low false positive rate (FPR). 
Since we aim to benchmark ImageNet-trained classifiers, the filtering mechanism should not include ImageNet-trained models to reduce filtering biases. 
Previous methods \citep{prabhu2023lance,vendrowDatasetInterfacesDiagnosing2023} measure whether a concept is still present by computing the alignment of the image to the prompt template $p$ ``\verb|A picture of a <class>|'' using CLIP \citep{radford2021learning}.
Specifically, the text-based alignment is computed via the cosine similarity for an image of scale $k$:
\begin{equation}
\mathcal{A}_\text{text} = \cos\left(\text{CLIP}_{\text{img}}\left(I_{k}\right),\text{CLIP}_{\text{text}}\left(p\right)\right).
\end{equation}
We additionally compute the cosine similarity with respect to the prompt template  ``\verb|A picture of a <class> in <shift>|'' to also measure the class concept in the shifted setting.
However, it has been shown that CLIP captures the training data bias and thus sometimes fails to capture a concept correctly~\citep{wang2024sober}.
Therefore, we furthermore measure class discrepancy of the shifted image with respect to the original image via the cosine similarities of image features $\mathcal{F}_0$ and $\mathcal{F}_k$ at the two scales $0$ and $k$:
\begin{equation}
    \mathcal{A}_\text{feat} = \cos\left(\mathcal{F}_0,\mathcal{F}_k\right).
\end{equation}
Here, in addition to the CLIP image features, we utilize the DINOv2 CLS token since it captures semantic similarity via a purely image-based self-supervised learning objective~\citep{oquab2023dinov2}. 
The final OOC filter is composed of four filters with two filters based on text alignment and two based on image feature similarities and we filter out an image if two out of four filters are active. 
We select the filtering threshold for each filter such that more than 90\% of the OOC images that do not correspond to the original class are removed. 
Note that none of these filters is trained on ImageNet data.

\section{Evaluation of CNS-Bench}
In this section, we present experimental details about the training of the class-specific sliders and the OOC filtering strategy.
Additionally, we apply our benchmarking strategy to the classes and with the weather shifts of the OOD-CV benchmark to compare our distribution shifts to a real-world OOD dataset \cite{zhao_ood-cv_2022}.

\subsection{Distribution Gap to ImageNet}
As pointed out in \cref{sec:in_star}, we use textual inversions to replicate the ImageNet distribution, and we call it IN*.
To evaluate the relevance of this approach, we generate 200 images of 100 randomly selected ImageNet classes using Stable Diffusion with the standard text template ``\verb|A picture of a <class>|'' and with the text embeddings acquired via textual inversions of IN. 
To quantify the distribution gap, we compute the FID to ImageNet of the selected classes and the classification accuracies for an ImageNet-trained ResNet-50 classifier, and we present the results in \cref{tab:ins_accs}. 
The results show that the IN* approach leads to unshifted generated images that are closer to the ImageNet distribution. 
Therefore, we perform all experiments using the IN* distribution.

\begin{figure}
    \centering
    \includegraphics[width=0.9\linewidth]{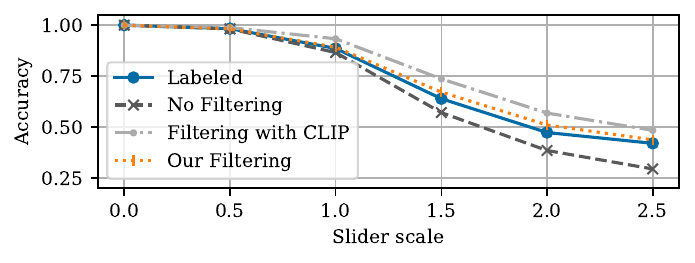}
    \caption{\textbf{The classification accuracy drops on our filtered dataset are closer to the human-filter version.} 
    The accuracy drop curves of a ResNet-50 classifier on our filtered dataset are closer to the accuracy curve of the labeled dataset than the CLIP-filtered variant, supporting the reduced bias of the robustness evaluation and demonstrating the effectiveness of our filtering strategy.
    } 
    \label{fig:accs_filtered_labeled}
\end{figure}

\begin{table}[t]
  \caption{\textbf{IN* distribution and OOC filtering  enhance realism.}} 
  \label{tab:table_cns_eval}
  \vspace{-5pt}
  \centering
  \begin{subtable}[t]{0.48\linewidth}
  \caption{FID to ImageNet and ResNet-50 classification accuracies for generated images from IN* and SD.}\label{tab:ins_accs}
    \centering
    \begin{footnotesize}
    \begin{tabular}{lcc} 
      \toprule 
               & FID($\cdot$,IN)  & RN50 acc.  \\ 
      \midrule 
      SD    & 33.8 & 0.68 \\
      IN*    & 27.1 & 0.74 \\
      \bottomrule
    \end{tabular}
    \end{footnotesize}
    
  \end{subtable}%
  \hfill
  \begin{subtable}[t]{0.48\linewidth}
  \caption{OOC filtering results for CLIP-based filtering and our filtering strategy.}\label{tab:ooc_filtering}
    \centering
    \begin{footnotesize}
    \begin{tabular}{lccc} 
      \toprule 
               & TPR & FPR & Acc  \\ 
      \midrule 
      CLIP   & 0.90  & 0.36 & 0.65\\
      Ours   & 0.88  & 0.12 & 0.88\\
      \bottomrule
    \end{tabular}
    \end{footnotesize}
    
  \end{subtable}
  
\end{table}

\subsection{OOC Filtering Strategy}

We evaluate our proposed filtering mechanism on our manually labeled dataset, and we present the results in \cref{tab:ooc_filtering}.
While our filter removes a similar number of out-of-class images as the CLIP-based approach (TPR), it removes significantly fewer hard samples (lower FPR), resulting in a higher filter accuracy. 
\cref{fig:accs_filtered_labeled} presents the classification accuracy of an ImageNet-trained ResNet-50 classifier for the labeled, the filtered, and the non-filtered versions. 
We observe comparable accuracy drops on both the manually labeled and the datasets filtered by our filter. 
At the same time, the CLIP-based filtering removes more hard samples, resulting in a smaller accuracy drop.
Since the unfiltered version contains failure cases, the classification accuracies are significantly lower.
To further support the realism of our generated images, we fine-tune a ResNet-50 classifier with our data and show more than 10\% gains on ImageNet-R (see \cref{sec:finetuning_inr}). 
We also conducted a user study to evaluate whether our filtered dataset contains images that do not represent the class, which showed that the benchmark contains 1\% of out-of-class samples. 
We refer to the supplementary for further details.

\begin{figure}[t]
    \centering
    \includegraphics[width=\linewidth]{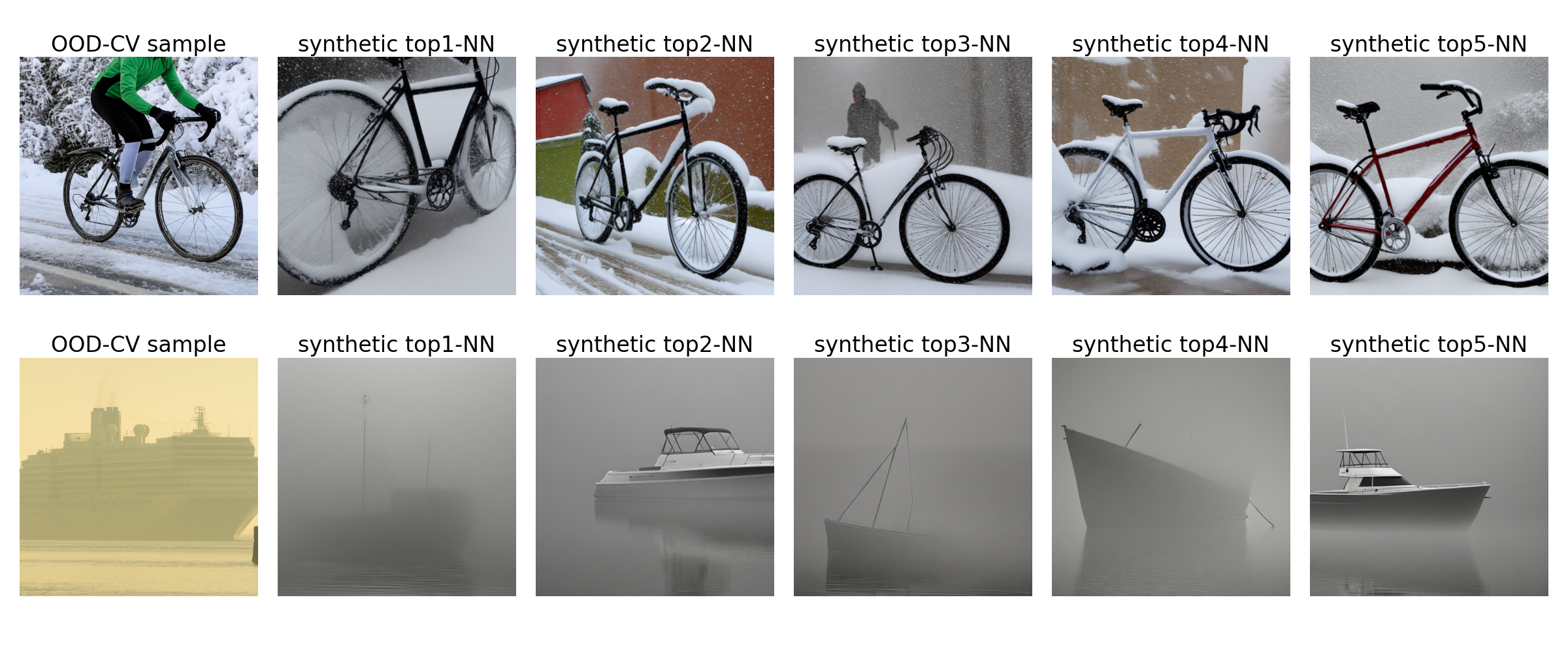}
    \vspace{-5pt}
    \caption{\textbf{Generated OOD images resemble real OOD-CV images.} 
    We find the top-5 nearest neighbors to two example OOD-CV \cite{zhao_ood-cv_2022} images from our benchmark using cosine similarity with CLIP image embedding, illustrating that the benchmark contains images with realistic distribution shifts.}
    \label{fig:nn_oodcv_main}
    \vspace{-5pt}
\end{figure}

\subsection{Comparing Shift Realism with OOD-CV}
\citet{zhao_ood-cv_2022,zhao_ood-cv-v2_2023} introduce OOD-CV to measure out-of-distribution (OOD) robustness of computer vision (CV) models, a benchmark dataset that includes OOD examples of ten object categories for five different individual nuisance factors (\eg, weather) on real data.
OOD-CV is the only real-world dataset that provides accurate labels of various individual weather shifts. 
This allows us to compare our generated images with real-world weather realizations of the considered shifts. 
We use our trained LoRA adapters to create a benchmark for the OOD-CV classes and scales up to $3.0$ to directly compare with the original manually labeled dataset.  
As shown in \cref{fig:nn_oodcv_main}, our generated shifted images resemble exemplary OOD-CV samples.  Additional examples are provided in the supplementary.

\begin{figure}
  \centering
  \includegraphics[width=1.\columnwidth]{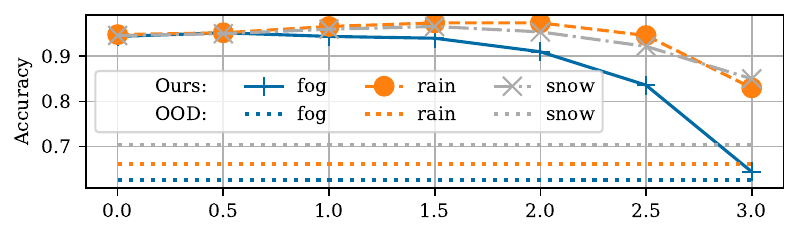}
  \caption{
  \textbf{The classification accuracy degrades gradually for different scales but remains higher than for OOD-CV.} 
  We report the accuracies of a ResNet-50 classifier on OOD-CV (horizontal lines) and our benchmark for multiple scales.
  While the OOD-CV data only allows reporting one OOD accuracy, our benchmark enables the analysis for gradually increasing weather nuisances.
  The accuracy remains higher than for the OOD-CV dataset, indicating the presence of other strong nuisance factors in the OOD-CV dataset.
  }
  \label{fig:exps_oodcv}
  \vspace{-5pt}
\end{figure}

Furthermore, we aim to compare the classifier performance on the OOD-CV benchmark and on our generated images. 
For this purpose, we train a ResNet-50 classifier on the training set of the OOD-CV benchmark.
Then, we evaluate the performance of our data and the OOD-CV benchmark. \cref{fig:exps_oodcv} presents the results for each nuisance independently.
The accuracies remain more or less constant with an accuracy around $95\%$ up to a nuisance scale of $1.5$.
This means that the classifier is not impacted by slight modulations of the image, \eg, some parts of the surroundings covered in snow.
However, from a nuisance scale of $2.0$, the accuracy starts dropping, with the nuisance of \textit{fog} having the biggest impact.
This could be explained by the fact that fog can lead to severe occlusion, while rain and snow can be considered as corruption factors. 
We hypothesize that the larger accuracy drop for the OOD-CV benchmark is due to a significant limitation of its dataset: 
The nuisances are not completely disentangled, and part of the accuracy drop originates from various other factors (\eg, image quality, image size, and noise), as we show in the supplementary (\cref{fig:sup:ood_cv_train_examples}). 
In contrast, our benchmark allows for fine-grained control of nuisances with multiple shift levels, leading to a more disentangled and scalable analysis of model robustness.

\section{Large-Scale Study}

In this section, we first detail the experimental setup and the evaluated models for benchmarking. We then perform a large-scale study on our CNS-Bench.

\subsection{Choices for Generation of Images}
For the generation of images, we use SD2.0, and we activate the LoRA adapters with the selected scale for the last 75\% of the noise steps.
Due to the computational complexity, we consider 100 ImageNet classes. 
To get an estimate of the robustness on the full scale of ImageNet, we classify based on 1000 classes using off-the-shelf classifiers without applying classifier masking, as done by \citet{hendrycks_many_2021}.
We ablate how the number of classes influences the robustness evaluations in \cref{sec:ablation_img_gen}.

\subsection{Evaluated Models and Experimental Setup} 
We use our large-scale benchmark to evaluate models along the following axes:\\
(\romannumeral1) \textit{Architecture.} 
To compare architectures with a comparable number of parameters, we consider both CNN and ViT architectures with different training recipes: ResNet-152 \citep{he2016deep}, ViT-B/16 \citep{dosovitskiy2020image}, DeiT-3-B/16 \citep{touvron2022deit}, and ConvNeXt-B \citep{liu2022convnet}. 
Besides, we also compare the VMamba~\citep{liu2024vmamba} architecture.
All models are trained in a supervised manner.\\
(\romannumeral2) \textit{Model size.} For ViT, we consider the small, medium, base, large, and huge variants of DeiT-3~\cite{touvron2022deit}. 
For CNN, we consider the ResNet~\cite{he2016deep} variants: 18, 34, 50, 101, and 152.\\
(\romannumeral3) \textit{Pre-training paradigm and data.}
We evaluate a set of models with the same backbone but different pre-training paradigms, including both supervised~\cite{dosovitskiy2020image,touvron2021training,touvron2022deit} and self-supervised~\cite{chen2020simple,he2020momentum,zhan2020online,chen2021empirical,caron2021emerging,xie2021unsupervised,xie2022delving,he2022masked,xie2023masked,li2023correlational} pre-training.
Specifically, the following models are pre-trained on IN1k with a self-supervised objective: 
MAE \citep{he2022masked}, DINOv1 \citep{caron2021emerging}, and MoCov3 \citep{chen2021empirical}. % \textcolor{red}{and VMAMBA ()}.
We compare these pre-training strategies to a model that was pre-trained using more data on ImageNet-21k in a supervised manner. 
All transformer-based models use ViT-B/16 as the backbone.
Furthermore, we evaluate an ImageNet-trained diffusion classifier \citep{li2023diffusion} on a smaller subset due to its heavy computational cost.

\noindent\textbf{Metrics.} 
We report the average accuracy drops, i.e., the ratio of failed images, averaged over the images of one shift or all shifts in the value range $[ 0,1]$.
In \cref{table:rce_avg_axes}, we report the mean relative corruption error (rCE) as introduced by \cite{hendrycks_benchmarking_2019} with respect to AlexNet~\citep{krizhevsky2012imagenet}.
It is defined by the average over all relative corruption errors for a given shift 
\begin{equation}
    \text{CE}_\text{shift}=\frac{\sum_s  E^f_{\text{shift},s} - E^f_{\text{shift},0}}{\sum_s  E^\text{alex}_{\text{shift},s} - E^\text{alex}_{\text{shift},0}}
\end{equation}
with the average error $E$ for scale $s$, and model $f$.

\noindent\textbf{Selection of nuisance shifts.}
The selection of the shifts is mainly inspired by ImageNet-R \cite{hendrycks_many_2021} (8 shifts) and the OOD-CV dataset \cite{zhao_ood-cv_2022} (6 shifts) to consider a diverse set of nuisance shifts that modulate the appearance and style or the background and occlusion.
Specifically, we consider the following 14 shifts: cartoon style, plush toy style, pencil sketch style, painting style, design of sculpture, graffiti style, video game renditions style, style of a tattoo, heavy snow, heavy rain, heavy fog, heavy smog, heavy dust, and heavy sandstorm.

The filtered benchmarking dataset contains $192,168$ images in total, with $32,028$ images per scale.

\begin{figure*}
    \centering
    \includegraphics[width=1.\linewidth]{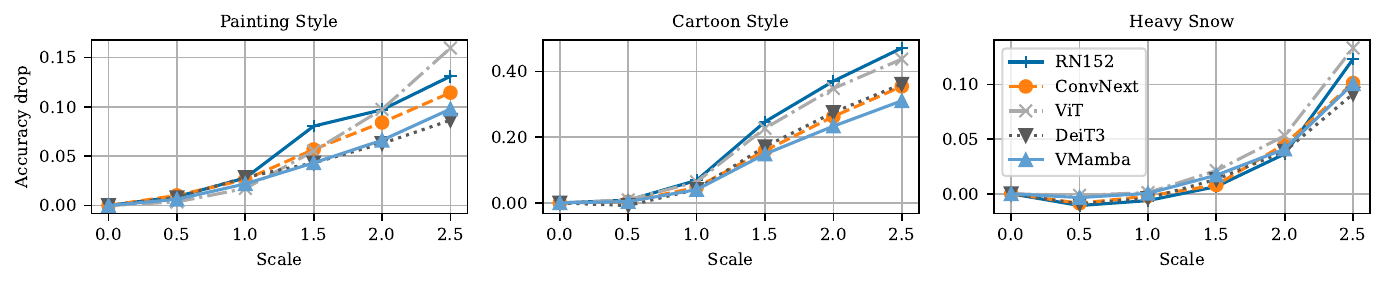}
    \vspace{-15pt}
    \caption{\textbf{Accuracy drops vary for different shifts and scales.} 
    Models exhibit varying performance changes depending on the considered shift.
    Model performances behave differently when increasing the painting style shift (\textit{left}).
    For the cartoon style shift (\textit{center}), the gaps between models increase for larger shift scales, while the accuracy gaps evolve comparably for all models (\textit{right}).
    %all model performances degrade similarly for a gradual increase of the snow shift (\textit{right}), the gaps increase for the cartoon style (\textit{center}) and behave inconsistently for different models when increasing the painting style shift (\textit{left}).
    %For snow and painting shifts, the ranking of the models changes. In contrast, the cartoon style shift results in a consistent model ranking. However, the OOD performance on cartoon-shifted images is drastically worse than the other shifts..
    } 
    \label{fig:acc_drops_individual_shifts}
\end{figure*}

\begin{figure}
    \centering
    \vspace{-5pt}
    \includegraphics[width=1.\linewidth]{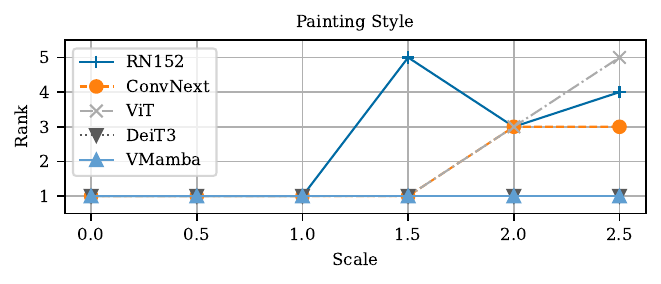}
    \vspace{-15pt}
    \caption{\textbf{Model rankings change for some shifts when increasing the nuisance shift scale.}
    We exemplarily show that model rankings along the painting style shift and the architecture axis change.
    Two models have the same ranking if their one-sigma confidence intervals of the accuracy estimates intersect. 
    %We apply Bonferroni correction \cite{Bonferroni1936} to account for the adjusted confidence interval when ranking multiple models. 
    } 
    \label{fig:rankings_individual_shifts}
    \vspace{-5pt}
\end{figure}

\subsection{Analysis and Findings}
\begin{figure*}
    \centering
    \includegraphics[width=\linewidth]{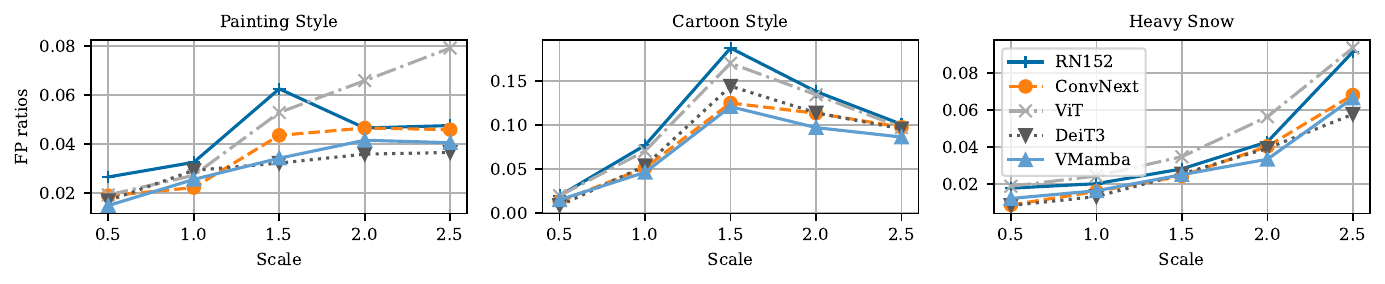}
         \caption{\textbf{Failure points vary for different models and shifts.}
         While the number of failure points gradually increases for the snow shift, most failure points occur around scale $1.5$ for the cartoon-style shift. 
         The failure point distribution clearly varies for different models for the painting style shift.
         We provide results for all shifts in the supplementary (\cref{fig:bps_all_shifts}).
         }
         \label{fig:example_bps}
         \vspace{-5pt}
\end{figure*}

In this subsection, we discuss the main findings of our benchmark. 
Following \cite{hendrycks_benchmarking_2019}, we report the average relative corruption errors as an aggregated measure for the OOD robustness of various models.
%\citet{hendrycks_many_2021}, 
%we report the accuracy drops for 5 scales averaged over 14 diverse shifts as a measure of robustness in \cref{fig:acc_drop_models}. \cref{table:rce_avg_axes} compares models using the average relative corruption errors as proposed by \cite{hendrycks_benchmarking_2019}.
We also provide accuracy drops for various shift scales for three exemplary shifts in \cref{fig:acc_drops_individual_shifts}. 
In addition, we report exemplary failure point distributions in \cref{fig:example_bps}.
We present more evaluations in \cref{sec:more_results}.

\noindent\textbf{Considering multiple scales of a shift allows a more nuanced analysis of OOD robustness.} 
The results in \cref{fig:acc_drops_individual_shifts} demonstrate that the model rankings measured by the accuracy drop change for different scales and shifts.
For example, while the rankings remain consistent for the cartoon style (\textit{right}) for all scales, the model rankings change significantly for the painting style shift: 
Here, ViT outperforms the other models on a lower scale but performs worse on large shift scales.
\cref{fig:rankings_individual_shifts} demonstrates that rankings change significantly.
Varying rankings also occur for other shifts (\cref{fig:acc_drops_all_shifts} in the supplementary). 
We conclude from this observation that the average accuracy drop and the accuracy drops at specific nuisance scales do not always indicate the same model behavior, which provides experimental evidence for the need for a multi-scale robustness benchmarking dataset and adequate metrics.

\noindent\textbf{Model failure points differ across different types of shifts.} 
A failure point is the first scale at which a model fails.
Comparing the failure point distribution of various models reveals significant differences for different shift types, as exemplified in \cref{fig:example_bps}.
We provide more results in the supplementary (\cref{fig:bps_all_shifts}).
Weather shifts, such as snow, typically correspond to slight appearance changes and mainly add a disturbance factor or occlusions to the image.
Therefore, the failure rate increases gradually compared to some style shifts, for which models tend to fail more abruptly at a specific scale, as, \eg, for the cartoon style at scale $s=1.5$.
An exemplary explanation for the abrupt shift in the cartoon shift might be the wrong classification of a class as the ImageNet class \textit{comic book}.

\begin{table}[t]
\caption{\textbf{Model robustness varies along the three considered axes.} 
  We present the average relative corruption error \cite{hendrycks_benchmarking_2019} (lower is better) as a single metric to measure the performance of models along the three explored axes.
We present more results in the  supplementary: 
  Average accuracies over all scales in \cref{fig:acc_drop_models} and the results for all models in \cref{table:mce_and_rCE_all}.
  }
  \centering

\begin{footnotesize}

\begin{tabular}{lr|lr|lr} 
\toprule 
\multicolumn{2}{c|}{Architecture} & \multicolumn{2}{c|}{Size} & \multicolumn{2}{c}{Pre-Training} \\ 
\midrule 
RN152 & 0.790 & DeiT3-S & 0.747 & SUP-IN1k & 0.926 \\
 ConvNeXt & 0.686 & DeiT3-M 
& 0.758 & DINOv1-IN1k & \textbf{0.636} \\ 
ViT & 0.926 & DeiT3-B 
& 0.610 & MAE-IN1k & 0.732 \\ 
DeiT3 & \underline{0.610} 
& DeiT3-L 
& \textbf{0.574} & MoCov3-IN1k & \underline{0.669} \\ 
VMamba& \textbf{0.574}& DeiT3-H & \underline{0.582}& SUP-IN21k-1k& 0.722\\ 
\bottomrule
\end{tabular}

\end{footnotesize}

  \label{table:rce_avg_axes}
  \vspace{-5pt}

\end{table}

\noindent\textbf{Visual state-space models are more robust than transformers and CNNs.}
\cref{table:rce_avg_axes} (\textit{left}) presents the aggregated robustness for classifiers with the same training data and a comparable number of parameters along the architecture axis. 
VMamba outperforms transformers and CNNs on CNS-Bench distribution shifts, although the ImageNet accuracies are comparable. 

\noindent\textbf{Transformers with modern training recipes outperform modern CNNs across all shift severities.}
DeiT3 achieves competing robustness on our benchmark with the VMamba architecture, increasing the gap towards ViT for stronger shifts.
While ResNet-152 is more robust than the standard ViT variant, ConvNeXt still clearly outperforms it. 

A modern CNN (ConvNeXt) outperforms baseline vision transformers (ViT) of a similar size but it is less robust than a transformer with modern training recipes (DeiT3), despite having a higher ID accuracy. 
While the gap between ConvNeXt and DeiT3 does not increase for stronger shifts when averaged over all shifts, we observe that this behavior is not consistent for all shifts. 
Consider, \eg, the failure point distribution in \cref{fig:example_bps} (\textit{Painting Style}), where DeiT3 has a gradually increasing failure point rate, while ConvNeXt depicts a sharp increase for scale $s=1.5$.

\noindent\textbf{Larger models improve the robustness, but this effect is also due to the higher in-distribution accuracy.}
We observe that larger models tend to have a stronger robustness, as shown \cref{table:rce_avg_axes} (\textit{Model size}).
However, larger model counts typically also improve the in-domain accuracy \citep{Miller2021AccuracyOT}, which we further discuss in the supplementary.

\noindent\textbf{Diffusion classifiers are less robust than discriminative models.}
In addition, we also compare the robustness of an ImageNet-trained diffusion classifier \citep{li2023diffusion} on our benchmark.
Due to the high computational cost, we evaluate the accuracy drop of the DiT-based diffusion classifier for 1,000 images on a subset of our dataset (approximately 12,000 images) for the snow and cartoon style shifts.
We apply the L1 loss computation strategy as proposed by \citet{li2023diffusion} since it results in the best performance. 
We compute the average accuracy drops as $0.106$ / $0.07$ / $0.05$ for DiT / supervised ViT / MAE.
Compared with discriminative models evaluated on the same subset, the diffusion classifier demonstrates a lower robustness on the evaluated shifts than the compared discriminative models.
The gap is increasing for larger severity levels (\cref{fig:results_dit} in the supplementary).

\noindent\textbf{More supervised training data improves the robustness, but self-supervised pre-training improves the OOD robustness even stronger.} 
To study the impact of the pre-training paradigm, we compare different learning objectives with the same ViT-B backbone and the same training data and \cref{table:rce_avg_axes} (\textit{right}).
We consider both the supervised and self-supervised (MAE, DINOv1, and MoCov3) paradigms.

First, we observe that more training data benefits OOD robustness: 
Pre-training on IN21k positively impacts the OOD robustness aggregated over all scales compared to a supervised model trained on IN1k. 
This might be explained by the fact that the tested distribution is less OOD for the model~\citep{Miller2021AccuracyOT}. 
However, using a self-supervised objective for pre-training followed by a fine-tuning protocol results in an even better robustness for the same training data and model size. 
Considering the rCE metric in \cref{table:rce_avg_axes} (\textit{right}), the fine-tuned DINOv1 model achieves the best performance.

\section{Conclusion}

The key advantage of using generative models for benchmarking is the ability to perform diverse nuisance shifts in a controlled and scalable way. 
This work filled a gap in generative benchmarking by introducing CNS-Bench, an evaluation method that performs diverse, realistic, fine-grained, and continuous nuisance shifts at multiple scales. 
We studied the necessity of removing out-of-class samples when benchmarking with diffusion-generated images and presented a filter with a higher filtering accuracy.

With the benchmark, we performed a systematic large-scale study of robustness for classifiers along three axes (architecture, number of parameters, pre-training paradigm, and data).
Our study underscored that considering multiple-scale nuisance shifts provides a more nuanced view of the model's robustness, as the performance drops can vary across different nuisance shifts and scales.
Therefore, instead of aggregating the robustness evaluation into a single metric, we encourage the community to report accuracy with different shift scales to foster a more comprehensive understanding of model robustness in various out-of-distribution scenarios.

\noindent\textbf{Limitations and future work.}
While our approach allows for diverse continuous nuisance shifts, it does not eliminate all confounders inherently present due to biases in the training data of CLIP, \ie, failures cannot always be solely attributed to the targeted nuisance concept. 
This highlights an inherent challenge for generative benchmarking approaches, and future advances in generative models could help mitigate these confounders. 
Additionally, while we have carefully addressed this issue in our work, we acknowledge that using generated images can lead to biases arising from the real vs.~synthetic distribution shift.

We hope this benchmark can encourage the community to continue working on more high-quality generative benchmarks and to adopt generated images as an additional source for systematically evaluating the robustness of vision models in a scalable and flexible manner.

\section*{Acknowledgments}
AK acknowledges support via his Emmy Noether Research Group funded by the German Research Foundation (DFG) under Grant No. 468670075.

{
    \small
    \bibliographystyle{ieeenat_fullname}
    \bibliography{main}

\begin{thebibliography}{72}
\providecommand{\natexlab}[1]{#1}
\providecommand{\url}[1]{\texttt{#1}}
\expandafter\ifx\csname urlstyle\endcsname\relax
  \providecommand{\doi}[1]{doi: #1}\else
  \providecommand{\doi}{doi: \begingroup \urlstyle{rm}\Url}\fi

\bibitem[Angelopoulos et~al.(2023)Angelopoulos, Bates, Fannjiang, Jordan, and Zrnic]{angelopoulos2023prediction}
Anastasios~N Angelopoulos, Stephen Bates, Clara Fannjiang, Michael~I Jordan, and Teodor Zrnic.
\newblock Prediction-powered inference.
\newblock \emph{Science}, 2023.

\bibitem[Barbu et~al.(2019)Barbu, Mayo, Alverio, Luo, Wang, Gutfreund, Tenenbaum, and Katz]{barbu_objectnet_2019}
Andrei Barbu, David Mayo, Julian Alverio, William Luo, Christopher Wang, Dan Gutfreund, Josh Tenenbaum, and Boris Katz.
\newblock {ObjectNet}: A large-scale bias-controlled dataset for pushing the limits of object recognition models.
\newblock In \emph{NeurIPS}, 2019.

\bibitem[Baumann et~al.(2025)Baumann, Krause, Neumayr, Stracke, Sevi, Hu, and Ommer]{baumann_continuous_2024}
Stefan~Andreas Baumann, Felix Krause, Michael Neumayr, Nick Stracke, Melvin Sevi, Vincent~Tao Hu, and Bj{\"o}rn Ommer.
\newblock Continuous, subject-specific attribute control in t2i models by identifying semantic directions.
\newblock In \emph{CVPR}, 2025.

\bibitem[Bordes et~al.(2024)Bordes, Shekhar, Ibrahim, Bouchacourt, Vincent, and Morcos]{bordes2024pug}
Florian Bordes, Shashank Shekhar, Mark Ibrahim, Diane Bouchacourt, Pascal Vincent, and Ari Morcos.
\newblock Pug: Photorealistic and semantically controllable synthetic data for representation learning.
\newblock In \emph{NeurIPS}, 2024.

\bibitem[Caron et~al.(2021)Caron, Touvron, Misra, J{\'e}gou, Mairal, Bojanowski, and Joulin]{caron2021emerging}
Mathilde Caron, Hugo Touvron, Ishan Misra, Herv{\'e} J{\'e}gou, Julien Mairal, Piotr Bojanowski, and Armand Joulin.
\newblock Emerging properties in self-supervised vision transformers.
\newblock In \emph{ICCV}, 2021.

\bibitem[Chen et~al.(2020)Chen, Kornblith, Norouzi, and Hinton]{chen2020simple}
Ting Chen, Simon Kornblith, Mohammad Norouzi, and Geoffrey Hinton.
\newblock A simple framework for contrastive learning of visual representations.
\newblock In \emph{ICML}, 2020.

\bibitem[Chen et~al.(2021)Chen, Xie, and He]{chen2021empirical}
Xinlei Chen, Saining Xie, and Kaiming He.
\newblock An empirical study of training self-supervised vision transformers.
\newblock In \emph{ICCV}, 2021.

\bibitem[Deng et~al.(2009)Deng, Dong, Socher, Li, Li, and Fei-Fei]{deng2009imagenet}
Jia Deng, Wei Dong, Richard Socher, Li-Jia Li, Kai Li, and Li Fei-Fei.
\newblock Imagenet: A large-scale hierarchical image database.
\newblock In \emph{CVPR}, 2009.

\bibitem[Deng(2012)]{deng2012mnist}
Li Deng.
\newblock The mnist database of handwritten digit images for machine learning research.
\newblock In \emph{IEEE Signal Processing Magazine}, 2012.

\bibitem[Dosovitskiy et~al.(2020)Dosovitskiy, Beyer, Kolesnikov, Weissenborn, Zhai, Unterthiner, Dehghani, Minderer, Heigold, Gelly, et~al.]{dosovitskiy2020image}
Alexey Dosovitskiy, Lucas Beyer, Alexander Kolesnikov, Dirk Weissenborn, Xiaohua Zhai, Thomas Unterthiner, Mostafa Dehghani, Matthias Minderer, Georg Heigold, Sylvain Gelly, et~al.
\newblock An image is worth 16x16 words: Transformers for image recognition at scale.
\newblock In \emph{ICLR}, 2020.

\bibitem[Drenkow et~al.(2021)Drenkow, Sani, Shpitser, and Unberath]{drenkow_systematic_2022}
Nathan Drenkow, Numair Sani, Ilya Shpitser, and Mathias Unberath.
\newblock Robustness in deep learning for computer vision: Mind the gap?
\newblock \emph{arXiv preprint arXiv:2112.00639}, 2021.

\bibitem[Dutta and Zisserman(2019)]{dutta2019vgg}
Abhishek Dutta and Andrew Zisserman.
\newblock The {VIA} annotation software for images, audio and video.
\newblock In \emph{MM}, 2019.

\bibitem[Dutta et~al.(2016)Dutta, Gupta, and Zissermann]{dutta2016via}
A. Dutta, A. Gupta, and A. Zissermann.
\newblock {VGG} image annotator ({VIA}).
\newblock http://www.robots.ox.ac.uk/~vgg/software/via/, 2016.

\bibitem[Efron(2011)]{efron2011tweedie}
Bradley Efron.
\newblock Tweedie’s formula and selection bias.
\newblock \emph{Journal of the American Statistical Association}, 2011.

\bibitem[Everingham et~al.(2010)Everingham, Van~Gool, Williams, Winn, and Zisserman]{Everingham10}
Mark Everingham, Luc Van~Gool, Christopher~KI Williams, John Winn, and Andrew Zisserman.
\newblock The pascal visual object classes (voc) challenge.
\newblock \emph{IJCV}, 2010.

\bibitem[Gal et~al.(2023)Gal, Alaluf, Atzmon, Patashnik, Bermano, Chechik, and Cohen-or]{galimage_2023}
Rinon Gal, Yuval Alaluf, Yuval Atzmon, Or Patashnik, Amit~Haim Bermano, Gal Chechik, and Daniel Cohen-or.
\newblock An image is worth one word: Personalizing text-to-image generation using textual inversion.
\newblock In \emph{ICLR}, 2023.

\bibitem[Gandelsman et~al.(2022)Gandelsman, Sun, Chen, and Efros]{gandelsman2022test}
Yossi Gandelsman, Yu Sun, Xinlei Chen, and Alexei Efros.
\newblock Test-time training with masked autoencoders.
\newblock In \emph{NeurIPS}, 2022.

\bibitem[Gandikota et~al.(2024)Gandikota, Materzy{\'n}ska, Zhou, Torralba, and Bau]{gandikota_concept_2023}
Rohit Gandikota, Joanna Materzy{\'n}ska, Tingrui Zhou, Antonio Torralba, and David Bau.
\newblock Concept sliders: Lora adaptors for precise control in diffusion models.
\newblock In \emph{ECCV}, 2024.

\bibitem[Gebru et~al.(2021)Gebru, Morgenstern, Vecchione, Vaughan, Wallach, Iii, and Crawford]{gebru2021datasheets}
Timnit Gebru, Jamie Morgenstern, Briana Vecchione, Jennifer~Wortman Vaughan, Hanna Wallach, Hal~Daum{\'e} Iii, and Kate Crawford.
\newblock Datasheets for datasets.
\newblock \emph{Communications of the ACM}, 2021.

\bibitem[Geirhos et~al.(2018)Geirhos, Rubisch, Michaelis, Bethge, Wichmann, and Brendel]{geirhos2018imagenet}
Robert Geirhos, Patricia Rubisch, Claudio Michaelis, Matthias Bethge, Felix~A Wichmann, and Wieland Brendel.
\newblock Imagenet-trained cnns are biased towards texture; increasing shape bias improves accuracy and robustness.
\newblock In \emph{ICLR}, 2018.

\bibitem[He et~al.(2016)He, Zhang, Ren, and Sun]{he2016deep}
Kaiming He, Xiangyu Zhang, Shaoqing Ren, and Jian Sun.
\newblock Deep residual learning for image recognition.
\newblock In \emph{CVPR}, 2016.

\bibitem[He et~al.(2020)He, Fan, Wu, Xie, and Girshick]{he2020momentum}
Kaiming He, Haoqi Fan, Yuxin Wu, Saining Xie, and Ross Girshick.
\newblock Momentum contrast for unsupervised visual representation learning.
\newblock In \emph{CVPR}, 2020.

\bibitem[He et~al.(2022)He, Chen, Xie, Li, Doll{\'a}r, and Girshick]{he2022masked}
Kaiming He, Xinlei Chen, Saining Xie, Yanghao Li, Piotr Doll{\'a}r, and Ross Girshick.
\newblock Masked autoencoders are scalable vision learners.
\newblock In \emph{CVPR}, 2022.

\bibitem[Hendrycks and Dietterich(2018)]{hendrycks_benchmarking_2019}
Dan Hendrycks and Thomas Dietterich.
\newblock Benchmarking neural network robustness to common corruptions and perturbations.
\newblock In \emph{ICLR}, 2018.

\bibitem[Hendrycks et~al.(2021{\natexlab{a}})Hendrycks, Basart, Mu, Kadavath, Wang, Dorundo, Desai, Zhu, Parajuli, Guo, et~al.]{hendrycks_many_2021}
Dan Hendrycks, Steven Basart, Norman Mu, Saurav Kadavath, Frank Wang, Evan Dorundo, Rahul Desai, Tyler Zhu, Samyak Parajuli, Mike Guo, et~al.
\newblock The many faces of robustness: A critical analysis of out-of-distribution generalization.
\newblock In \emph{ICCV}, 2021{\natexlab{a}}.

\bibitem[Hendrycks et~al.(2021{\natexlab{b}})Hendrycks, Zhao, Basart, Steinhardt, and Song]{hendrycks2021nae}
Dan Hendrycks, Kevin Zhao, Steven Basart, Jacob Steinhardt, and Dawn Song.
\newblock Natural adversarial examples.
\newblock In \emph{CVPR}, 2021{\natexlab{b}}.

\bibitem[Ho et~al.(2020)Ho, Jain, and Abbeel]{ho_denoising_diffusion}
Jonathan Ho, Ajay Jain, and Pieter Abbeel.
\newblock Denoising diffusion probabilistic models.
\newblock In \emph{NeurIPS}, 2020.

\bibitem[Hu et~al.(2022)Hu, Shen, Wallis, Allen-Zhu, Li, Wang, Wang, Chen, et~al.]{hu_lora_2021}
Edward~J Hu, Yelong Shen, Phillip Wallis, Zeyuan Allen-Zhu, Yuanzhi Li, Shean Wang, Lu Wang, Weizhu Chen, et~al.
\newblock Lora: Low-rank adaptation of large language models.
\newblock In \emph{ICLR}, 2022.

\bibitem[Idrissi et~al.(2022)Idrissi, Bouchacourt, Balestriero, Evtimov, Hazirbas, Ballas, Vincent, Drozdzal, Lopez-Paz, and Ibrahim]{idrissi_imagenet-x_2022}
Badr~Youbi Idrissi, Diane Bouchacourt, Randall Balestriero, Ivan Evtimov, Caner Hazirbas, Nicolas Ballas, Pascal Vincent, Michal Drozdzal, David Lopez-Paz, and Mark Ibrahim.
\newblock Imagenet-x: Understanding model mistakes with factor of variation annotations.
\newblock \emph{arXiv preprint arXiv:2211.01866}, 2022.

\bibitem[Kar et~al.(2022)Kar, Yeo, Atanov, and Zamir]{kar_3d_2022}
O{\u{g}}uzhan~Fatih Kar, Teresa Yeo, Andrei Atanov, and Amir Zamir.
\newblock 3d common corruptions and data augmentation.
\newblock In \emph{CVPR}, 2022.

\bibitem[Kim et~al.(2024)Kim, Bader, Alaniz, Schmid, and Akata]{kim2024datadream}
Jae~Myung Kim, Jessica Bader, Stephan Alaniz, Cordelia Schmid, and Zeynep Akata.
\newblock Datadream: Few-shot guided dataset generation.
\newblock In \emph{ECCV}, 2024.

\bibitem[Krizhevsky et~al.(2012)Krizhevsky, Sutskever, and Hinton]{krizhevsky2012imagenet}
Alex Krizhevsky, Ilya Sutskever, and Geoffrey~E Hinton.
\newblock Imagenet classification with deep convolutional neural networks.
\newblock In \emph{NeurIPS}, 2012.

\bibitem[Li et~al.(2023{\natexlab{a}})Li, Prabhudesai, Duggal, Brown, and Pathak]{li2023diffusion}
Alexander~C Li, Mihir Prabhudesai, Shivam Duggal, Ellis Brown, and Deepak Pathak.
\newblock Your diffusion model is secretly a zero-shot classifier.
\newblock In \emph{ICCV}, 2023{\natexlab{a}}.

\bibitem[Li et~al.(2023{\natexlab{b}})Li, Xie, and Loy]{li2023correlational}
Wei Li, Jiahao Xie, and Chen~Change Loy.
\newblock Correlational image modeling for self-supervised visual pre-training.
\newblock In \emph{CVPR}, 2023{\natexlab{b}}.

\bibitem[Li et~al.(2023{\natexlab{c}})Li, Chen, Zhu, Wang, Zhang, and Xue]{li_imagenet-e_2023}
Xiaodan Li, Yuefeng Chen, Yao Zhu, Shuhui Wang, Rong Zhang, and Hui Xue.
\newblock Imagenet-e: Benchmarking neural network robustness via attribute editing.
\newblock In \emph{CVPR}, 2023{\natexlab{c}}.

\bibitem[Lin et~al.(2014)Lin, Maire, Belongie, Hays, Perona, Ramanan, Doll{\'a}r, and Zitnick]{cocodataset}
Tsung-Yi Lin, Michael Maire, Serge Belongie, James Hays, Pietro Perona, Deva Ramanan, Piotr Doll{\'a}r, and C~Lawrence Zitnick.
\newblock Microsoft coco: Common objects in context.
\newblock In \emph{ECCV}, 2014.

\bibitem[Liu et~al.(2023)Liu, Wei, Guo, Yu, Yuille, Feizi, Lau, and Chellappa]{liu_instruct2attack_2023}
Jiang Liu, Chen Wei, Yuxiang Guo, Heng Yu, Alan Yuille, Soheil Feizi, Chun~Pong Lau, and Rama Chellappa.
\newblock Instruct2attack: Language-guided semantic adversarial attacks.
\newblock \emph{arXiv preprint arXiv:2311.15551}, 2023.

\bibitem[Liu et~al.(2024)Liu, Tian, Zhao, Yu, Xie, Wang, Ye, and Liu]{liu2024vmamba}
Yue Liu, Yunjie Tian, Yuzhong Zhao, Hongtian Yu, Lingxi Xie, Yaowei Wang, Qixiang Ye, and Yunfan Liu.
\newblock Vmamba: Visual state space model.
\newblock In \emph{NeurIPS}, 2024.

\bibitem[Liu et~al.(2022)Liu, Mao, Wu, Feichtenhofer, Darrell, and Xie]{liu2022convnet}
Zhuang Liu, Hanzi Mao, Chao-Yuan Wu, Christoph Feichtenhofer, Trevor Darrell, and Saining Xie.
\newblock A convnet for the 2020s.
\newblock In \emph{CVPR}, 2022.

\bibitem[Mao et~al.(2022)Mao, Chen, Li, Qi, Duan, Zhang, and Xue]{mao2022easyrobust}
Xiaofeng Mao, Yuefeng Chen, Xiaodan Li, Gege Qi, Ranjie Duan, Rong Zhang, and Hui Xue.
\newblock Easyrobust: A comprehensive and easy-to-use toolkit for robust computer vision.
\newblock \url{https://github.com/alibaba/easyrobust}, 2022.

\bibitem[Meng et~al.(2021)Meng, He, Song, Song, Wu, Zhu, and Ermon]{meng2021sdedit}
Chenlin Meng, Yutong He, Yang Song, Jiaming Song, Jiajun Wu, Jun-Yan Zhu, and Stefano Ermon.
\newblock Sdedit: Guided image synthesis and editing with stochastic differential equations.
\newblock In \emph{ICLR}, 2021.

\bibitem[Metzen et~al.(2023)Metzen, Hutmacher, Hua, Boreiko, and Zhang]{metzen_identification_2023}
Jan~Hendrik Metzen, Robin Hutmacher, N~Grace Hua, Valentyn Boreiko, and Dan Zhang.
\newblock Identification of systematic errors of image classifiers on rare subgroups.
\newblock In \emph{ICCV}, 2023.

\bibitem[Miller et~al.(2021)Miller, Taori, Raghunathan, Sagawa, Koh, Shankar, Liang, Carmon, and Schmidt]{Miller2021AccuracyOT}
John Miller, Rohan Taori, Aditi Raghunathan, Shiori Sagawa, Pang~Wei Koh, Vaishaal Shankar, Percy Liang, Yair Carmon, and Ludwig Schmidt.
\newblock Accuracy on the line: on the strong correlation between out-of-distribution and in-distribution generalization.
\newblock In \emph{ICML}, 2021.

\bibitem[Mofayezi and Medghalchi(2023)]{mofayezi_benchmarking_2023}
Mohammadreza Mofayezi and Yasamin Medghalchi.
\newblock Benchmarking robustness to text-guided corruptions.
\newblock In \emph{CVPRW}, 2023.

\bibitem[Oquab et~al.(2023)Oquab, Darcet, Moutakanni, Vo, Szafraniec, Khalidov, Fernandez, HAZIZA, Massa, El-Nouby, et~al.]{oquab2023dinov2}
Maxime Oquab, Timoth{\'e}e Darcet, Th{\'e}o Moutakanni, Huy~V Vo, Marc Szafraniec, Vasil Khalidov, Pierre Fernandez, Daniel HAZIZA, Francisco Massa, Alaaeldin El-Nouby, et~al.
\newblock Dinov2: Learning robust visual features without supervision.
\newblock \emph{TMLR}, 2023.

\bibitem[Prabhu et~al.(2023)Prabhu, Yenamandra, Chattopadhyay, and Hoffman]{prabhu2023lance}
Viraj Prabhu, Sriram Yenamandra, Prithvijit Chattopadhyay, and Judy Hoffman.
\newblock Lance: Stress-testing visual models by generating language-guided counterfactual images.
\newblock In \emph{NeurIPS}, 2023.

\bibitem[Press et~al.(2023)Press, Schneider, K{\"u}mmerer, and Bethge]{press2023rdumb}
Ori Press, Steffen Schneider, Matthias K{\"u}mmerer, and Matthias Bethge.
\newblock Rdumb: A simple approach that questions our progress in continual test-time adaptation.
\newblock In \emph{NeurIPS}, 2023.

\bibitem[Radford et~al.(2021)Radford, Kim, Hallacy, Ramesh, Goh, Agarwal, Sastry, Askell, Mishkin, Clark, et~al.]{radford2021learning}
Alec Radford, Jong~Wook Kim, Chris Hallacy, Aditya Ramesh, Gabriel Goh, Sandhini Agarwal, Girish Sastry, Amanda Askell, Pamela Mishkin, Jack Clark, et~al.
\newblock Learning transferable visual models from natural language supervision.
\newblock In \emph{ICML}, 2021.

\bibitem[Recht et~al.(2019)Recht, Roelofs, Schmidt, and Shankar]{recht2019imagenet}
Benjamin Recht, Rebecca Roelofs, Ludwig Schmidt, and Vaishaal Shankar.
\newblock Do imagenet classifiers generalize to imagenet?
\newblock In \emph{ICML}, 2019.

\bibitem[Rombach et~al.(2022)Rombach, Blattmann, Lorenz, Esser, and Ommer]{rombach2022high}
Robin Rombach, Andreas Blattmann, Dominik Lorenz, Patrick Esser, and Bj{\"o}rn Ommer.
\newblock High-resolution image synthesis with latent diffusion models.
\newblock In \emph{CVPR}, 2022.

\bibitem[Schneider et~al.(2020)Schneider, Rusak, Eck, Bringmann, Brendel, and Bethge]{schneider2020improving}
Steffen Schneider, Evgenia Rusak, Luisa Eck, Oliver Bringmann, Wieland Brendel, and Matthias Bethge.
\newblock Improving robustness against common corruptions by covariate shift adaptation.
\newblock In \emph{NeurIPS}, 2020.

\bibitem[Shu et~al.(2020)Shu, Liu, Qiu, and Yuille]{shu_identifying_2019}
Michelle Shu, Chenxi Liu, Weichao Qiu, and Alan Yuille.
\newblock Identifying model weakness with adversarial examiner.
\newblock In \emph{AAAI}, 2020.

\bibitem[Sohl-Dickstein et~al.(2015)Sohl-Dickstein, Weiss, Maheswaranathan, and Ganguli]{sohl_2015_thermo}
Jascha Sohl-Dickstein, Eric~A. Weiss, Niru Maheswaranathan, and Surya Ganguli.
\newblock Deep unsupervised learning using nonequilibrium thermodynamics.
\newblock \emph{JMLR}, 2015.

\bibitem[Song et~al.(2021)Song, Meng, and Ermon]{songdenoising_2021}
Jiaming Song, Chenlin Meng, and Stefano Ermon.
\newblock Denoising diffusion implicit models.
\newblock In \emph{ICLR}, 2021.

\bibitem[Sun et~al.(2022)Sun, Segu, Postels, Wang, Van~Gool, Schiele, Tombari, and Yu]{sun_shift_2022}
Tao Sun, Mattia Segu, Janis Postels, Yuxuan Wang, Luc Van~Gool, Bernt Schiele, Federico Tombari, and Fisher Yu.
\newblock {SHIFT}: A synthetic driving dataset for continuous multi-task domain adaptation.
\newblock In \emph{CVPR}, 2022.

\bibitem[Sun et~al.(2020)Sun, Wang, Liu, Miller, Efros, and Hardt]{sun2020test}
Yu Sun, Xiaolong Wang, Zhuang Liu, John Miller, Alexei Efros, and Moritz Hardt.
\newblock Test-time training with self-supervision for generalization under distribution shifts.
\newblock In \emph{ICLR}, 2020.

\bibitem[Taori et~al.(2020)Taori, Dave, Shankar, Carlini, Recht, and Schmidt]{taori_measuring_2020}
Rohan Taori, Achal Dave, Vaishaal Shankar, Nicholas Carlini, Benjamin Recht, and Ludwig Schmidt.
\newblock Measuring robustness to natural distribution shifts in image classification.
\newblock In \emph{NeurIPS}, 2020.

\bibitem[Touvron et~al.(2021)Touvron, Cord, Douze, Massa, Sablayrolles, and J{\'e}gou]{touvron2021training}
Hugo Touvron, Matthieu Cord, Matthijs Douze, Francisco Massa, Alexandre Sablayrolles, and Herv{\'e} J{\'e}gou.
\newblock Training data-efficient image transformers \& distillation through attention.
\newblock In \emph{ICML}, 2021.

\bibitem[Touvron et~al.(2022)Touvron, Cord, and J{\'e}gou]{touvron2022deit}
Hugo Touvron, Matthieu Cord, and Herv{\'e} J{\'e}gou.
\newblock Deit iii: Revenge of the vit.
\newblock In \emph{ECCV}, 2022.

\bibitem[Vendrow et~al.(2023)Vendrow, Jain, Engstrom, and Madry]{vendrowDatasetInterfacesDiagnosing2023}
Joshua Vendrow, Saachi Jain, Logan Engstrom, and Aleksander Madry.
\newblock Dataset interfaces: Diagnosing model failures using controllable counterfactual generation.
\newblock \emph{arXiv preprint arXiv:2302.07865}, 2023.

\bibitem[von Platen et~al.(2022)von Platen, Patil, Lozhkov, Cuenca, Lambert, Rasul, Davaadorj, Nair, Paul, Berman, Xu, Liu, and Wolf]{von-platen-etal-2022-diffusers}
Patrick von Platen, Suraj Patil, Anton Lozhkov, Pedro Cuenca, Nathan Lambert, Kashif Rasul, Mishig Davaadorj, Dhruv Nair, Sayak Paul, William Berman, Yiyi Xu, Steven Liu, and Thomas Wolf.
\newblock Diffusers: State-of-the-art diffusion models.
\newblock \url{https://github.com/huggingface/diffusers}, 2022.

\bibitem[Wang et~al.(2019)Wang, Ge, Lipton, and Xing]{wang_learning_2019}
Haohan Wang, Songwei Ge, Zachary Lipton, and Eric~P Xing.
\newblock Learning robust global representations by penalizing local predictive power.
\newblock In \emph{NeurIPS}, 2019.

\bibitem[Wang et~al.(2024)Wang, Lin, Chen, Schmidt, Han, and Zhang]{wang2024sober}
Qizhou Wang, Yong Lin, Yongqiang Chen, Ludwig Schmidt, Bo Han, and Tong Zhang.
\newblock A sober look at the robustness of clips to spurious features.
\newblock In \emph{NeurIPS}, 2024.

\bibitem[Xie et~al.(2021)Xie, Zhan, Liu, Ong, and Loy]{xie2021unsupervised}
Jiahao Xie, Xiaohang Zhan, Ziwei Liu, Yew~Soon Ong, and Chen~Change Loy.
\newblock Unsupervised object-level representation learning from scene images.
\newblock In \emph{NeurIPS}, 2021.

\bibitem[Xie et~al.(2022)Xie, Zhan, Liu, Ong, and Loy]{xie2022delving}
Jiahao Xie, Xiaohang Zhan, Ziwei Liu, Yew-Soon Ong, and Chen~Change Loy.
\newblock Delving into inter-image invariance for unsupervised visual representations.
\newblock \emph{IJCV}, 2022.

\bibitem[Xie et~al.(2023)Xie, Li, Zhan, Liu, Ong, and Loy]{xie2023masked}
Jiahao Xie, Wei Li, Xiaohang Zhan, Ziwei Liu, Yew~Soon Ong, and Chen~Change Loy.
\newblock Masked frequency modeling for self-supervised visual pre-training.
\newblock In \emph{ICLR}, 2023.

\bibitem[Xie et~al.(2025)Xie, Tonioni, Rauschmayr, Tombari, and Schiele]{xie2025test}
Jiahao Xie, Alessio Tonioni, Nathalie Rauschmayr, Federico Tombari, and Bernt Schiele.
\newblock Test-time visual in-context tuning.
\newblock In \emph{CVPR}, 2025.

\bibitem[Zhan et~al.(2020)Zhan, Xie, Liu, Ong, and Loy]{zhan2020online}
Xiaohang Zhan, Jiahao Xie, Ziwei Liu, Yew-Soon Ong, and Chen~Change Loy.
\newblock Online deep clustering for unsupervised representation learning.
\newblock In \emph{CVPR}, 2020.

\bibitem[Zhang et~al.(2024)Zhang, Pan, Kim, Kweon, and Mao]{zhang_imagenet-d_2024}
Chenshuang Zhang, Fei Pan, Junmo Kim, In~So Kweon, and Chengzhi Mao.
\newblock Imagenet-d: Benchmarking neural network robustness on diffusion synthetic object.
\newblock In \emph{CVPR}, 2024.

\bibitem[Zhao et~al.(2022)Zhao, Yu, Ma, Yu, Mei, Wang, He, Yuille, and Kortylewski]{zhao_ood-cv_2022}
Bingchen Zhao, Shaozuo Yu, Wufei Ma, Mingxin Yu, Shenxiao Mei, Angtian Wang, Ju He, Alan Yuille, and Adam Kortylewski.
\newblock Ood-cv: A benchmark for robustness to out-of-distribution shifts of individual nuisances in natural images.
\newblock In \emph{ECCV}, 2022.

\bibitem[Zhao et~al.(2024)Zhao, Wang, Ma, Jesslen, Yang, Yu, Zendel, Theobalt, Yuille, and Kortylewski]{zhao_ood-cv-v2_2023}
Bingchen Zhao, Jiahao Wang, Wufei Ma, Artur Jesslen, Siwei Yang, Shaozuo Yu, Oliver Zendel, Christian Theobalt, Alan Yuille, and Adam Kortylewski.
\newblock Ood-cv-v2: An extended benchmark for robustness to out-of-distribution shifts of individual nuisances in natural images.
\newblock \emph{TPAMI}, 2024.

\bibitem[Zhao et~al.(2023)Zhao, Bai, Rao, Zhou, and Lu]{zhao2023unipc}
Wenliang Zhao, Lujia Bai, Yongming Rao, Jie Zhou, and Jiwen Lu.
\newblock Unipc: A unified predictor-corrector framework for fast sampling of diffusion models.
\newblock In \emph{NeurIPS}, 2023.

\end{thebibliography}
}

\clearpage
%\input{sec/X_supplementary_wo_cite}

%%%%%%%%%%%%%%%%%%%%%%%%%%%%%%%%%%%%%%%%%%%%%%%%%%%%%%%%%%%%
\clearpage
\setcounter{page}{1}
 {
   \newpage
       \twocolumn[
        \centering
        \Large
        \textbf{\thetitle}\\
        \vspace{0.5em}Supplementary Material \\
        \vspace{1.5em}
       ] %< twocolumn
   }

\appendix

%\section{Appendix} \label{sec:supplementary}
%In the following, we provide supplementary information that is not elaborated in our main paper: We will discuss more details about the benchmarking dataset, the filtering, and image generation strategy. 
%Additionally, we will provide more results.

\section{Benchmark Details}
This section provides more details about the benchmarking dataset.

\subsection{List of Shifts, Classes, and Example Images} \label{sec:list_shifts}
\begin{figure*}[th]
     \centering

\begin{subfigure}[b]{1.\textwidth}
    \centering
    \includegraphics[width=0.85\textwidth]{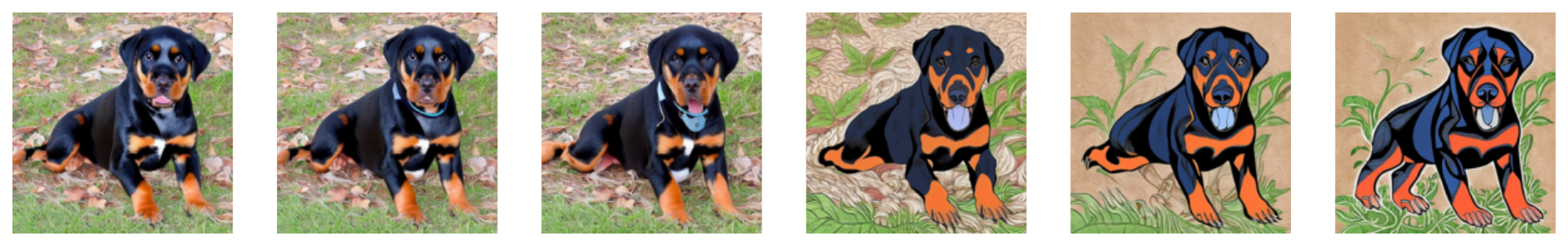}
    \caption{Style of a tattoo.}
    \label{fig:shift_tatoo}
\end{subfigure}

\hfill
\begin{subfigure}[b]{1.\textwidth}
    \centering
    \includegraphics[width=0.85\textwidth]{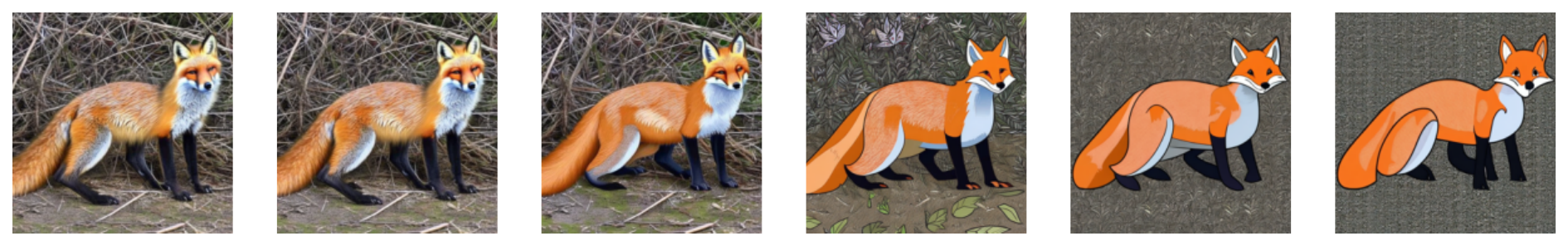}
    \caption{Cartoon style.}
    \label{fig:shift_cartoon}
\end{subfigure}

\hfill
\begin{subfigure}[b]{1.\textwidth}
    \centering
    \includegraphics[width=0.85\textwidth]{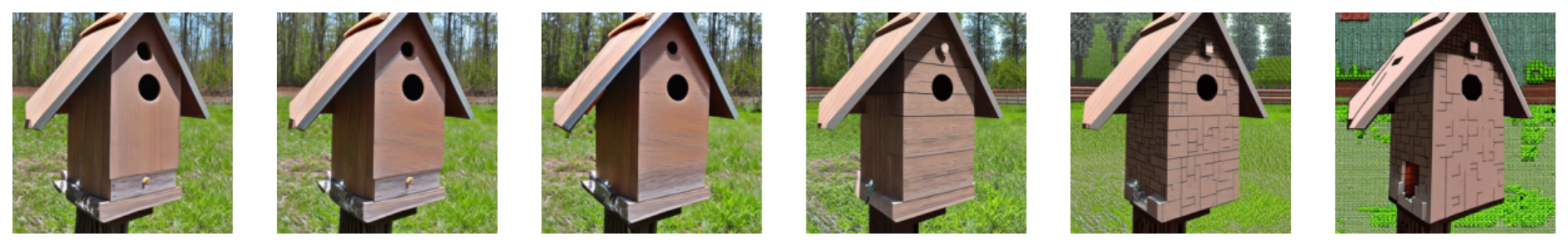}
    \caption{Style of a video game.}
    \label{fig:shift_video_game}
\end{subfigure}

\hfill
\begin{subfigure}[b]{1.\textwidth}
    \centering
    \includegraphics[width=0.85\textwidth]{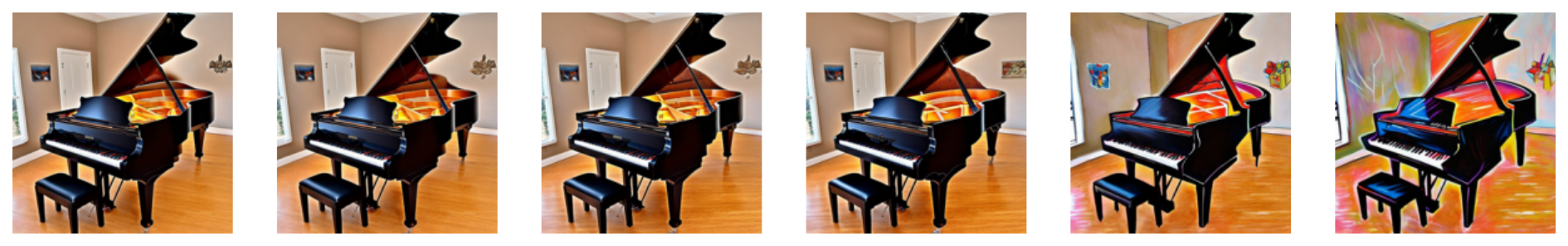}
    \caption{Graffiti style.}
    \label{fig:shift_graffiti}
\end{subfigure}

\hfill
\begin{subfigure}[b]{1.\textwidth}
    \centering
    \includegraphics[width=0.85\textwidth]{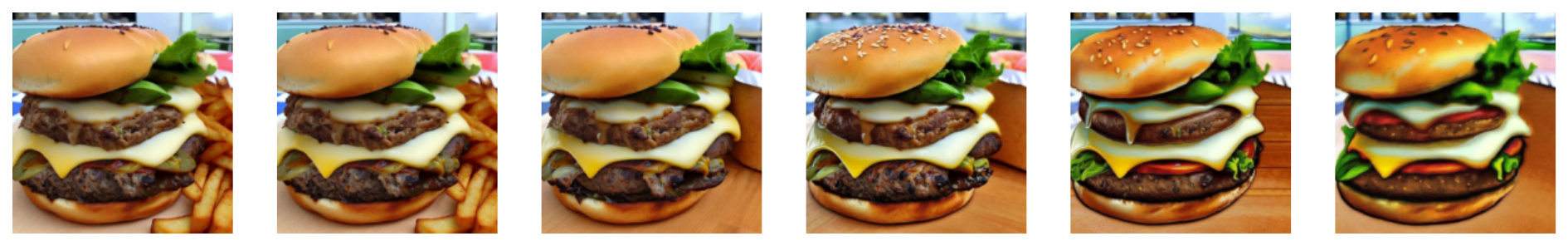}
    \caption{Painting style.}
    \label{fig:shift_painting_style}
\end{subfigure}

\hfill
\begin{subfigure}[b]{1.\textwidth}
    \centering
    \includegraphics[width=0.85\textwidth]{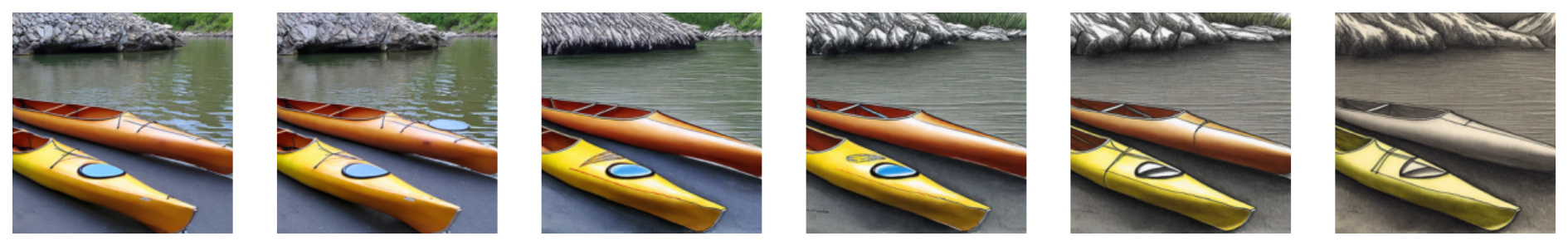}
    \caption{Pencil sketch style.}
    \label{fig:shift_pencil_sketch}
\end{subfigure}

\hfill
\begin{subfigure}[b]{1.\textwidth}
    \centering
    \includegraphics[width=0.85\textwidth]{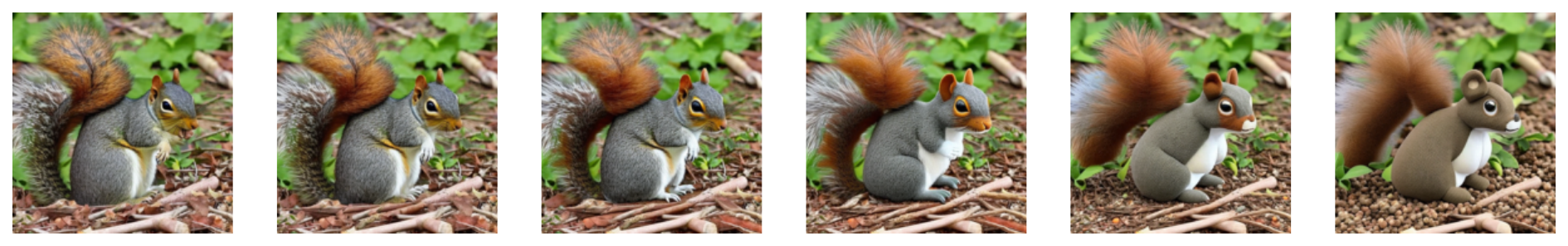}
    \caption{Plush toy style.}
    \label{fig:shift_plush_toy}
\end{subfigure}

\hfill
\begin{subfigure}[b]{1.\textwidth}
    \centering
    \includegraphics[width=0.85\textwidth]{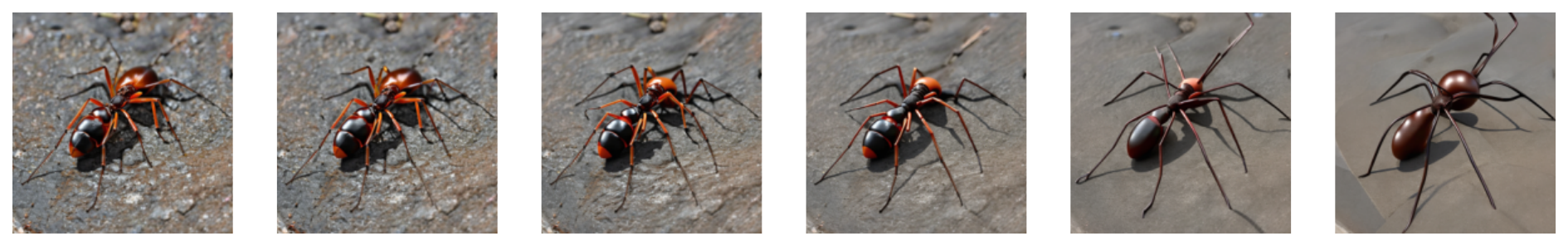}
    \caption{Design of a sculpture.}
    \label{fig:shift_sculpture}
\end{subfigure}

    \caption{\textbf{Example sliding for various nuisance shifts.} We visualize six generated images with the corresponding scales as 0, 0.5, 1, 1.5, 2, and 2.5.}
        \label{fig:shifts_1}
\end{figure*}

\begin{figure*}[th]
     \centering
     \begin{subfigure}[b]{1.\textwidth}
    \centering
    \includegraphics[width=0.85\textwidth]{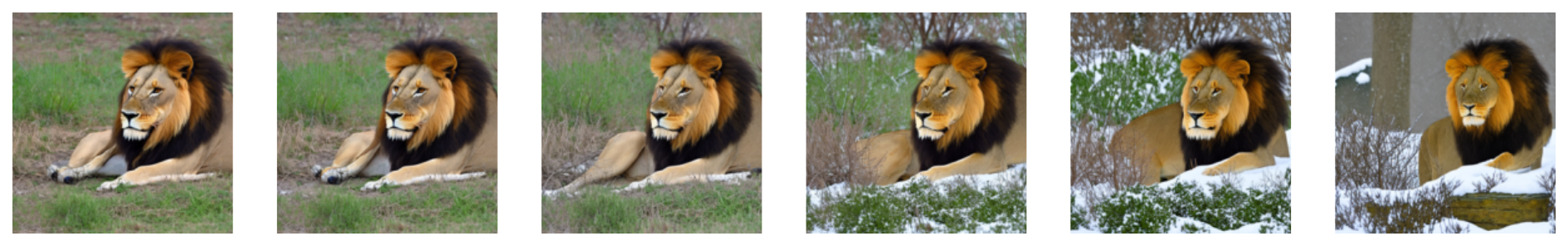}
    \caption{In heavy snow.}
    \label{fig:shift_snow}
\end{subfigure}

\hfill
\begin{subfigure}[b]{1.\textwidth}
    \centering
    \includegraphics[width=0.85\textwidth]{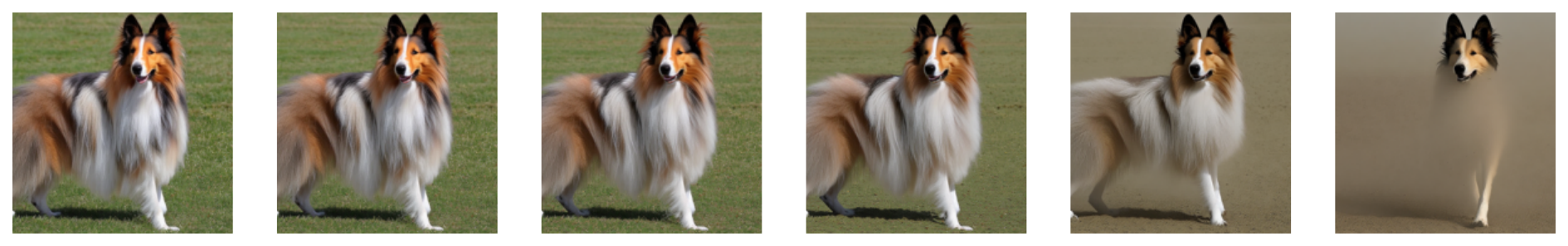}
    \caption{In a sandstorm.}
    \label{fig:shift_sand}
\end{subfigure}
\hfill
\begin{subfigure}[b]{1.\textwidth}
    \centering
    \includegraphics[width=0.85\textwidth]{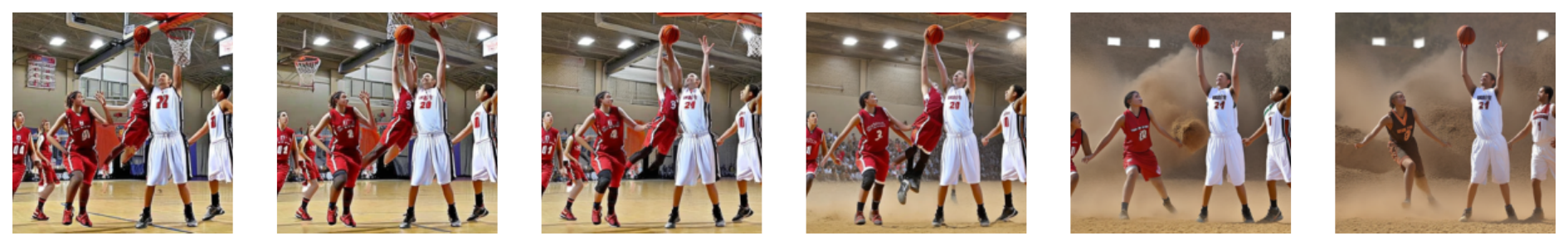}
    \caption{In dust.}
    \label{fig:shift_dust}
\end{subfigure}

\hfill
\begin{subfigure}[b]{1.\textwidth}
    \centering
    \includegraphics[width=0.85\textwidth]{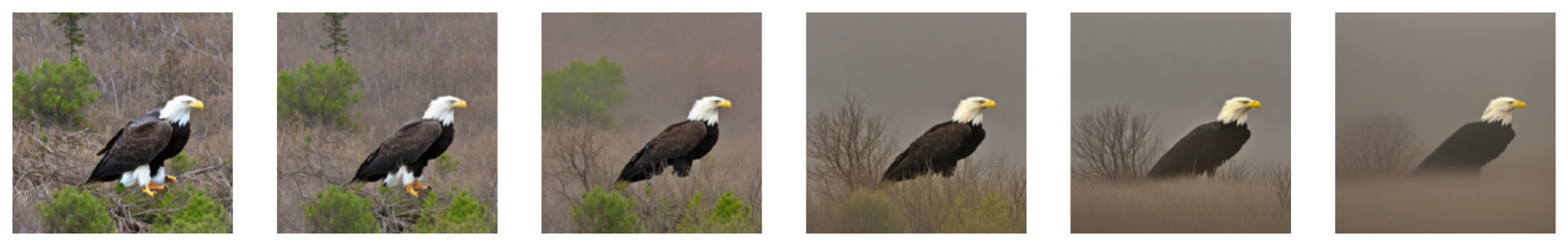}
    \caption{In smog.}
    \label{fig:shift_smog}
\end{subfigure}

\hfill
\begin{subfigure}[b]{1.\textwidth}
    \centering
    \includegraphics[width=0.85\textwidth]{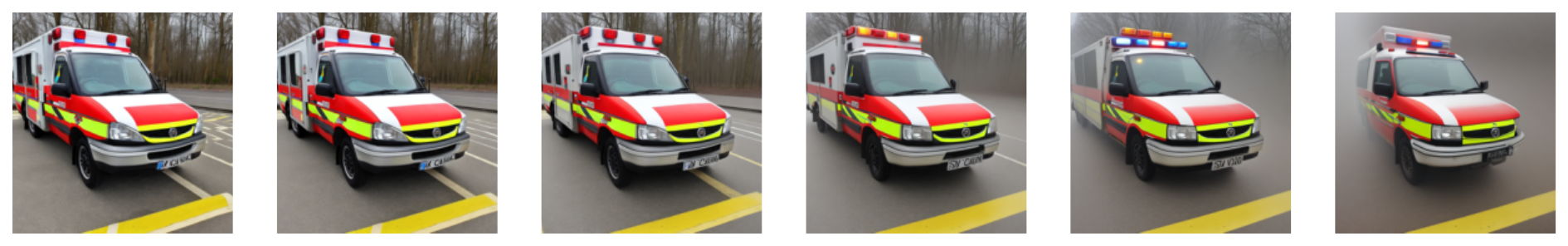}
    \caption{In fog.}
    \label{fig:shift_fog}
\end{subfigure}
\hfill
\begin{subfigure}[b]{1.\textwidth}
    \centering
    \includegraphics[width=0.85\textwidth]{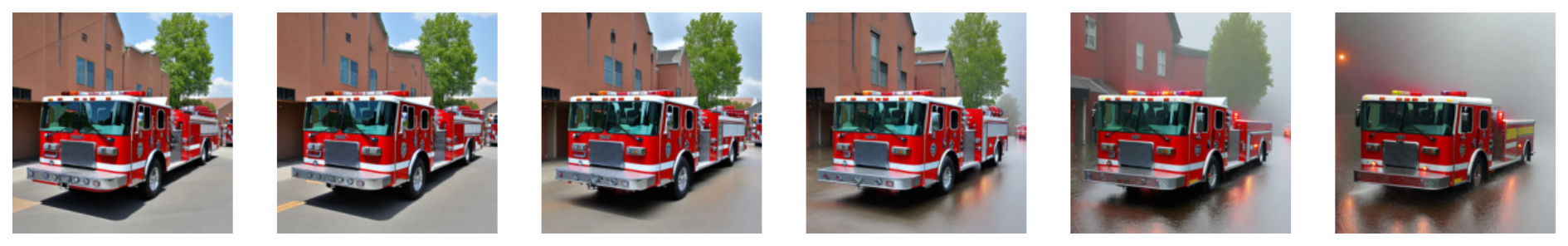}
    \caption{In heavy rain.}
    \label{fig:shift_rain}
\end{subfigure}
    \caption{\textbf{Example sliding for various nuisance shifts.} We visualize six generated images with the corresponding scales as 0, 0.5, 1, 1.5, 2, and 2.5.}
        \label{fig:shifts_2}
\end{figure*}

The results are averaged over the following 14 shifts: \textit{cartoon style, plush toy style, pencil sketch style, painting style, design of sculpture, graffiti style, video game renditions style, style of a tattoo, heavy snow, heavy rain, heavy fog, heavy smog, heavy dust, heavy sandstorm} (see examples in \cref{fig:shifts_1} and \cref{fig:shifts_2}).
We train the sliders using the prompt template ``\verb|A picture of a {class} in {shift}|''. 
Here, we consider the following classes: 
\textit{hammerhead, hen, ostrich, junco, bald eagle, common newt, tree frog, african chameleon, scorpion, centipede, peacock, toucan, goose, koala, jellyfish, hermit crab, pelican, sea lion, afghan hound, bloodhound, italian greyhound, whippet, weimaraner, golden retriever, collie, border collie, rottweiler, french bulldog, s
aint bernard, siberian husky, dalmatian, pug, pembroke, red fox, leopard, snow leopard, lion, ladybug, ant, mantis, starfish, wood rabbit, fox squirrel, beaver, hog, hippopotamus, bison, skunk, gibbon, baboon, giant panda, eel, puffer, accordion, ambulance, basketball, binoculars, birdhouse, bow tie, broom, bucket, cannon, canoe, carousel, cowboy hat, fire engine, flute, gasmask, grand piano, hammer, harp, hatchet, jeep, joystick, lipstick, mailbox, mitten, parachute, pickup, sax, school bus, soccer ball, submarine, tennis ball, warplane, ice cream, bagel, pretzel, cheeseburger, hotdog, head cabbage, broccoli, cucumber, bell pepper, granny smith, lemon, burrito, espresso, volcano, ballplayer}.

\section{More Benchmarking Results}\label{sec:more_results}
\begin{figure*}[t]
         \centering
         \includegraphics[width=1.\linewidth]{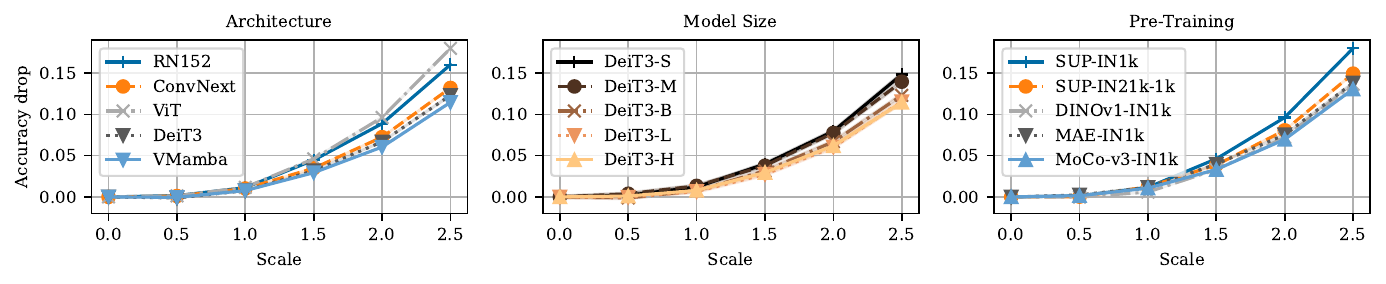}
         \caption{\textbf{Accuracy drops averaged over the whole benchmark.} Architecture (\emph{left}): We show models with the same training data and similar parameter counts. 
         The selection of the architecture influences the accuracy drop. Model size (\emph{center}): We show DeiT3 with various numbers of parameters. Increasing the model capacity results in lower accuracy drops. %\textcolor{red}{Add some ID-OOD plots} %We provide the ResNet results in the supplementary material, which has similar observations. 
         Pre-training paradigm and data (\emph{right}): We show different pre-training paradigms: supervised, self-supervised (MAE, DINO, MoCo), and more data (IN21k), all using ViT-B/16.
         We present results for all shifts in \cref{fig:acc_drops_all_shifts}.}
         \label{fig:acc_drop_models}
 \end{figure*}

\begin{figure}[t]
  \centering
  \includegraphics[width=1.\columnwidth]{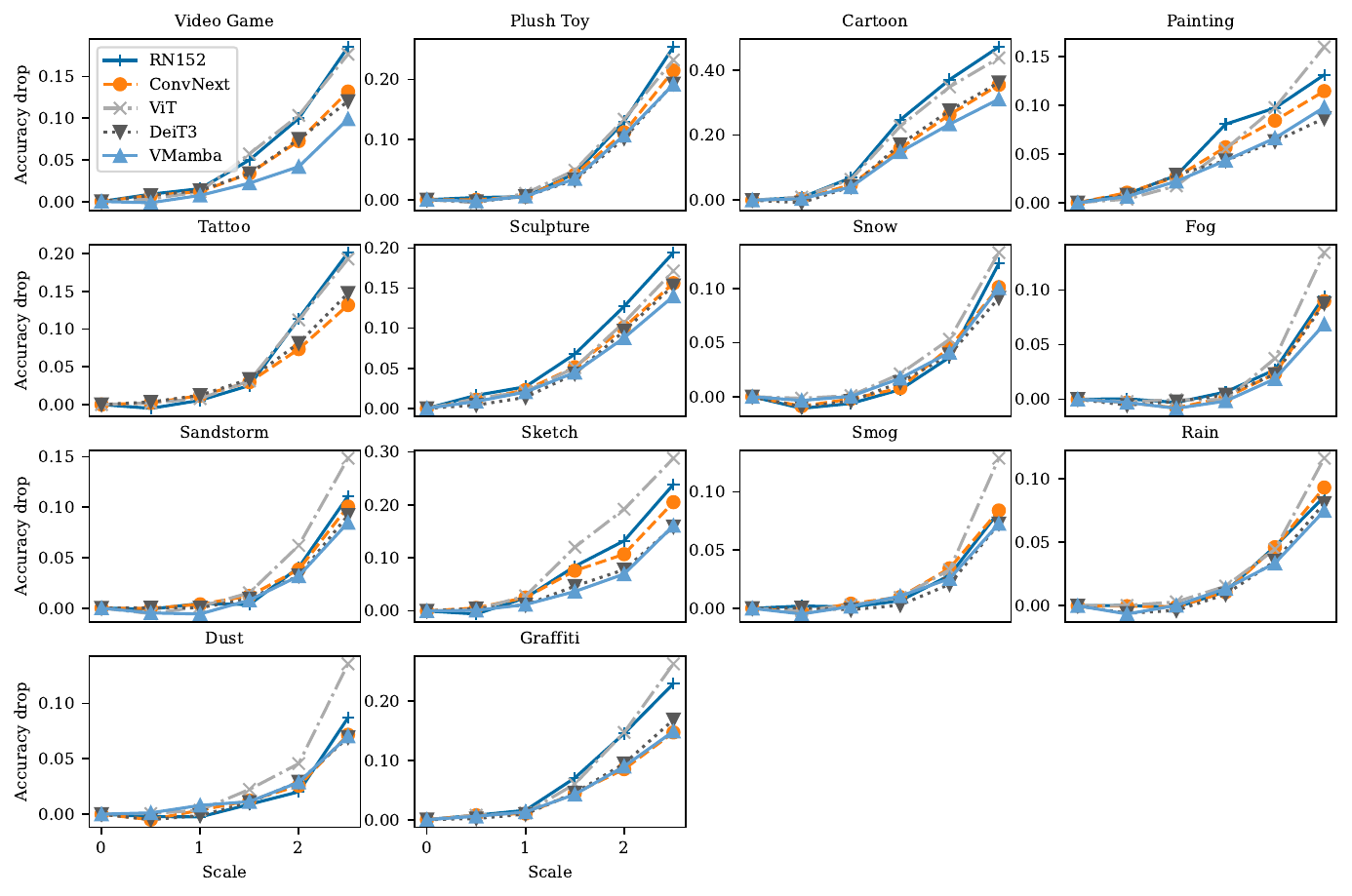}
  \caption{\textbf{Accuracy drops of various architectures for all shifts.} We present the accuracy drops for all shifts in our benchmark. The performance gaps vary for different shifts and scales.}
  \label{fig:acc_drops_all_shifts}
  \vspace{-10pt}
\end{figure}

\begin{figure}
    \centering
    \includegraphics[width=\linewidth]{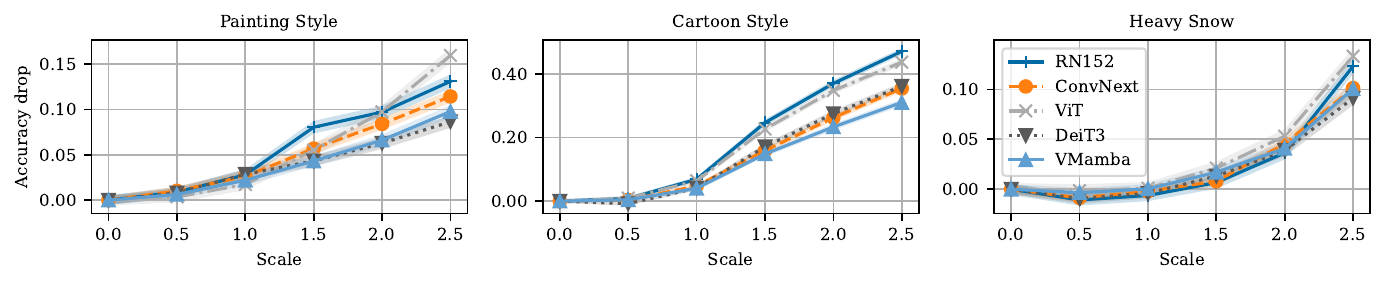}
    \caption{\textbf{Accuracy drops with confidence intervals.} The accuracy drops are depicted for the three shifts along the model axes including the one-sigma confidence interval of the accuracy computation. The results show that some ranking changes are statistically stable.} 
    \label{fig:acc_drops_model_wconf}
    \vspace{-10pt}
\end{figure}

\begin{figure}[th]
    \centering
    \includegraphics[width=0.7\columnwidth]{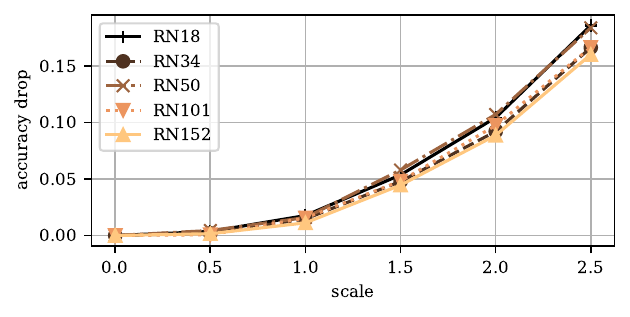}
    \caption{\textbf{Robustness evaluation for ResNet model family.} We compute the accuracy drops for all scales when varying the model size for a set of ResNet models. Larger models result in a better OOD robustness. % but the effect is not as consistent as for DeiT3.
    }
    \label{fig:results_model_size_resnet}
    \vspace{-10pt}
\end{figure}

\begin{figure}[t]
  \centering
  \includegraphics[width=1.\columnwidth]{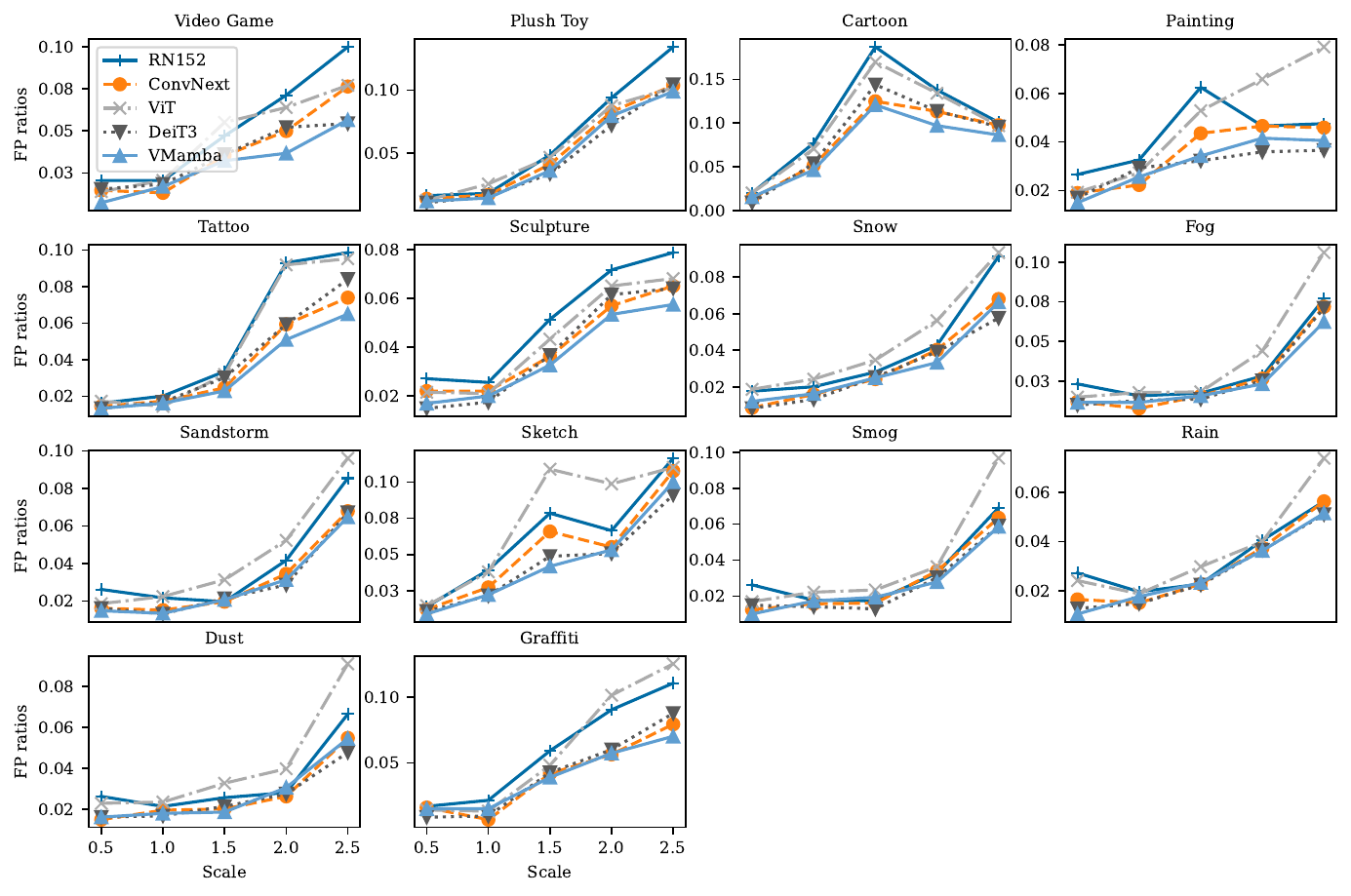}
  \caption{\textbf{Failure point distributions for all shifts.} We present the failure point distributions for all shifts in our benchmark. The failure point distributions vary for different shifts, quantifying the different ways the shifts influence model performance.}
  \label{fig:bps_all_shifts}
\end{figure}

\begin{figure}[t]
  \centering
  \includegraphics[width=\linewidth]{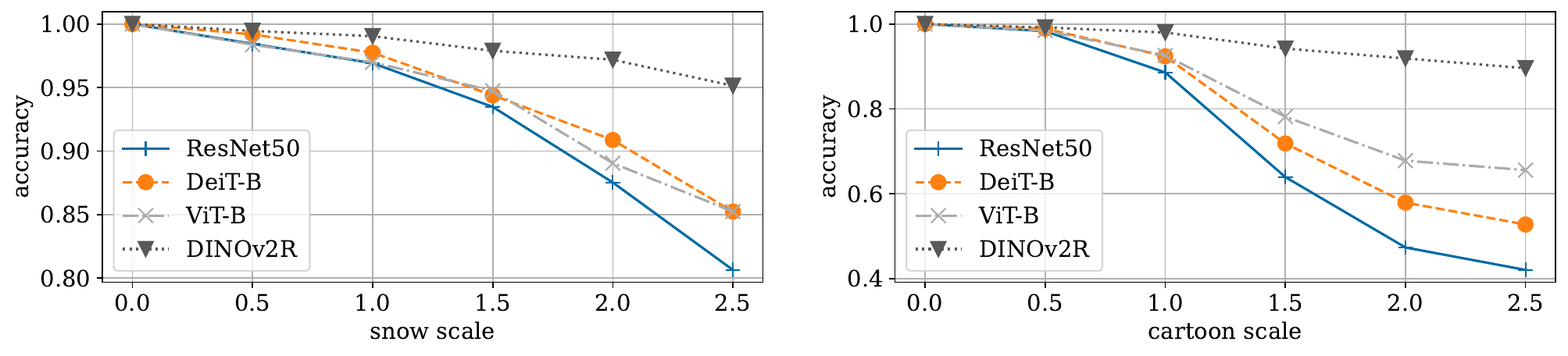}
  \caption{\textbf{Classification accuracy on the labeled dataset for snow and cartoon shifts.} The accuracy drops on the labeled dataset showcase that various classifiers have varying sensitivities on different shifts.}
  \label{fig:acc_bp_labeled_dataset}
  \vspace{-5pt}
\end{figure}

\cref{fig:acc_drop_models} presents the accuracy drops averaged over all shifts and \cref{table:accuracies} lists all average accuracies and accuracy drops for all evaluated models and shift scales.
\cref{fig:acc_drops_model_wconf} plots the accuracy drops for painting, cartoon, and snow shifts with confidence intervals.
As discussed in the main paper, we also provide the accuracy drops for the ResNet family in \cref{fig:results_model_size_resnet}. 
Similar to the observations in \cref{table:rce_avg_axes}, larger models result in a lower accuracy drop in average.
\cref{fig:acc_drops_all_shifts} provides a more nuanced view on the model performances across various architectures on all shifts.
We also plot failure point distributions in \cref{fig:bps_all_shifts}.
\cref{fig:acc_bp_labeled_dataset} presents more classifier results on the labeled dataset.

\begin{table*}[t]
  \caption{\textbf{mCE and mean rCE.} We present the mean corruption error and the mean relative corruption error for all evaluated models.}
  \label{table:mce_and_rCE_all}
  \centering

\begin{tiny}

\begin{tabular}{lrr}
\toprule
 & CE & rCE \\
\midrule
alexnet & 1.000 & 1.000 \\
clip\_resnet101 & 0.532 & 0.563 \\
clip\_resnet50 & 0.715 & 0.587 \\
clip\_vit\_base\_patch16\_224 & 0.420 & 0.230 \\
clip\_vit\_base\_patch32\_224 & 0.487 & 0.591 \\
clip\_vit\_large\_patch14\_224 & 0.445 & 0.228 \\
clip\_vit\_large\_patch14\_336 & 0.419 & 0.274 \\
convnext\_base.fb\_in1k & 0.359 & 0.686 \\
convnext\_large.fb\_in1k & 0.354 & 0.672 \\
convnext\_small.fb\_in1k & 0.353 & 0.609 \\
convnext\_tiny.fb\_in1k & 0.393 & 0.809 \\
convnextv2\_base.fcmae\_ft\_in1k & 0.322 & 0.680 \\
convnextv2\_huge.fcmae\_ft\_in1k & 0.283 & 0.553 \\
convnextv2\_large.fcmae\_ft\_in1k & 0.297 & 0.568 \\
deit3\_base\_patch16\_224.fb\_in1k & 0.396 & 0.610 \\
deit3\_huge\_patch14\_224.fb\_in1k & 0.353 & 0.583 \\
deit3\_large\_patch16\_224.fb\_in1k & 0.382 & 0.574 \\
deit3\_medium\_patch16\_224.fb\_in1k & 0.387 & 0.758 \\
deit3\_small\_patch16\_224.fb\_in1k & 0.400 & 0.747 \\
deit\_base\_patch16\_224.fb\_in1k & 0.437 & 0.746 \\
dino\_vit\_base\_patch16 & 0.504 & 0.851 \\
dinov1\_vit\_base\_patch16 & 0.412 & 0.676 \\
dinov2\_vit\_base\_patch14 & 0.350 & 0.524 \\
dinov2\_vit\_base\_patch14\_reg & 0.311 & 0.456 \\
dinov2\_vit\_giant\_patch14 & 0.321 & 0.431 \\
dinov2\_vit\_giant\_patch14\_reg & 0.311 & 0.426 \\
dinov2\_vit\_large\_patch14 & 0.298 & 0.349 \\
dinov2\_vit\_large\_patch14\_reg & 0.296 & 0.370 \\
dinov2\_vit\_small\_patch14 & 0.351 & 0.639 \\
dinov2\_vit\_small\_patch14\_reg & 0.330 & 0.627 \\
mae\_vit\_base\_patch16 & 0.386 & 0.732 \\
mae\_vit\_huge\_patch14 & 0.303 & 0.542 \\
mae\_vit\_large\_patch16 & 0.328 & 0.571 \\
mocov3\_vit\_base\_patch16 & 0.379 & 0.669 \\
resnet101.a1\_in1k & 0.491 & 0.842 \\
resnet152.a1\_in1k & 0.498 & 0.790 \\
resnet18.a1\_in1k & 0.493 & 0.954 \\
resnet34.a1\_in1k & 0.440 & 0.843 \\
resnet50.a1\_in1k & 0.485 & 0.945 \\
vit\_base\_patch16\_224.augreg\_in1k & 0.569 & 0.926 \\
vit\_base\_patch16\_224.augreg\_in21k\_ft\_in1k & 0.460 & 0.722 \\
vit\_base\_patch16\_clip\_224.openai\_ft\_in1k & 0.282 & 0.482 \\
vssm\_base\_v0 & 0.371 & 0.574 \\
\bottomrule
\end{tabular}

\end{tiny}

\end{table*}

\begin{table*}[t]
  \caption{\textbf{Accuracy evaluations.} We present the accuracies and accuracy drops of all evaluated classifiers.}
  \label{table:accuracies}
  \centering

\begin{tiny}

\begin{tabular}{l|rrrrrrr|rrrrrr}
\hline
\multicolumn{1}{c}{} & \multicolumn{12}{c}{Shift Scale} \\
\cline{2-13}
\multicolumn{1}{c}{} & \multicolumn{7}{c}{Accuracy} & \multicolumn{5}{c}{Accuracy Drop} \\
\hline
 model & 0 & 0.5 & 1 & 1.5 & 2 & 2.5 & avg  & 1 & 1.5 & 2 & 2.5 & avg \\
 \hline
 clip\_resnet50                             & 0.81 & 0.81 & 0.8  & 0.78 & 0.74 & 0.67 & 0.77 &  0.01 & 0.03 & 0.07 & 0.14 & 0.04 \\
 clip\_resnet101                            & 0.86 & 0.86 & 0.85 & 0.83 & 0.81 & 0.74 & 0.82 &  0.01 & 0.03 & 0.06 & 0.12 & 0.04 \\
 clip\_vit\_base\_patch16\_224                 & 0.87 & 0.88 & 0.88 & 0.87 & 0.86 & 0.81 & 0.86 & -0.00    & 0.01 & 0.02 & 0.06 & 0.02 \\
 clip\_vit\_base\_patch32\_224                 & 0.87 & 0.87 & 0.86 & 0.85 & 0.83 & 0.77 & 0.84 &  0.01 & 0.02 & 0.04 & 0.1  & 0.03 \\
 clip\_vit\_large\_patch14\_224                & 0.87 & 0.87 & 0.87 & 0.86 & 0.85 & 0.82 & 0.86 & -0.00    & 0.01 & 0.02 & 0.05 & 0.01 \\
 clip\_vit\_large\_patch14\_336                & 0.88 & 0.88 & 0.88 & 0.87 & 0.86 & 0.83 & 0.87 &  0.00    & 0.01 & 0.02 & 0.05 & 0.01 \\
 convnext\_tiny.fb\_in1k                     & 0.92 & 0.92 & 0.91 & 0.88 & 0.84 & 0.77 & 0.87 &  0.01 & 0.04 & 0.08 & 0.15 & 0.05 \\
convnext\_small.fb\_in1k                    & 0.92 & 0.93 & 0.92 & 0.89 & 0.86 & 0.8  & 0.89 &  0.01 & 0.03 & 0.07 & 0.13 & 0.04 \\
convnext\_base.fb\_in1k                     & 0.93 & 0.93 & 0.92 & 0.89 & 0.85 & 0.79 & 0.89 &  0.01 & 0.03 & 0.07 & 0.13 & 0.04 \\
 convnext\_large.fb\_in1k                    & 0.93 & 0.92 & 0.92 & 0.89 & 0.86 & 0.8  & 0.89 &  0.01 & 0.04 & 0.07 & 0.12 & 0.04 \\
 convnextv2\_base.fcmae\_ft\_in1k             & 0.93 & 0.93 & 0.92 & 0.9  & 0.87 & 0.82 & 0.9  &  0.01 & 0.04 & 0.07 & 0.12 & 0.04 \\
 convnextv2\_large.fcmae\_ft\_in1k            & 0.94 & 0.93 & 0.93 & 0.91 & 0.88 & 0.84 & 0.91 &  0.01 & 0.03 & 0.05 & 0.1  & 0.03 \\
 convnextv2\_huge.fcmae\_ft\_in1k             & 0.94 & 0.93 & 0.93 & 0.91 & 0.89 & 0.84 & 0.91 &  0.01 & 0.03 & 0.05 & 0.09 & 0.03 \\
deit3\_small\_patch16\_224.fb\_in1k           & 0.92 & 0.92 & 0.91 & 0.88 & 0.84 & 0.77 & 0.87 &  0.01 & 0.04 & 0.08 & 0.15 & 0.05 \\
deit3\_base\_patch16\_224.fb\_in1k            & 0.91 & 0.91 & 0.9  & 0.88 & 0.84 & 0.79 & 0.87 &  0.01 & 0.03 & 0.07 & 0.12 & 0.04 \\
 deit3\_medium\_patch16\_224.fb\_in1k          & 0.92 & 0.92 & 0.91 & 0.88 & 0.84 & 0.78 & 0.88 &  0.01 & 0.04 & 0.08 & 0.14 & 0.05 \\
 deit3\_large\_patch16\_224.fb\_in1k           & 0.91 & 0.91 & 0.9  & 0.88 & 0.85 & 0.8  & 0.88 &  0.01 & 0.03 & 0.06 & 0.12 & 0.04 \\
 deit3\_huge\_patch14\_224.fb\_in1k            & 0.92 & 0.92 & 0.91 & 0.89 & 0.86 & 0.81 & 0.89 &  0.01 & 0.03 & 0.06 & 0.11 & 0.04 \\
 deit\_base\_patch16\_224.fb\_in1k             & 0.9  & 0.9  & 0.89 & 0.87 & 0.83 & 0.76 & 0.86 &  0.01 & 0.04 & 0.08 & 0.15 & 0.05 \\
 dino\_lp\_vit\_base\_patch16                     & 0.9  & 0.9  & 0.89 & 0.85 & 0.8  & 0.71 & 0.84 &  0.01 & 0.05 & 0.1  & 0.19 & 0.06 \\
 dinov1\_ft\_vit\_base\_patch16                     & 0.91  & 0.91  & 0.90 & 0.88 & 0.84  & 0.84 & 0.87 &  0.01 & 0.03  & 0.07 & 0.04 & 0.03 \\
 dinov2\_vit\_small\_patch14                  & 0.92 & 0.92 & 0.91 & 0.89 & 0.86 & 0.81 & 0.89 &  0.01 & 0.03 & 0.06 & 0.11 & 0.04 \\
 dinov2\_vit\_small\_patch14\_reg              & 0.93 & 0.93 & 0.92 & 0.9  & 0.87 & 0.81 & 0.89 &  0.01 & 0.03 & 0.06 & 0.11 & 0.04 \\
 dinov2\_vit\_base\_patch14                   & 0.91 & 0.91 & 0.91 & 0.89 & 0.87 & 0.82 & 0.89 &  0.00    & 0.02 & 0.04 & 0.09 & 0.02 \\
 dinov2\_vit\_base\_patch14\_reg               & 0.92 & 0.92 & 0.92 & 0.9  & 0.88 & 0.84 & 0.9  &  0.00    & 0.02 & 0.04 & 0.08 & 0.02 \\
 dinov2\_vit\_large\_patch14                  & 0.92 & 0.92 & 0.92 & 0.91 & 0.89 & 0.86 & 0.9  &  0.00    & 0.01 & 0.03 & 0.06 & 0.02 \\
 dinov2\_vit\_large\_patch14\_reg              & 0.92 & 0.92 & 0.91 & 0.91 & 0.89 & 0.86 & 0.9  &  0.00    & 0.01 & 0.03 & 0.06 & 0.02 \\
 dinov2\_vit\_giant\_patch14                  & 0.91 & 0.91 & 0.91 & 0.9  & 0.88 & 0.84 & 0.89 &  0.00    & 0.01 & 0.04 & 0.07 & 0.02 \\
 dinov2\_vit\_giant\_patch14\_reg              & 0.92 & 0.92 & 0.91 & 0.9  & 0.88 & 0.85 & 0.9  &  0.00    & 0.01 & 0.03 & 0.07 & 0.02 \\
 mae\_vit\_base\_patch16                      & 0.92 & 0.92 & 0.91 & 0.88 & 0.84 & 0.78 & 0.88 &  0.01 & 0.04 & 0.08 & 0.14 & 0.05 \\
 mae\_vit\_huge\_patch14                      & 0.93 & 0.93 & 0.92 & 0.9  & 0.88 & 0.84 & 0.9  &  0.01 & 0.03 & 0.05 & 0.1  & 0.03 \\
 mae\_vit\_large\_patch16                     & 0.93 & 0.92 & 0.92 & 0.9  & 0.87 & 0.83 & 0.9  &  0.01 & 0.03 & 0.05 & 0.1  & 0.03 \\
 mocov3\_vit\_base\_patch16                   & 0.92 & 0.92 & 0.91 & 0.88 & 0.85 & 0.79 & 0.88 &  0.01 & 0.03 & 0.07 & 0.13 & 0.04 \\
 resnet18.a1\_in1k                          & 0.9  & 0.9  & 0.88 & 0.85 & 0.8  & 0.72 & 0.84 &  0.02 & 0.05 & 0.1  & 0.19 & 0.06 \\
 resnet34.a1\_in1k                          & 0.91 & 0.91 & 0.9  & 0.86 & 0.82 & 0.75 & 0.86 &  0.01 & 0.05 & 0.09 & 0.17 & 0.05 \\
 resnet50.a1\_in1k                          & 0.91 & 0.9  & 0.89 & 0.85 & 0.8  & 0.72 & 0.85 &  0.02 & 0.06 & 0.11 & 0.18 & 0.06 \\
 resnet101.a1\_in1k                         & 0.9  & 0.9  & 0.88 & 0.85 & 0.8  & 0.73 & 0.84 &  0.02 & 0.05 & 0.1  & 0.17 & 0.06 \\
 resnet152.a1\_in1k                         & 0.89 & 0.89 & 0.88 & 0.85 & 0.8  & 0.73 & 0.84 &  0.01 & 0.04 & 0.09 & 0.16 & 0.05 \\
 vit\_base\_patch16\_224.augreg\_in1k          & 0.87 & 0.87 & 0.86 & 0.82 & 0.77 & 0.69 & 0.81 &  0.01 & 0.05 & 0.1  & 0.18 & 0.06 \\
 vit\_base\_patch16\_224.augreg\_in21k\_ft\_in1k & 0.9  & 0.9  & 0.89 & 0.86 & 0.82 & 0.75 & 0.85 &  0.01 & 0.04 & 0.08 & 0.15 & 0.05 \\
 vit\_base\_patch16\_clip\_224.openai\_ft\_in1k  & 0.93 & 0.93 & 0.92 & 0.91 & 0.89 & 0.86 & 0.91 &  0.01 & 0.02 & 0.04 & 0.08 & 0.03 \\
 vssm\_base\_v0                                 & 0.91 & 0.91 & 0.91 & 0.89 & 0.85 & 0.80 & 0.88 & 0.01 & 0.03 & 0.06 & 0.11 & 0.04 \\
\hline
\end{tabular}

\end{tiny}

\end{table*}

\begin{figure}[t]
     \centering
     \begin{subfigure}[b]{.49\linewidth}
         \centering
         \includegraphics[width=\linewidth]{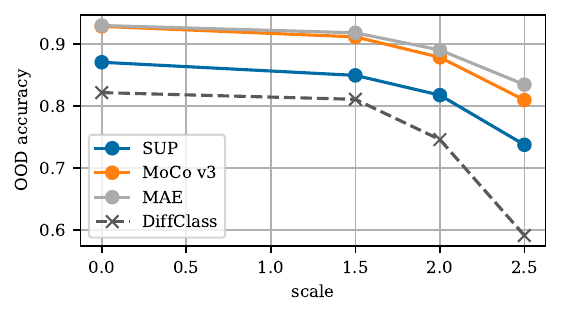}
         \caption{Accuracies for heavy snow.
         }
         \label{fig:acc_dit_snow}
     \end{subfigure}
     \hfill
     \begin{subfigure}[b]{.49\linewidth}
         \centering
         \includegraphics[width=\linewidth]{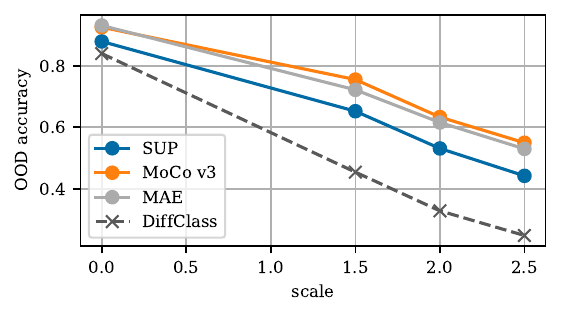}
         \caption{Accuracies for cartoon style.
         }
         \label{fig:acc_dit_cartoon}
     \end{subfigure}
             \caption{\textbf{Comparison of DiT classifier.} We report the OOD accuracies for two shifts for the DiT classifier \citep{li2023diffusion} and discriminative classifiers. All models were trained on ImageNet-1k and are evaluated on the same subset of our benchmark. The diffusion classifier performs worse than the discriminative models.
             }
        \label{fig:results_dit}
\end{figure}

The accuracies for the diffusion classifier are depicted in \cref{fig:results_dit}.
Similar to the discussion in the paper, the results showcase that the generative classifier is less robust than a supervised classifier.
We use the DiT-based diffusion classifier trained on ImageNet-1k using the available framework \citep{li2023diffusion} and the default hyper-parameters with a resolution of 256.
Due to high computational costs, we compute the results for 100 classes, four scales, for the snow and cartoon style shift, and for at most 20 seeds per class, scale, and shift.

\section{Fine-tuning with Synthetic Data}\label{sec:finetuning_inr}

We fine-tune a ResNet-50 classifier using our synthetic data. We compare the original ImageNet-trained model to a model fine-tuned using 50\% synthetic data and 50\% ImageNet training data. As shown in \cref{table:finetuning_rn50}, the fine-tuned model leads to improved performance on the shifted real-world dataset, without a significant decline on the original ImageNet dataset.

\begin{table}[t]
\caption{\textbf{ImageNet-R performance after fine-tuning on our benchmark data.}ImageNet-R accuracy of the original ResNet-50 without fine-tuning and our model, fine-tuned on our benchmark.}
\label{table:finetuning_rn50}
\vspace{-5pt}
\begin{center}
\begin{tabular}{l|cc}
\toprule
Evaluation Dataset & wo/ fine-tuning & w/ fine-tuning \\
\midrule
IN/val & 80.15 & 78.11 \\
IN/R   & 27.34 & 37.57 \\
\bottomrule
\end{tabular}
\end{center}
\end{table}

\section{Accuracy Drops on ImageNet-C}\label{sec:inc_ranking}
\begin{figure}[ht]
        \centering
\includegraphics[width=1.\linewidth]{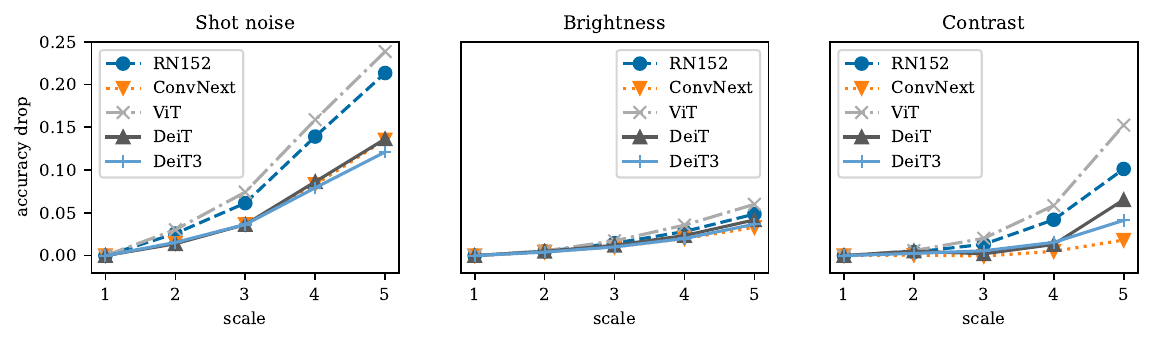}
    \caption{\textbf{Accuracy drops for three ImageNet-C corruptions and various architectures.} The model rankings change for different corruptions, underlining the importance of the selection of the corruption types or nuisance shifts for benchmarking the OOD robustness. Additionally, it can also be observed that the accuracy drops at varying rates for different shifts.}
    \label{fig:acc_drops_inc_various}
\end{figure}

\begin{figure}[!ht]
        \centering
\includegraphics[width=0.6\linewidth]{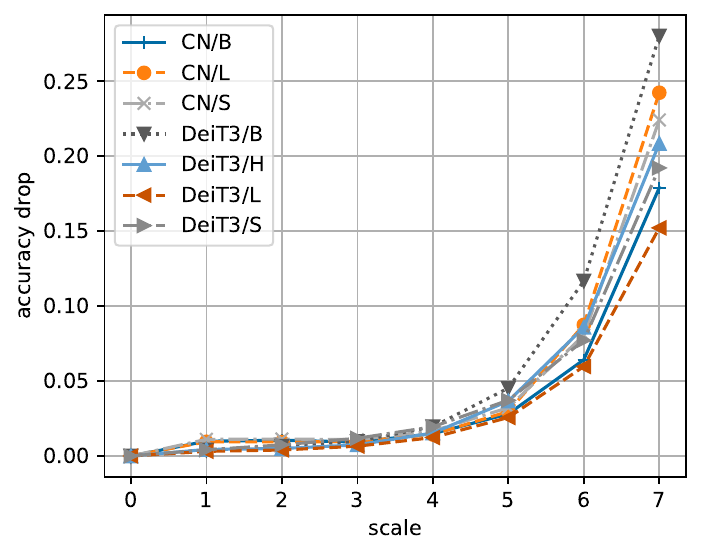}
    \caption{\textbf{Accuracy drops for contrast corruption.} We report the accuracy drops for seven severities of the contrast corruption, as defined in \citep{hendrycks_benchmarking_2019}. The model rankings change for different scales.}
    \label{fig:acc_drops_inc_contrast}
\end{figure}

We provide more evidence that the model rankings can change for different scales for ImageNet-C as well. 
We consider seven levels of contrast as a deterministic example corruption from ImageNet-C, based on the implementation of \citet{hendrycks_benchmarking_2019}. 
We present the accuracy drops for all corruption levels in \cref{fig:acc_drops_inc_various} and \cref{fig:acc_drops_inc_contrast}.
%As shown in Tab. 1 of the rebuttal, model rankings shift across different levels. 
Similar to our benchmark, a global averaged metric fails to capture such variations.

\section{Comparison to ImageNet-C/R}
\begin{figure}[t]
     \centering
     \begin{subfigure}[b]{.25\linewidth}
         \centering
         \includegraphics[width=\linewidth]{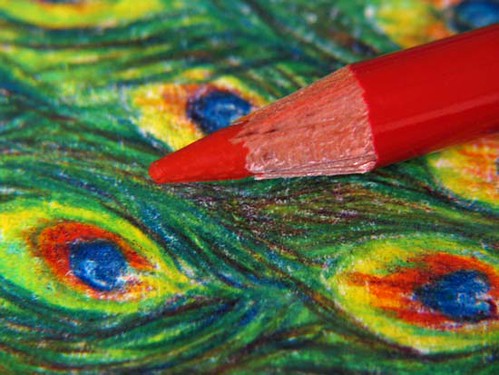}
         \label{fig:art_0}
     \end{subfigure}
     \hfill
      \begin{subfigure}[b]{.25\linewidth}
         \centering
         \includegraphics[width=\linewidth]{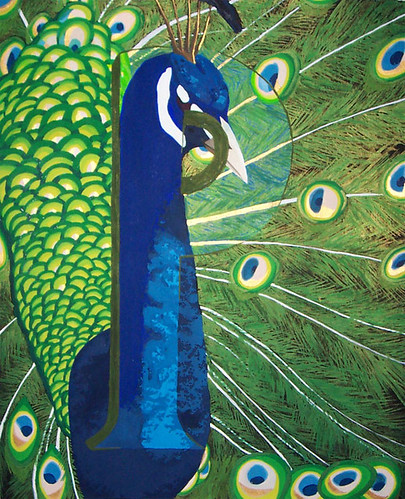}
         \label{fig:art_10}
     \end{subfigure}
     \hfill
      \begin{subfigure}[b]{.2\linewidth}
         \centering
         \includegraphics[width=\linewidth]{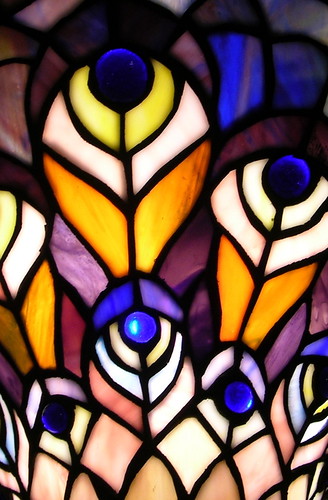}
         \label{fig:art_6}
     \end{subfigure}
                  \caption{\textbf{ImageNet-R examples.} Example images of one class where the shape and perspective significantly change.}
        \label{fig:inr_examples}
        \vspace{-10pt}
\end{figure}

While ImageNet-R evaluates style shifts, it includes confounders, such as heavy shape and perspective changes (\cref{fig:inr_examples}).
Our approach aims at reducing such factors by reducing variations of the spatial structure of the image when gradually applying the shift.

\begin{figure}[ht]
    \centering
    \includegraphics[width=1\linewidth]{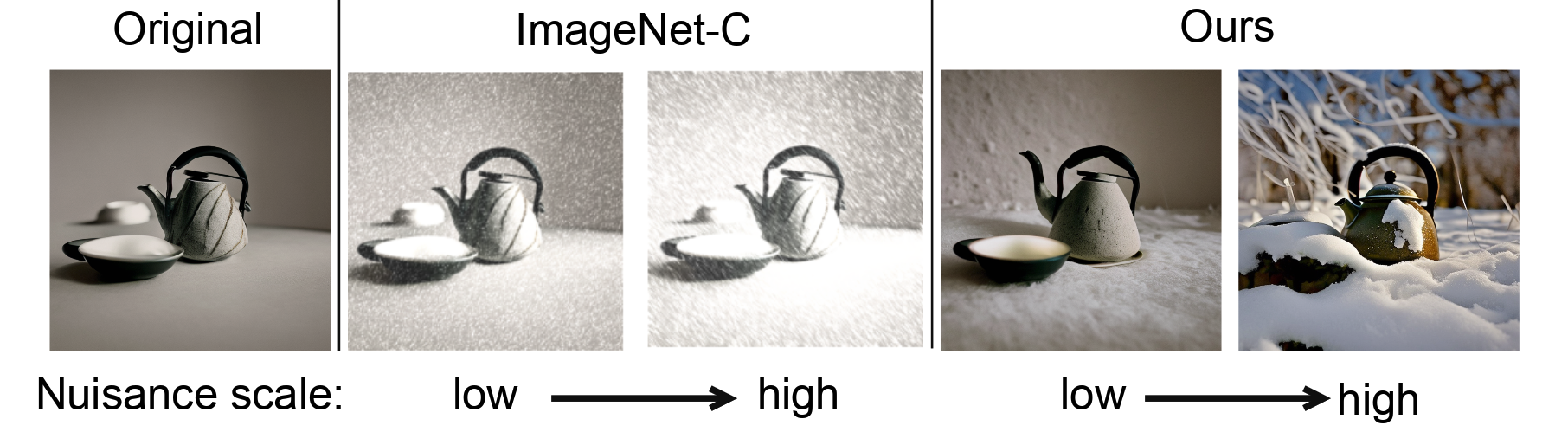}
    \vspace{-20pt}
    \caption{\textbf{Illustration of difference between ImageNet-C and CNS.} While ImageNet-C analyzes only synthetic shifts, CNS capture real-world distribution shifts..}
    \label{fig:compare_inc_and_cns}
    \vspace{-7pt}
\end{figure}

\section{Discussion of Accuracy-on-the-Line}\label{sec:statistics_acc_on_the_line}

\begin{figure}
    \centering
    \includegraphics[width=\linewidth]{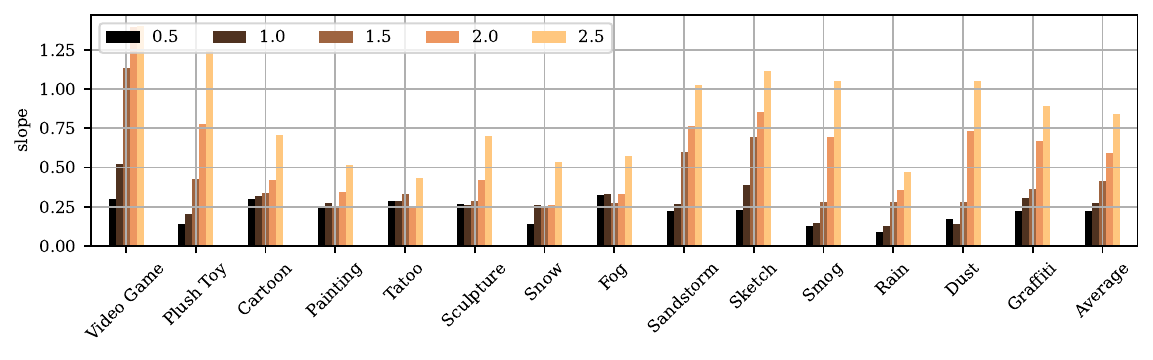}
    \caption{\textbf{Slope of ID and OOD accuracies.} We report the slope computed for 16 ImageNet-trained models.
    } 
    \label{fig:id_ood_slope}
    %\vspace{-10pt}
\end{figure}

\begin{figure}[h!]
        \centering
\includegraphics[width=1.\linewidth]{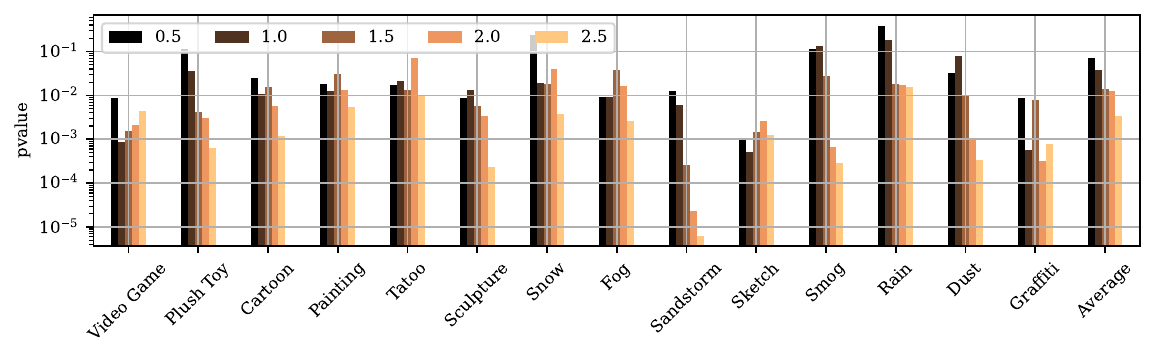}
    \caption{p-values of the linear regressions corresponding to the plot in \cref{fig:id_ood_slope}: The p-value is smaller than $0.05$ for most scales and shifts, providing evidence for the statistical significance of our statements.}
    \label{fig:id_ood_slope_pvalue}
\end{figure}

\begin{figure}[!ht]
        \centering
\includegraphics[width=1.\linewidth]{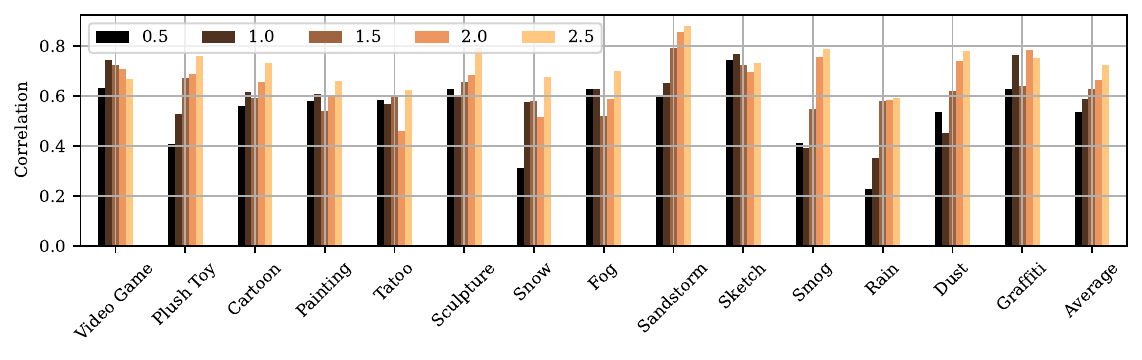}
    \caption{We report the linear correlation coefficients between
ID and OOD accuracy using 16 supervised ImageNet-trained models for all evaluated shifts. The
relation varies for different shifts and scales between 0.5 and 2.5.}
    \label{fig:correlation}
\end{figure}

We observe that larger models obtain higher OOD accuracies, \ie, smaller accuracy drops, as shown in \cref{fig:acc_drop_models} (\textit{Model size}).
However, ID and OOD accuracy are correlated, as we show in \cref{fig:correlation}. 
As prior work~\citep{Miller2021AccuracyOT} has shown that ID and OOD accuracy relate linearly, \ie, \textit{accuracy on the line}, we want to study whether the larger parameter count explains the higher robustness or whether this is solely explained by the accuracy-on-the-line observation. 
Therefore, we remove the effect of the ID accuracy on the OOD accuracy by computing the partial correlation between model size and OOD accuracy.
\cref{fig:id_ood_slope} show the slopes for various shifts and \cref{fig:id_ood_slope_pvalue} provides the p-values of the linear regression corresponding to the presented results in \cref{fig:id_ood_slope}.
This partial correlation coefficient is significantly negative ($\rho_{\text{size,OOD}\cdot \text{ID}}=\mbox{-}0.358$ for the DeiT3 family). 
Therefore, we conclude from our analysis that the improvements can be explained by the improved ID accuracy but not by more parameters. 

\begin{figure}[t]
     \centering
     \begin{subfigure}[b]{.49\linewidth}
         \centering
    
         \includegraphics[width=0.9\linewidth]{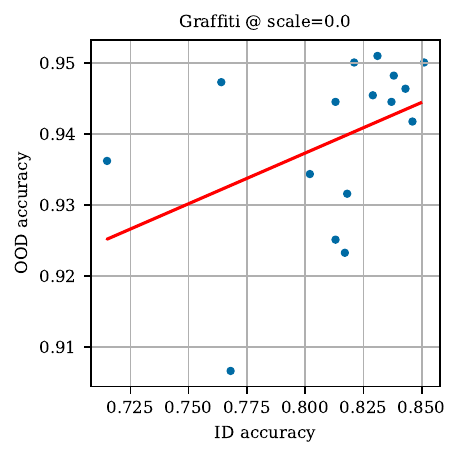}
         \caption{Fit for no applied shift.}
         \label{fig:id_ood_slope_1}
     \end{subfigure}
     \hfill
     \begin{subfigure}[b]{.49\linewidth}
         \centering
         \includegraphics[width=0.9\linewidth]{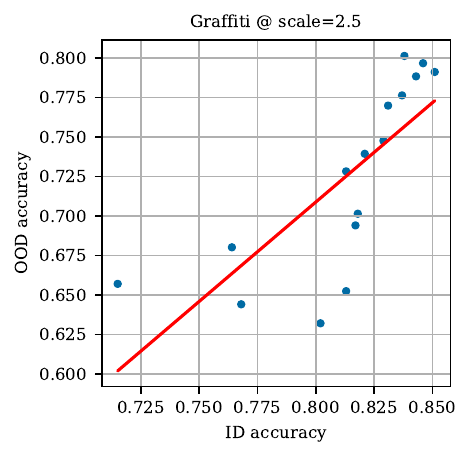}
         \caption{Fit for a graffiti shift with scale 2.5.}
         \label{fig:id_ood_slope_2}
     \end{subfigure}
             \caption{\textbf{Linear fits of the ID and OOD accuracies.} We plot example linear fits of ID and OOD accuracies for the graffiti style. It can be observed that the slope increases for a larger scale.}
        \label{fig:id_ood_slopes}
\end{figure}

\begin{figure}[ht]
        \centering
    \includegraphics[width=.7\linewidth]{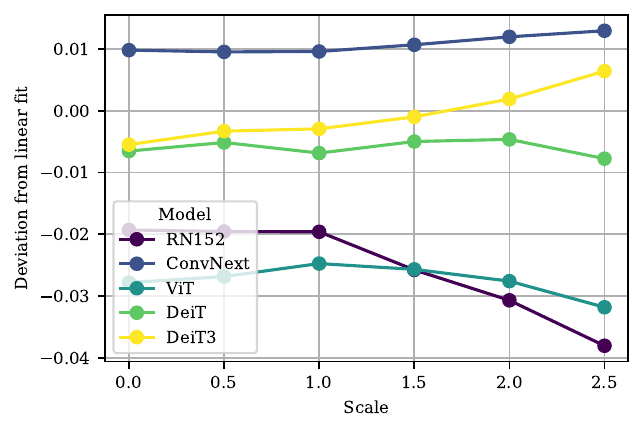}
    \caption{\textbf{Accuracy gains of models along the architecture axis.} 
    We plot the accuracy gains averaged over all shifts after correcting for the effect of the ID-OOD accuracy slope. These gains are computed by substracting the effect of the linear fit (consider \cref{fig:id_ood_slopes} for an example) from the OOD accuracies. After that correction, ConvNext performs  better than DeiT3.}
    \label{fig:acc_gain_after_id_ood_slope_correction}
\end{figure}

We further explore how removing the linear relation (as, e.g., in \cref{fig:id_ood_slopes}) explains the better OOD accuracy in \cref{fig:acc_gain_after_id_ood_slope_correction}.

\begin{table}[ht]
  \caption{\textbf{ImageNet validation accuracies and parameter count}. One the left, we plot model accuracies on the ImageNet validation dataset for all evaluated classifiers. On the right, we present the parameter counts for the used architectures.}
  \label{table:in_val_accs}
  \centering

\begin{tiny}

\centering
\begin{minipage}{0.45\linewidth}
\centering
\begin{tabular}{lc}
\toprule
Model & IN/val \\
\midrule
clip\_resnet101 & 58.00 \\
clip\_resnet50 & 55.00 \\
clip\_vit\_base\_patch16\_224 & 67.70 \\
clip\_vit\_base\_patch32\_224 & 62.60 \\
clip\_vit\_large\_patch14\_224 & 75.00 \\
clip\_vit\_large\_patch14\_336 & 76.30 \\
convnext\_base.fb\_in1k & 83.80 \\
convnext\_large.fb\_in1k & 84.30 \\
convnext\_small.fb\_in1k & 83.10 \\
convnext\_tiny.fb\_in1k & 82.10 \\
convnextv2\_base.fcmae\_ft\_in1k & 84.90 \\
convnextv2\_huge.fcmae\_ft\_in1k & 86.20 \\
convnextv2\_large.fcmae\_ft\_in1k & 85.80 \\
deit3\_base\_patch16\_224.fb\_in1k & 83.70 \\
deit3\_huge\_patch14\_224.fb\_in1k & 85.10 \\
deit3\_large\_patch16\_224.fb\_in1k & 84.60 \\
deit3\_medium\_patch16\_224.fb\_in1k & 82.90 \\
deit3\_small\_patch16\_224.fb\_in1k & 81.30 \\
deit\_base\_patch16\_224.fb\_in1k & 81.80 \\
dino\_lp\_vit\_base\_patch16 & 78.10 \\
dino\_v1\_vit\_base\_patch16 & 82.49 \\
dinov2\_vit\_base\_patch14 & 84.50 \\
dinov2\_vit\_base\_patch14\_reg & 84.60 \\
dinov2\_vit\_giant\_patch14 & 86.60 \\
dinov2\_vit\_giant\_patch14\_reg & 87.10 \\
dinov2\_vit\_large\_patch14 & 86.40 \\
dinov2\_vit\_large\_patch14\_reg & 86.70 \\
dinov2\_vit\_small\_patch14 & 81.40 \\
dinov2\_vit\_small\_patch14\_reg & 80.90 \\
mae\_vit\_base\_patch16 & 83.70 \\
mae\_vit\_huge\_patch14 & 86.90 \\
mae\_vit\_large\_patch16 & 86.00 \\
mocov3\_vit\_base\_patch16 & 83.20 \\
resnet101.a1\_in1k & 81.30 \\
resnet152.a1\_in1k & 81.70 \\
resnet18.a1\_in1k & 71.50 \\
resnet34.a1\_in1k & 76.40 \\
resnet50.a1\_in1k & 80.20 \\
vit\_base\_patch16\_224.augreg\_in1k & 76.80 \\
vit\_base\_patch16\_224.augreg\_in21k\_ft\_in1k & 77.70 \\
vit\_base\_patch16\_clip\_224.openai\_ft\_in1k & 85.20 \\
\bottomrule
\end{tabular}
\end{minipage}%
\hfill
\begin{minipage}{0.45\linewidth}
\centering
\begin{tabular}{lr}
\toprule
Model & Number of parameters (in M) \\
\midrule
convnext\_tiny & 29 \\
convnext\_small & 50 \\
convnext\_base & 89 \\
convnext\_large & 198 \\
convnextv2\_base & 89 \\
convnextv2\_huge & 660 \\
convnextv2\_large & 198 \\
deit3\_small & 22 \\
deit3\_medium & 39 \\
deit3\_base & 87 \\
deit3\_huge & 632 \\
deit3\_large & 304 \\
deit\_base & 87 \\
vit\_base & 87 \\
vit\_huge & 632 \\
vit\_large & 307 \\
resnet18 & 12 \\
resnet34 & 22 \\
resnet50 & 26 \\
resnet101 & 45 \\
resnet152 & 60 \\
\bottomrule
\end{tabular}

\end{minipage}

\end{tiny}
\end{table}

\begin{figure}[t]
     \centering
     \begin{subfigure}[b]{.45\linewidth}
         \centering
         \includegraphics[width=\linewidth]{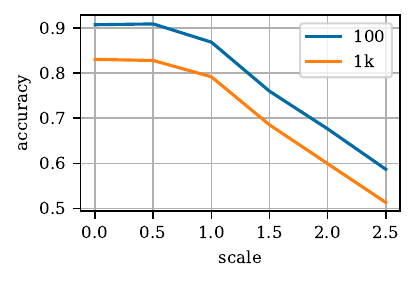}
         \caption{Accuracy over various scales.}
         \label{fig:IN1k_accs}
     \end{subfigure}
     \hfill
      \begin{subfigure}[b]{.45\linewidth}
         \centering
         \includegraphics[width=\linewidth]{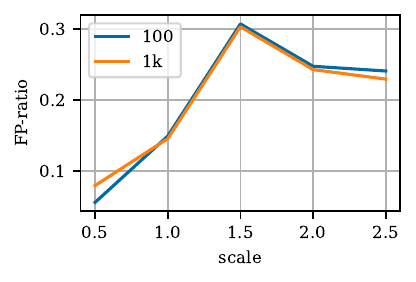}
         \caption{Failure point distribution.}
         \label{fig:IN1k_fps}
     \end{subfigure}
                  \caption{\textbf{Ablation of the number of ImageNet classes.} We compare the accuracies and failure points averaged over the selected 100 classes and all 1000 ImageNet classes for two shifts (snow and cartoon style). We report the results with ResNet-50. 
                  The results indicate that the initial accuracy estimate is overestimated but the accuracy drops averaged over the two shifts are in line.
                  The failure point distribution is normalized.)
                  }
        \label{fig:in1k_ablation}
\end{figure}

\newpage

\section{Implementation Details}
\label{sup:sec:implementation_details}
In this section, we provide more implementation details about the dataset generation process.

\subsection{Implementation Details for Image Generation}
We use the standard diffusers \citep{von-platen-etal-2022-diffusers} pipeline for Stable Diffusion 2.0, the DDIM \citep{songdenoising_2021} sampler with 100 steps and a guidance scale of 7.5, seeds ranging from 1 to 50.

\subsection{Implementation Details for Benchmarking} 
We integrate our new benchmark and additional models in the easyrobust \citep{mao2022easyrobust} framework.

\subsection{Details about the Used Compute}
We used the internal cluster consisting of NVIDIA A40, A100, and RTX 8000 GPUs for running most of the experiments. 
Small-scale experiments are conducted on workstations equipped with RTX 3090.
Training one LoRA adapter requires 1 to 2 hours depending on the used GPU, generating the images for one shift and class with 50 seeds and 6 scales requires 10 to 20 minutes.
Thus, the training of the 1400 LoRA adapters took around 2000 GPU hours and the generation of the images around 350 GPU hours.
%which, respectively, equaled around 2000 GPU hours and around 7500 GPU hours for the published benchmark in total. %14 shifts, 100 classes
Benchmarking all models using \textit{easyrobust} required around 1000 GPU hours.
The experiments to perform classification using the diffusion-classifier required around 4000 GPU hours.

\newpage

\subsection{Effect of Reduced Number of Classes for Benchmark Evaluation}\label{sec:ablation_img_gen}

We ablate how the number of classes influences the robustness evaluations in \cref{fig:in1k_ablation}. For a more efficient computation, we use the \verb|UniPCMultistepScheduler| sampler with 20 steps \citep{zhao2023unipc}.

\begin{figure}[t]
    \centering
    \includegraphics[width=1\columnwidth]{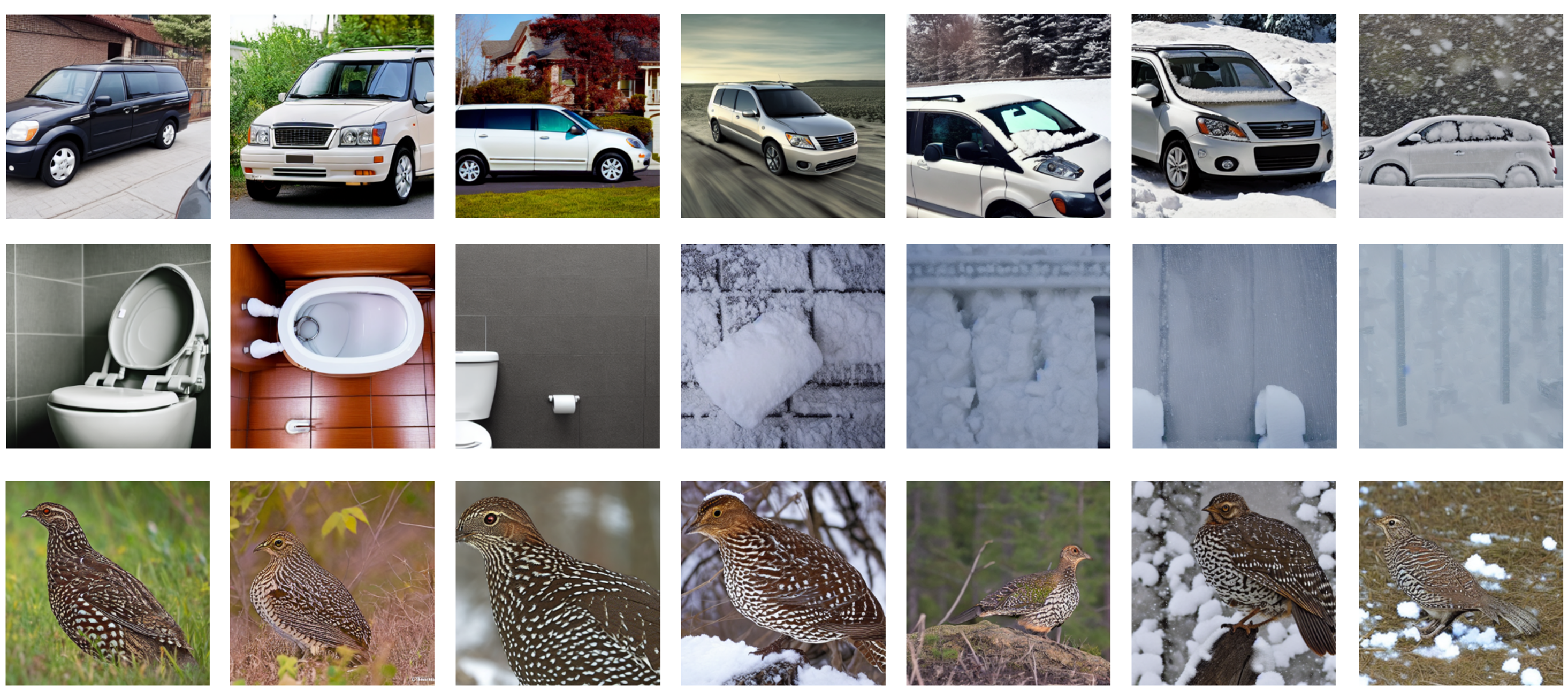}
    \caption{\textbf{Examples for text-based continuous shift.} The gradual increase can be successful. However, we observe that it fails for some classes (middle row) and is not always consistently increasing (bottom row).}
    \label{fig:examples_texsliding}
\end{figure}

\section{Design Choice for Text-Based Continuous Shift}
A naive approach for realizing continuous shifts involves computing the difference between two corresponding CLIP embeddings.
We explored this strategy following the implementation of \citet{baumann_continuous_2024}, but we did not achieve robust nuisance shifts for the variety of classes we considered and we present some examples in \cref{fig:examples_texsliding}.
We achieve reasonable results for some classes (\eg, upper row).
However, we observed that the spatial structures sometimes changes despite starting at later timesteps.
We observed that the naive approach is not very stable for some classes, resulting in OOD samples that do not represent realistic images (\eg, middle row). 
Applying the delta in text-embedding space also does not always result in a consistent increase of the considered shift (\eg, lower row).

We evaluate whether our sliders always increase the shift, as measured by the $\Delta$ CLIP score.
For this purpose, we compute the $\Delta$ CLIP scores when increasing the slider scale by $0.5$.
Here, the CLIP shift alignment increases for 73\% of all cases for scales $s>0$ and averaged over all shifts, demonstrating that increasing the slider weight results in a stronger severity of the desired shift.

\begin{table}[t]
  \caption{\textbf{Statistics of filtering process.} We report the number of in-class samples after various filtering stages. }
  \label{table:filtering_stages}
  \centering
  \begin{tabular}{lllll}
    \toprule
    Scale& Stage (i)& Stage (ii)&Stage (iii)&Stage (iv)\\
    \midrule
    0& 4000& 2966& 2966&2966\\
    0.5& 4000&  2966& 2929&2955\\
    1& 4000&  2966& 2813&2906\\
    1.5& 4000& 2966& 2479&2740\\
     2& 4000& 2966& 2143&2498\\
    2.5& 4000& 2966& 1729&2110\\
    \bottomrule

  \end{tabular}
\end{table}

\section{Labeling}\label{sec:labeling_details}
In this section, we provide more details about the labeling dataset and strategy.

\subsection{Details on the Creation of the Labeled Dataset}
To select a filter for detecting out-of-class (OOC) samples, we collected a manually labeled dataset.
For this, we pursued the following strategy:
(i) In the first stage, 24k images are generated for 20 seeds, 5 LoRA scales, and 2 shifts per class for 100 random ImageNet classes in total. 
We select two different shifts: One shift corresponds to a natural variation (snow), and the second shift corresponds to a style shift (cartoon style).
(ii) We aim to find OOC samples that are due to the application of the LoRA adapters. 
Therefore, we remove all images generated with a seed that results in a generated image with low CLIP text-alignment or that is not classified classified correctly even without the application of LoRA adapters.
After removing such images, the labeling dataset consists of around 18k images.
(iii) To reduce the labeling effort, we filter out all easy samples that (1) are correctly classified by DINOv2-ViT-L \citep{caron2021emerging,oquab2023dinov2} with a linear fine-tuned head and (2) one out of three classifiers (ResNet-50, DeiT-B/16, or ViT-B/16). (3) Additionally, we ensure a sufficiently high text alignment.
(iv) The remaining hard images are labeled by two human annotators.

Eeach annotator can choose from the labels `class', `partial class properties', and `not class', where the second option should be selected if the image partially includes some characteristics of the class.
An image is defined as an out-of-class sample if at least one annotator considers the image as an OOC sample. 
For the remaining samples, an image is considered IC (in-class) if at least one annotator labeled the image a clear sample of the corresponding class

\begin{figure}[t]
  \centering
  \includegraphics[width=0.5\columnwidth]{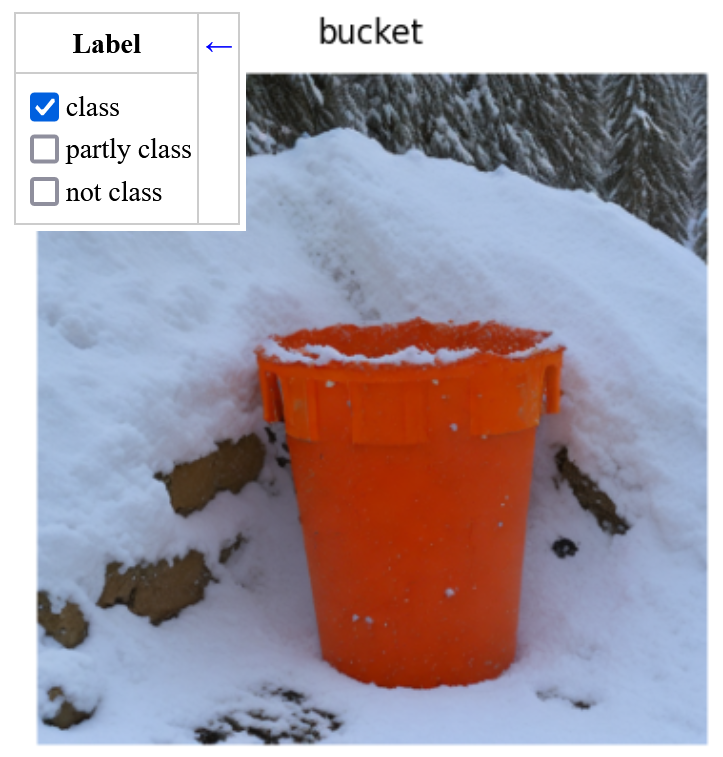}
  \caption{\textbf{Screenshot of labeling tool.} We plot a screenshot of an example image as it appeared during our labeling.}
  \label{fig:screenshot_labeling}
\end{figure}

\begin{figure}[t]
     \centering
     \begin{subfigure}[b]{.49\linewidth}
         \centering
         \includegraphics[width=\linewidth]{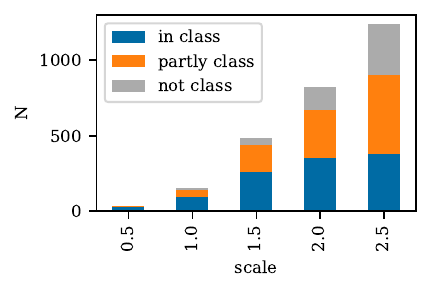}
         \caption{Human labeling dataset.}
         \label{fig:statistics_human_labels}
     \end{subfigure}
     \hfill
      \begin{subfigure}[b]{.49\linewidth}
         \centering
         \includegraphics[width=\linewidth]{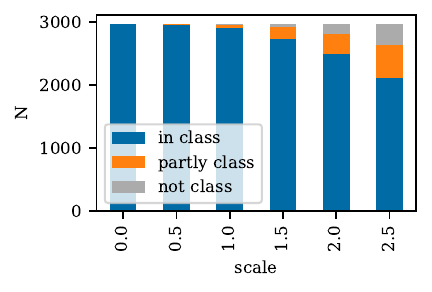}
         \caption{Complete filtering dataset.
        }
         \label{fig:statistics_labeling_dataset}
     \end{subfigure}
             \caption{\textbf{Statistics of labeling dataset.} We report the number of in-class, partially in-class, and out-of-class samples.}
        \label{fig:statistics_labeling}
        \vspace{-10pt}
\end{figure}

\begin{figure*}
    \centering
    \includegraphics[width=0.8\linewidth]{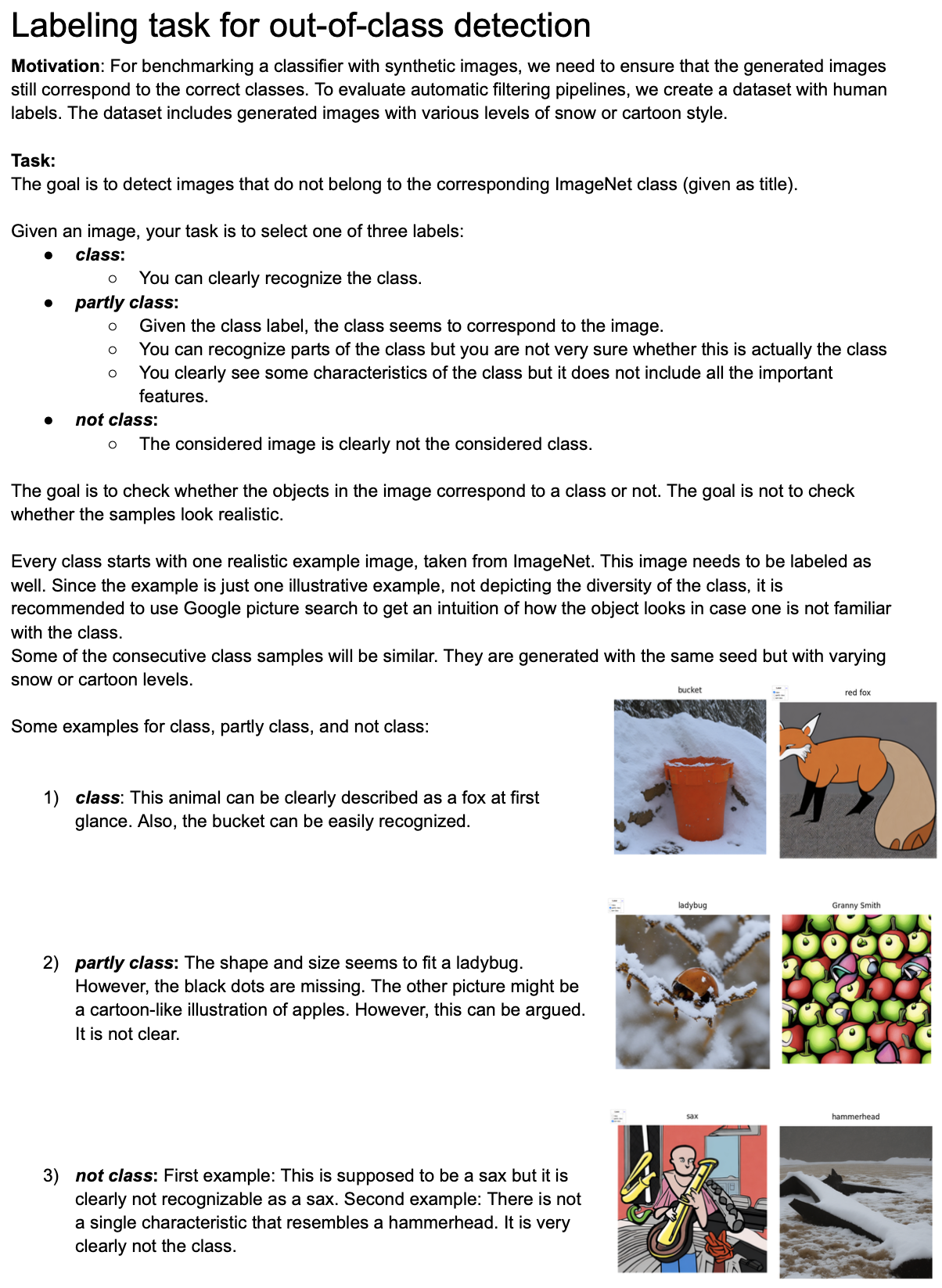}
    \caption{\textbf{Set of instructions for labeling.} Instructions provided to the human annotators to perform the labeling of the out-of-class filtering dataset.}
    \label{fig:instructions}
\end{figure*}

For the pre-filtering strategy (ii) and for the selection of easy samples (iii), we compute text-alignment using CLIP score and we remove all samples that have a CLIP similarity $s_\text{CLIP-text-alignment} >24$, which approximately includes 90\% of all ImageNet validation images \citep{vendrowDatasetInterfacesDiagnosing2023}. 
We use the implementation in \textit{torchmetrics} with VIT-B/16. 
After removing the easy samples in step (iii), 2.7k images remain for labeling.
We use the VIA annotation tool \citep{dutta2019vgg,dutta2016via} to create the annotations. Each image is labeled by two humans. 
In total, 14 graduate students are involved in the labeling process.
For all participants, we ensure sufficient motivation and they receive detailed instructions on how to perform the labeling (the full set of instructions is provided in \cref{fig:instructions}).
We provide the filtering statistics in \cref{table:filtering_stages} and the statistics of the labeled dataset in \cref{fig:statistics_labeling}.
An example screenshot of the labeling tool is visualized in \cref{fig:screenshot_labeling}.


\section{User Study}\label{sec:user_study}
We perform a user study on the final dataset using the same tooling as for the human labeling discussed in \cref{sec:labeling_details} (iv).
The user study includes 300 randomly sampled images from the benchmark and it is checked by two different individuals. In total, the user study involved seven people with different professions.
3 samples of our benchmark were considered as out-of-class samples, resulting in a ratio of 1\% of failure cases with a margin of error of 0.5\% for a one-sigma confidence level.

We also study when a shift is clearly visible and report it in \cref{tab:user_study_shift}. Model performance is evaluated only for 
030 seeds where all scales are valid.

\begin{table}[t]
\caption{\textbf{User study shift realism.} Distribution of images where the shift is clearly identifiable.}
    \label{tab:user_study_shift}
    \centering
    \begin{tabular}{lcc}
\toprule
Scale & Unclear & Clear \\
\midrule
1.0 & 0.76 & 0.24 \\
1.5 & 0.51 & 0.49 \\
2.0 & 0.24 & 0.76 \\
2.5 & 0.16 & 0.84 \\
\bottomrule
\end{tabular}
\end{table}

\section{Applications of Trained Sliders}
\begin{figure}[t]
  \centering
  \includegraphics[width=0.7\columnwidth]{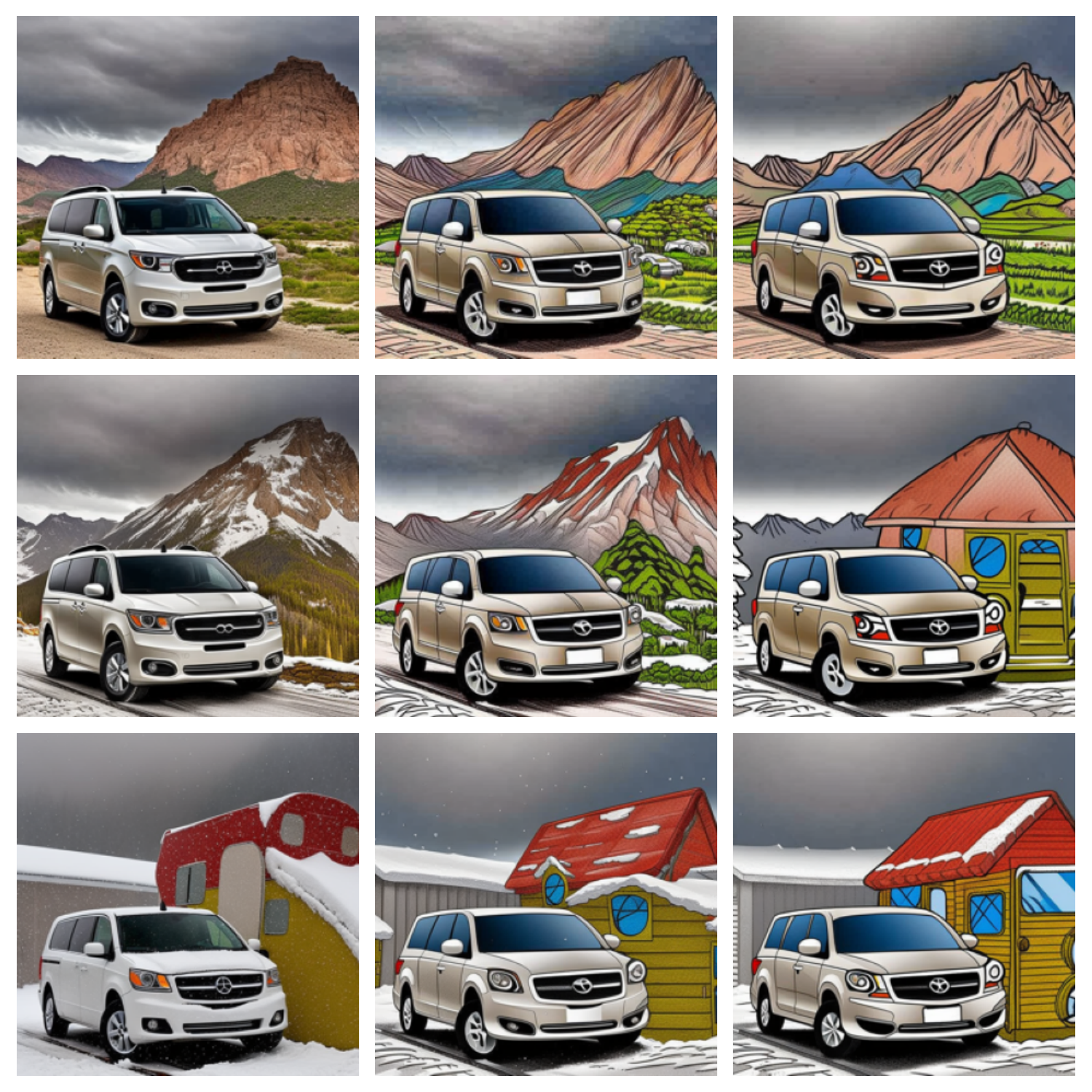}
  \caption{\textbf{Combination of Sliders.} We exemplarily show that sliders can be combined. Here, a snow slider (vertical axis) and a cartoon slider (horizontal axis) are linearly added for three scales.}
  \label{fig:combination_sliders}
  \vspace{-10pt}
\end{figure}

We can combine various sliders by simply adding the corresponding LoRA adapters. We show an example application in \cref{fig:combination_sliders}.

\section{OOD-CV Details}
\label{subsec:supp:ood_cv_details}
The Out-of-Distribution Benchmark for Robustness (OOD-CV) dataset includes real-world OOD examples of 10 object categories varying in terms of 5 nuisance factors: \textit{pose}, \textit{shape}, \textit{context}, \textit{texture}, and \textit{weather}.

\paragraph{Generation of images for synthetic OOD-CV}
We generate the images for the synthetic OOD-CV dataset using a larger number of noise steps (85\%) and more scales (between 0 and 3).
The shift sliders for these classes appear to be more robust potentially since these classes occur more often in the dataset for training CLIP and Stable Diffusion.
We use SD2.0 to generate the images.

\begin{figure*}[th]
    \centering
    \begin{subfigure}[b]{0.23\linewidth}
        \centering
        \includegraphics[height=3cm]{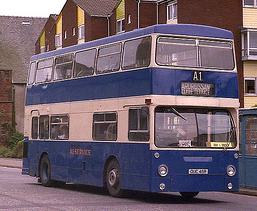}
        \caption{Train, ImageNet.}
    \end{subfigure}
    \hfill
    \begin{subfigure}[b]{0.23\linewidth}
        \centering
        \includegraphics[height=3cm]{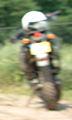}
        \caption{Train, ImageNet.}
    \end{subfigure}
    \hfill
    \begin{subfigure}[b]{0.23\linewidth}
        \centering
        \includegraphics[height=3cm]{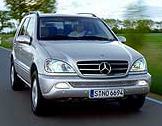}
        \caption{Train, ImageNet.}
    \end{subfigure}
    \hfill
    \begin{subfigure}[b]{0.23\linewidth}
        \centering
        \includegraphics[height=3cm]{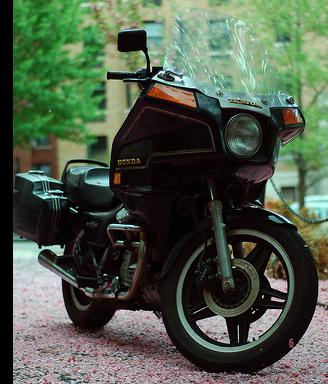}
        \caption{Train, ImageNet.}
    \end{subfigure}
    \vfill
    \begin{subfigure}[b]{0.23\linewidth}
        \centering
        \includegraphics[height=3cm]{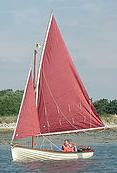}
        \caption{Train, Pascal-VOC.}
    \end{subfigure}
    \hfill
    \begin{subfigure}[b]{0.23\linewidth}
        \centering
        \includegraphics[height=3cm]{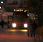}
        \caption{Train, Pascal-VOC.}
    \end{subfigure}
    \hfill
    \begin{subfigure}[b]{0.23\linewidth}
        \centering
        \includegraphics[height=3cm]{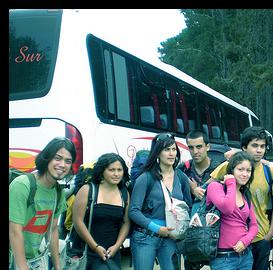}
        \caption{Train, Pascal-VOC.}
    \end{subfigure}
    \hfill
    \begin{subfigure}[b]{0.23\linewidth}
        \centering
        \includegraphics[height=3cm]{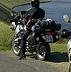}
        \caption{Train, Pascal-VOC.}
    \end{subfigure}
    \vfill
    \begin{subfigure}[b]{0.23\linewidth}
        \centering
        \includegraphics[height=2.5cm]{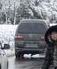}
        \caption{Test, snow shift.}
    \end{subfigure}
    \hfill
    \begin{subfigure}[b]{0.23\linewidth}
        \centering
        \includegraphics[height=2.5cm]{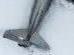}
        \caption{Test, snow shift.}
    \end{subfigure}
    \hfill
    \begin{subfigure}[b]{0.23\linewidth}
        \centering
        \includegraphics[height=2.5cm]{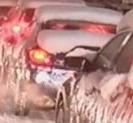}
        \caption{Test, snow shift.}
    \end{subfigure}
    \hfill
    \begin{subfigure}[b]{0.23\linewidth}
        \centering
        \includegraphics[height=2.5cm]{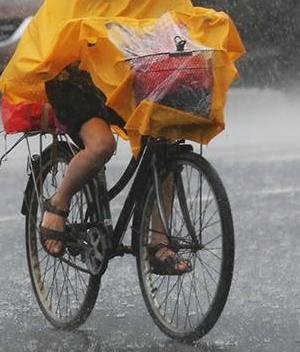}
        \caption{Test, rain shift.}
    \end{subfigure}
    \caption{\textbf{OOD-CV example images.} We illustrate a set of example images from the training and the testing dataset of OOD-CV: (a-h) example from the training set, from ImageNet or Pascal-VOC. (i-l) Some examples for weather nuisance shifts. In the training set, we observe that images from the Pascal-VOC subset are usually of lower quality (\eg, cropping, occlusion, resolution) compared to the ImageNet subset. In the test set, we see that they are not fully disentangled (\eg, (j) is only partially visible, (k) is partially occluded).}
    \label{fig:sup:ood_cv_train_examples}
\end{figure*}

\paragraph{Training subset}
The OOD-CV benchmark provides a training subset of $8627$ images. 
We train various classifiers (i.e., ResNet-50 \citep{he2016deep}, ViT-B/16 \citep{dosovitskiy2020image}, and DINO-v2-ViT \citep{oquab2023dinov2}) for classification. We finetune each baseline during 50 epochs with an early stopping set to 5 epochs. We apply standard data augmentations such as scale, rotation, and flipping during training. The training subset is composed of images originating from different datasets, notably ImageNet \citep{deng2009imagenet} and Pascal-VOC \citep{Everingham10}.
It is important to notice that the distribution of these two subsets is slightly different, with a higher data quality for the ImageNet subset and a lower quality for the latter subset (more noise, smaller objects, different image sizes). We visualize a few examples of the training data in \cref{fig:sup:ood_cv_train_examples}.
\paragraph{Test subset annotations}
In the test subset provided in the benchmark dataset, only the coarse individual nuisance factors (\eg, \textit{weather}, \textit{texture}) are provided. In our setup, we are interested in studying more fine-grained nuisance shifts, notably \textit{rain}, \textit{snow}, or \textit{fog}. Hence, we had to assign some fine-grained annotation to all images containing \textit{weather} nuisance shifts.
%``\verb|A picture of a {class}|''
Hence, we assign a fine-grained annotation by computing the CLIP similarity to the following texts: ``a picture of a \verb|class| in \verb|shift|'', where \verb|class| is the ground truth class and \verb|shift| the nuisance shift candidate \textit{rain}, \textit{snow}, or \textit{fog} and ``a picture of a \verb|class| without snow nor fog nor rain''. By applying a softmax on the similarity scores with the previous texts, we can assign the fine-grained nuisance shift \textit{rain}, \textit{snow}, \textit{fog} or \textit{unknown} for each image. We show more statistics in \cref{table:ood_cv_statistics}. 
By checking the results visually, we observe that all fine-grained nuisance shifts align with human perception and have a tendency towards classifying samples as \textit{unknown} as soon as there is a small doubt. Note that by applying the same strategies to our generated data, we obtain an accuracy close to $100\%$. 

\begin{figure*}[t]
    \centering
    \begin{subfigure}[b]{0.65\linewidth}
        \centering
        \includegraphics[width=\linewidth]{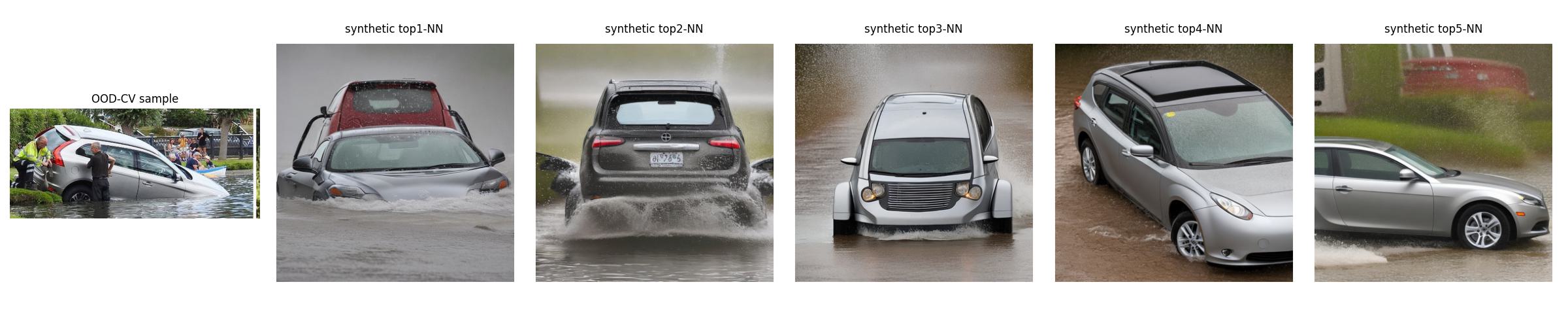}
    \end{subfigure}
    \begin{subfigure}[b]{0.65\linewidth}
        \centering
        \includegraphics[width=\linewidth]{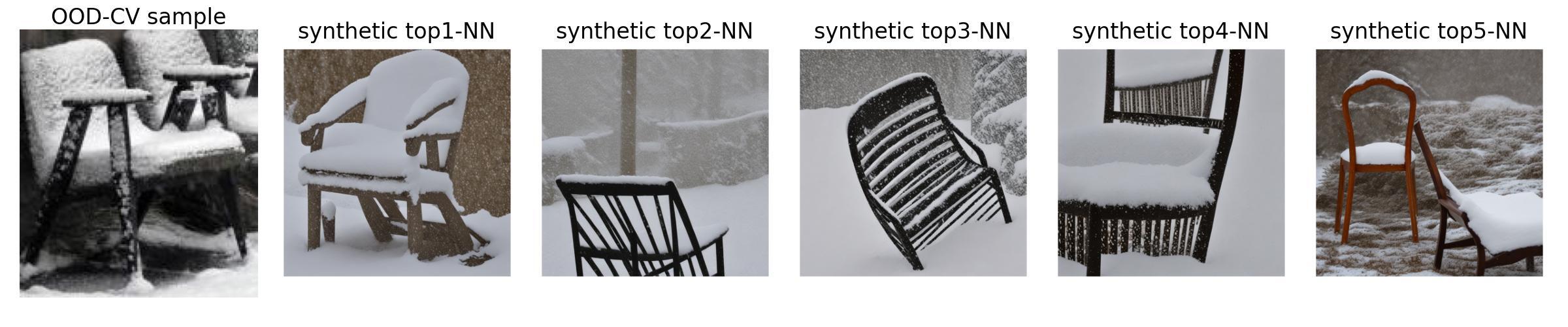}
    \end{subfigure}
    \begin{subfigure}[b]{0.65\linewidth}
        \centering
        \includegraphics[width=\linewidth]{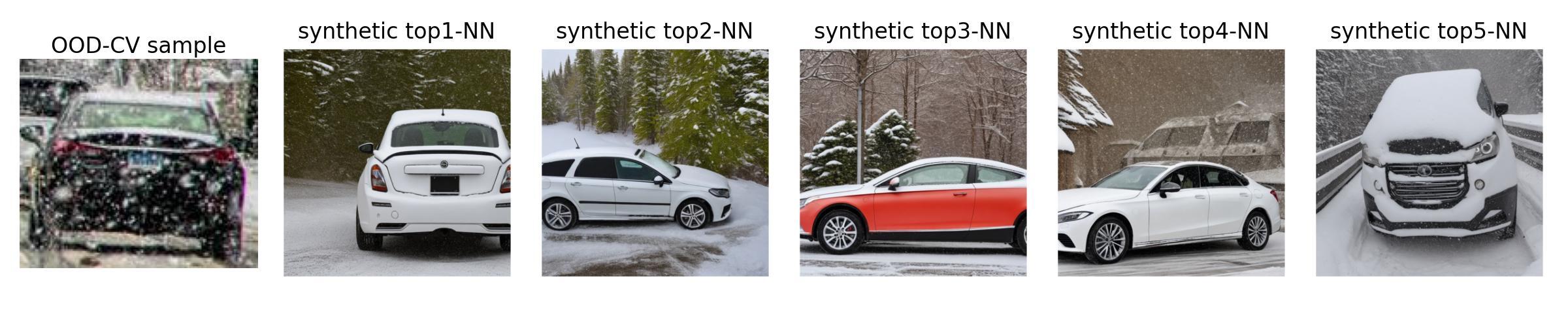}
    \end{subfigure}
    \begin{subfigure}[b]{0.65\linewidth}
        \centering
        \includegraphics[width=\linewidth]{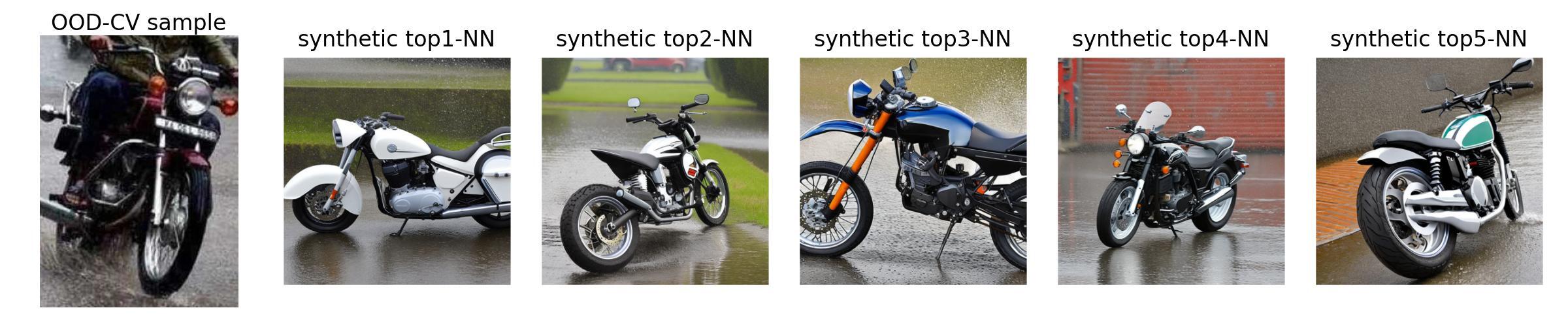}
    \end{subfigure}
    \begin{subfigure}[b]{0.65\linewidth}
        \centering
        \includegraphics[width=\linewidth]{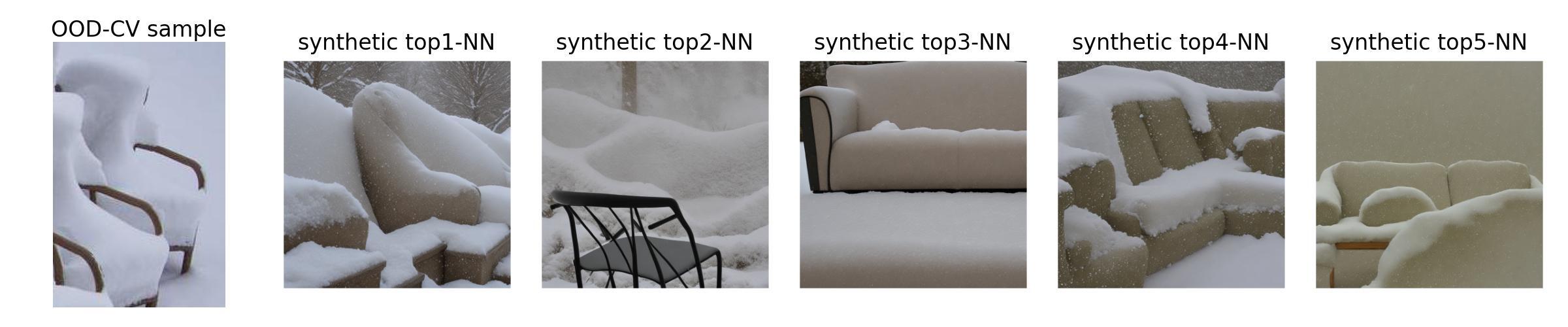}
    \end{subfigure}
    \begin{subfigure}[b]{0.65\linewidth}
        \centering
        \includegraphics[width=\linewidth]{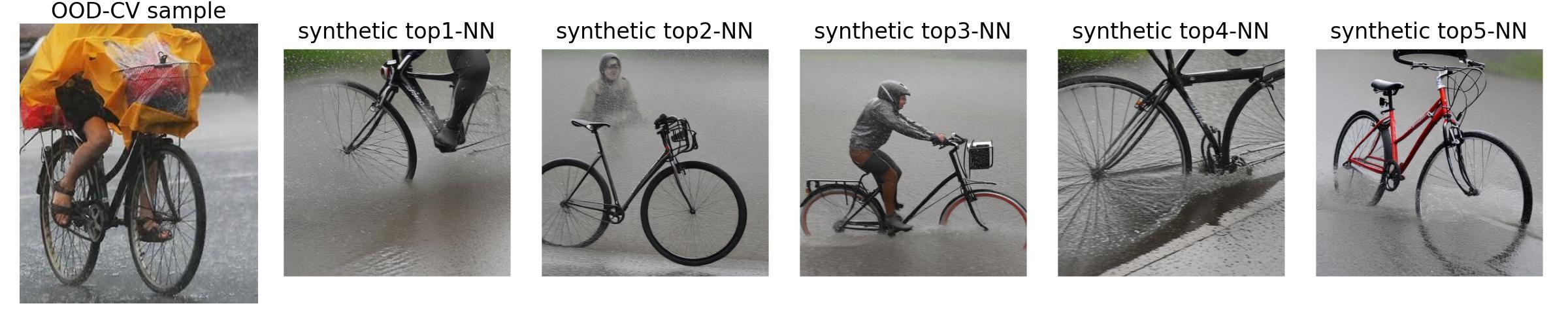}
    \end{subfigure}
    \begin{subfigure}[b]{0.65\linewidth}
        \centering
        \includegraphics[width=\linewidth]{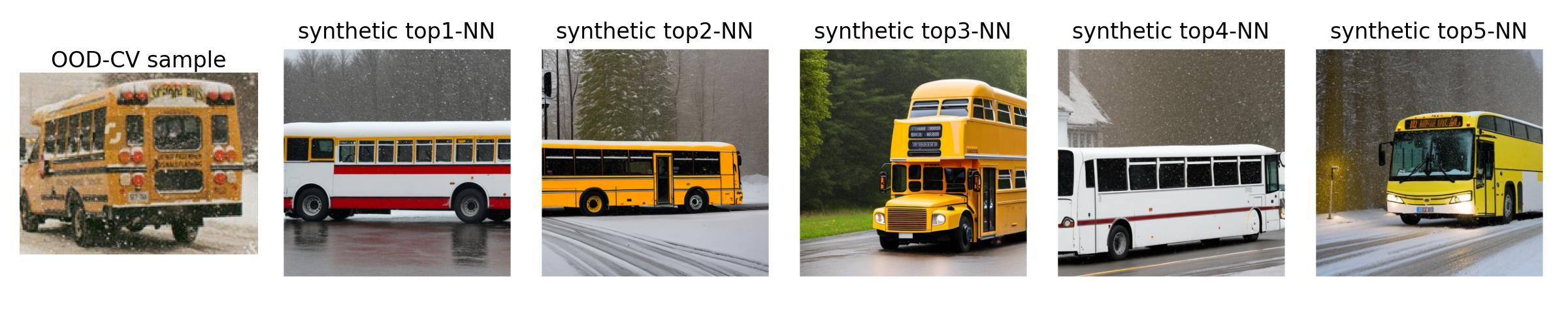}
    \end{subfigure}

    \caption{\textbf{Closest synthetic samples to two example OOD-CV images.} We find the top-5 nearest neighbours using cosine similarity with CLIP image embedding.}
    \label{fig:nn_oodcv}
\end{figure*}

\paragraph{Nearest neighbor images of OOD-CV and CNS-Bench.}
To illustrate the realism of our generated image, we compute the nearest neighbours using cosine similarity with CLIP image embedding and we plot it in \cref{fig:nn_oodcv}.

\begin{table}[t]
  \caption{\textbf{OOD-CV Statistics.} We report the number of images and accuracies for the weather subset.}
  \label{table:ood_cv_statistics}
  \centering
  \begin{tabular}{lll}
    \toprule
    Shift& \#images &Accuracy\\
    \midrule
    Snow& 273&70.3\\
    Fog& 24& 62.5\\ 
    Rain& 74&66.2\\
    Unknown&  129&66.7 \\
    Total& 500 & 68.4\\
    \bottomrule

  \end{tabular}
\end{table}

\begin{figure}[t]
  \centering
  \includegraphics[width=.7\columnwidth]{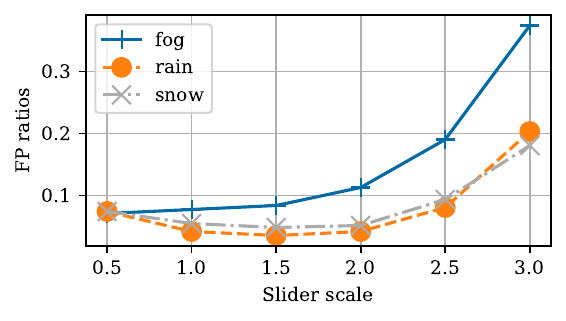}
  \caption{\textbf{Failure point distribution of a ResNet-50 classifier on our continuous OOD-CV benchmark.} 
  Our benchmark allows computing the failure distribution of failure points, allowing the analysis of when classifiers tend to fail, which was not possible using the manually labeled images.
  }
  \label{fig:exps_oodcv_bps}
  \vspace{-10pt}
\end{figure}

\textbf{Failure point distribution for CNS-Bench (OOD-CV)}
\cref{fig:exps_oodcv_bps} depicts the failure distribution for the three shifts.

~
\newpage
~
\clearpage

\section{Datasheet}
\label{sec:datasheet}
In the following, we answer the questions as proposed in \citet{gebru2021datasheets}.
\subsection{Motivation}
\customformat{For what purpose was the dataset created?}
{Was there a specific task in mind? Was there a specific gap that needed to be filled? Please provide a description.}
{%ANSWER
The dataset was created to evaluate the robustness of state-of-the-art models to specific continuous nuisance shifts.
Current approaches are not scalable and often include only a small variety of nuisance shifts, which are not always relevant in the real world. More importantly, current benchmark datasets define binary nuisance shifts by considering the existence or absence of that shift, which may contradict their continuous realization in real-world scenarios.
}
\customformat{Who created the dataset (e.g., which team, research group) and on behalf of which entity (e.g., company, institution, organization)?}{}
{
The paper was created by the authors of the CNS-Bench paper, which are affiliated with the listed organizations.
}
\customformat{Who funded the creation of the dataset?}{If there is an associated grant, please provide the name of the grantor and the grant name and number.}
{
The creation was funded by by the German Science Foundation (DFG) under Grant No. 468670075.
}
\subsection{Composition}

\customformat{What do the instances that comprise the dataset represent (e.g., documents, photos, people, countries)?}{}
{
The dataset consists of synthetic images that were generated using Stable Diffusion.
}
\customformat{How many instances are there in total (of each type, if appropriate)?}{}
{
The dataset contains $192,168$ images in total, with $32,028$ for each of the six scales with 14 shifts. Each shift has at least $5,000$ images and $100$ classes.
}
\customformat{Does the dataset contain all possible instances or is it a sample (not necessarily random) of instances from a larger set?}{ If the dataset is a sample, then what is the larger set? Is the sample representative of the larger set (e.g., geographic coverage)? If so, please describe how this representativeness was validated/verified. If it is not representative of the larger set, please describe why not (e.g., to cover a more diverse range of instances because instances were withheld or unavailable).}
{
The dataset contains the subset of images that were filtered using the selected filtering strategy. Originally, $420,000$ images were generated.
}
\customformat{What data does each instance consist of? ``Raw'' data (e.g., unprocessed text or images) or features?}{ In either case, please provide a description.}
{
``Raw'' synthetically generated data as described in the paper.
}
\customformat{Is there a label or target associated with each instance?} {If so, please provide a description.}
{
Yes, each image belongs to an ImageNet class and has a shift scale assigned to it.
}
\customformat{Is any information missing from individual instances?} {If so, please provide a description, explaining why this information is missing (e.g., because it was unavailable). This does not include intentionally removed information, but might include, e.g., redacted text.}
{
No, for each instance, we give the class label, the shift and its scale, and the parameters used for generating this image. 
However, the class label might be erroneous in rare cases where the generated image corresponds to an out-of-class sample.
}
\customformat{Are relationships between individual instances made explicit (e.g., users with their tweets, songs with their lyrics, nodes with edges)?}{If so, please describe how these relationships are made explicit.}
{
Yes, the relationships in terms of class, random seed for generation, shift, and scale of shift are provided in the dataset.
}
\customformat{Are there recommended data splits (e.g., training, development/validation, testing)?}{If so, please provide a description of these splits, explaining the rationale behind them.}
{
We offer a benchmark dataset specifically intended for testing the robustness of classifiers. 
Therefore, we recommend utilizing the entire dataset provided as the test dataset. 
}
\customformat{Are there any errors, sources of noise, or redundancies in the dataset?}{If so, please provide a description.}
{
We provided a dataset of generated images.
While we apply a filtering strategy to reduce the number of out-of-class and unrealistic samples, we cannot guarantee that all images of the dataset represent a realistic and visually appealing realization of the considered class.
We provide a statistical estimate of the number of failure samples in the paper.
The data might also include the redundancies that underlie the image generation process of Stable Diffusion.
}
\customformat{Is the dataset self-contained, or does it link to or otherwise rely on external resources (e.g., websites, tweets, other datasets)?}{If it links to or relies on external resources, a) are there guarantees that they will exist, and remain constant, over time; b) are there official archival versions of the complete dataset (i.e., including the external resources as they existed at the time the dataset was created); c) are there any restrictions (e.g., licenses, fees) associated with the use of these external resources?}
{
The dataset is fully self-contained.
}
\customformat{Does the dataset contain data that might be considered confidential (e.g., data that is protected by legal privilege or by doctor–patient confidentiality, data that includes the content of individuals’ non-public communications)?}{If so, please provide a description.}
{
No.
}
\customformat{Does the dataset contain data that, if viewed directly, might be offensive, insulting, threatening, or might otherwise cause anxiety?}{If so, please describe why.}
{
There is a small chance that our synthetically generated data can generate offensive images. However, we did not encounter any such sample during our extensive manual annotations.
}
\customformat{Does the dataset relate to people? If not, you may skip the remaining questions in this section.}{}
{
No.
}
\customformat{Does the dataset identify any subpopulations (e.g., by age, gender)?}{If so, please describe how these subpopulations are identified and provide a description of their respective distributions within the dataset.}
{
N/A.
}
\customformat{Is it possible to identify individuals (i.e., one or more natural persons), either directly or indirectly (i.e., in combination with other data) from the dataset?}{ If so, please describe how.}
{
N/A.
}
\customformat{Does the dataset contain data on individuals’ protected characteristics (e.g., age, gender, race, religion, sexual orientation)?}{If so, please describe this data and how it was obtained.}
{
N/A.
}
\customformat{Does the dataset contain data on individuals’ criminal history or other behaviors that would typically be considered sensitive or confidential?}{If so, please describe this data and how it was obtained.}
{
N/A.
}

\subsection{Collection Process}

\customformat{How was the data associated with each instance acquired? Was the data directly observable (e.g., raw text, movie ratings), reported by subjects (e.g., survey responses), or indirectly inferred/derived from other data (e.g., part-of-speech tags, model-based guesses)?}{}
{
N/A.
}
\customformat{What mechanisms or procedures were used to collect the data (e.g., hardware apparatus or sensor, manual human curation, software program, software API)? How were these mechanisms or procedures validated?}{}
{
We used Stable Diffusion 2.0 to generate all images.
Images were generated using NVIDIA A100 and A40 GPUs.
}
\customformat{If the dataset is a sample from a larger set, what was the sampling strategy (e.g., deterministic, probabilistic with specific sampling probabilities)?}{}
{
The dataset was filtered using a combinatorial selection approach using the alignment scores of DINOv2 and CLIP to the considered class.
}
\customformat{Who was involved in the data collection process (e.g., students, crowdworkers, contractors) and how were they compensated (e.g., how much were crowdworkers paid)?}{}
{
The authors of the paper and other PhD students of the institute. They were not additionally paid for the dataset collection process.
}
\customformat{Over what timeframe was the data collected? Does this timeframe match the creation timeframe of the data associated with the instances (e.g., recent crawl of old news articles)?}{ If not, please describe the timeframe in which the data associated with the instances was created.}
{
The images were generated and processed over a timeframe of four weeks.
}
\customformat{Were any ethical review processes conducted (e.g., by an institutional review board)?}{ If so, please provide a description of these review processes, including the outcomes, as well as a link or other access point to any supporting documentation.}
{
No ethical concerns.
}

\subsection{Preprocessing/cleaning/labeling}

\customformat{Was any preprocessing/cleaning/labeling of the data done (e.g., discretization or bucketing, tokenization, part-of-speech tagging, SIFT feature extraction, removal of instances, processing of missing values)?}{ If so, please provide a description. If not, you may skip the remaining questions in this section.}
{
Yes, cleaning of the generated data was conducted. 
The generated images underwent filtering to reduce the number of out-of-class samples using the proposed filtering mechanisms.
Instances that did not meet these criteria were removed from the dataset. 
For a detailed description of the filtering process, please refer to the corresponding section in the paper.
}
\customformat{Was the ``raw'' data saved in addition to the preprocessed/cleaned/labeled data (e.g., to support unanticipated future uses)?}{ If so, please provide a link or other access point to the “raw” data.}
{
The generated images remain in their original, unprocessed state and can be considered as ``raw'' data. However, we have not provided all the images that were filtered out.
}
\customformat{Is the software used to preprocess/clean/label the instances available?}{ If so, please provide a link or other access point.}
{
Generating the images was performed using commonly available Python libraries.
For annotating a subset of the dataset for filtering purposes, we have used the VIA annotation tool \citep{dutta2019vgg,dutta2016via}.
}

\subsection{Uses}

\customformat{Has the dataset been used for any tasks already?}{ If so, please provide a description.}
{
In our work, we demonstrate how this approach yields valuable insights into the robustness of state-of-the-art models, particularly in the context of classification tasks.
}
\customformat{Is there a repository that links to any or all papers or systems that use the dataset?}{ If so, please provide a link or other access point.}
{
The relevant links can be acquired via the project page \url{https://genintel.github.io/CNS}.
}
\customformat{What (other) tasks could the dataset be used for?}{}
{
Our work showcases the capability of our dataset to enhance control over data generation, which is particularly evident through continuous shifts. 
However, its applicability extends beyond this demonstration.
The dataset can be effectively utilized in various generation tasks that necessitate continuous parameter control. 
While we showcased its efficacy in providing insights for models tackling classification tasks, it can seamlessly extend to evaluate the robustness of state-of-the-art methods across diverse tasks such as segmentation, domain adaptation, and many others.
This is possible by combining our approach with other modes of conditioning Stable Diffusion.
In addition, our data can also be used for fine-tuning, which we also demonstrated in the supplementary material.
}
\customformat{Is there anything about the composition of the dataset or the way it was collected and cleaned that might impact future uses? For example, is there anything that might cause the dataset to be used inappropriately or misinterpreted (e.g., accidentally incorporating biases, reinforcing stereotypes)?}{}
{
Our dataset was synthesized using a generative model.
It, therefore, likely inherits any biases for its generator.
Similarly, filtering is performed by pre-trained models, which can indirectly also contribute to biases.
}
\customformat{Are there tasks for which the dataset should not be used?}{ If so, please provide a description.}
{
No, there are no tasks for which the dataset should not be used. 
Our dataset aims to enhance model robustness and provide deeper insights during model evaluation. Therefore, we see no reason to restrict its usage.
}

\subsection{Distribution}

\customformat{Will the dataset be distributed to third parties outside of the entity (e.g., company, institution, organization) on behalf of which the dataset was created?}{ If so, please provide a description.}
{
Yes, the dataset will be publicly available on the internet.
}
\customformat{How will the dataset be distributed (e.g., tarball on website, API, GitHub)? Does the dataset have a digital object identifier (DOI)?}{}
{
The dataset will be distributed as archive files on our servers.
}
\customformat{When will the dataset be distributed?}{}
{
The dataset will be distributed upon acceptance of the manuscript.
}
\customformat{Will the dataset be distributed under a copyright or other intellectual property (IP) license, and/or under applicable terms of use (ToU)?}{ If so, please describe this license and/or ToU, and provide a link or other access point to, or otherwise reproduce, any relevant licensing terms or ToU.}
{
CC-BY-4.0.
}
\customformat{Have any third parties imposed IP-based or other restrictions on the data associated with the instances?}{ If so, please describe these restrictions, and provide a link or other access point to, or otherwise reproduce, any relevant licensing terms.}
{
No, there are no IP-based or other restrictions on the data associated with the instances imposed by third parties.
}
\customformat{Do any export controls or other regulatory restrictions apply to the dataset or to individual instances?}{ If so, please describe these restrictions, and provide a link or other access point to, or otherwise reproduce, any supporting documentation.}
{
We are not aware of any export controls or other regulatory restrictions that apply to the dataset or to individual instances.
}

\subsection{Maintenance}

\customformat{Who is supporting/hosting/maintaining the dataset?}{}
{
The dataset is supported by the authors and their associated research groups. The dataset is hosted on our own servers.
}
\customformat{How can the owner/curator/manager of the dataset be contacted (e.g., email address)?}{}
{
The authors of this dataset will be reachable at their e-mail addresses.
}
\customformat{Is there an erratum?}{ If so, please provide a link or other access point.}
{
If errors are found, an erratum will be added to the website.
}
\customformat{Will the dataset be updated (e.g., to correct labeling errors, add new instances, delete instances)?}{ If so, please describe how often, when, and how updates will be provided.}
{
Yes, updates will be communicated via the website. The dataset will be versioned.
}
\customformat{If the dataset relates to people, are there applicable limits on the retention of the data associated with the instances (e.g., were individuals in question told that their data would be retained for a specific period of time and then deleted)?}{ If so, please describe these limits and explain how they will be enforced.}
{
Our dataset does not relate to people.
}
\customformat{Will older versions of the dataset continue to be supported/hosted/maintained?}{ If so, please describe how.}
{
No, older versions of the dataset will not be supported if the dataset is updated. We do not plan to extend or update the dataset. Any updates will be made solely to correct any hypothetical errors that may be discovered.
}
\customformat{If others want to extend/augment/build on/contribute to the dataset, is there a mechanism for them to do so?}{ If so, please provide a description. Will these contributions be made publicly available?}
{
Yes, we provide all the necessary tools and explanations to enable users to build continuous shifts for their own specific applications. 
Our dataset serves as a foundation to evaluate various classifiers. 
We encourage to build on top of this work and we are happy to link relevant works via our GitHub page.
}
\subsection{Author Statement of Responsibility}
The authors confirm all responsibility in case of violation of rights and confirm the license associated with the dataset and its images.
\end{document}